\documentclass{ieeeaccess_arxiv}
\pdfoutput=1
\usepackage{cite}
\usepackage{amsmath,amssymb,amsfonts}
\usepackage{algorithmic}
\usepackage{graphicx}
\usepackage{textcomp}
\usepackage{subcaption}
\usepackage[mode=buildnew]{standalone}
\usepackage{multirow}
\usepackage{booktabs}
\usepackage{makecell}

\usepackage[breaklinks]{hyperref}
\usepackage{tabularx}

\newcommand{\VEC}[1]{\mathbf{#1}}          %
\newcommand{\VECG}[1]{\boldsymbol{#1}}     %

\newcommand{\putindex}[3]{\vtop{\hbox{\hspace{#3} $#1$}
            \hbox{\raise 6mm \hbox{$\scriptscriptstyle #2$}}}}

\newcommand{\gradx}[0]{\vtop{\hbox{\rm grad}
            \hbox{\raise 2.5mm \hbox{\rm \hspace{2mm} \footnotesize x}}}}

\newcommand{\grady}[0]{\vtop{\hbox{\rm grad}
            \hbox{\raise 2.5mm \hbox{\rm \hspace{2mm} \footnotesize y}}}}

\newcommand{\grad}[1]{\vtop{\hbox{\rm grad}
            \hbox{\raise 2.5mm \hbox{#1}}}}

\newcommand{\stz}{\rule{0mm}{2.3ex}}

\newcommand{\btb}{     \begin{tabbing}             }
\newcommand{\bte}{     \end{tabbing}               }

\definecolor{tu0}{rgb}{0.7451, 0.1176, 0.2353}

\definecolor{tu1}{rgb}{1.0000, 0.8039, 0.0000}
\definecolor{tu11}{rgb}{1.0000, 0.8627, 0.3020}
\definecolor{tu12}{rgb}{1.0000, 0.9020, 0.4980}
\definecolor{tu13}{rgb}{1.0000, 0.9412, 0.6980}
\definecolor{tu14}{rgb}{1.0000, 0.9608, 0.8000}

\definecolor{tu2}{rgb}{0.9804, 0.4314, 0.0000}
\definecolor{tu21}{rgb}{0.9882, 0.6039, 0.3020}
\definecolor{tu22}{rgb}{0.9882, 0.7137, 0.4980}
\definecolor{tu23}{rgb}{0.9922, 0.8275, 0.6980}
\definecolor{tu24}{rgb}{0.9961, 0.8863, 0.8000}

\definecolor{tu3}{rgb}{0.6902, 0.0000, 0.2745}
\definecolor{tu31}{rgb}{0.7529, 0.2000, 0.4196}
\definecolor{tu32}{rgb}{0.8431, 0.4980, 0.6353}
\definecolor{tu33}{rgb}{0.9216, 0.7490, 0.8196}
\definecolor{tu34}{rgb}{0.9529, 0.8510, 0.8902}

\definecolor{tu4}{rgb}{0.4863, 0.8039, 0.9020}
\definecolor{tu41}{rgb}{0.6431, 0.8627, 0.9333}
\definecolor{tu42}{rgb}{0.7412, 0.9020, 0.9490}
\definecolor{tu43}{rgb}{0.8431, 0.9412, 0.9686}
\definecolor{tu44}{rgb}{0.8980, 0.9608, 0.9804}

\definecolor{tu5}{rgb}{0.0000, 0.5020, 0.7059}
\definecolor{tu51}{rgb}{0.3020, 0.6510, 0.7961}
\definecolor{tu52}{rgb}{0.5490, 0.7765, 0.8667}
\definecolor{tu53}{rgb}{0.7490, 0.8745, 0.9255}
\definecolor{tu54}{rgb}{0.8510, 0.9255, 0.9569}

\definecolor{tu6}{rgb}{0.0000, 0.3255, 0.4549}
\definecolor{tu61}{rgb}{0.2510, 0.4941, 0.5922}
\definecolor{tu62}{rgb}{0.5490, 0.6941, 0.7529}
\definecolor{tu63}{rgb}{0.7490, 0.8314, 0.8627}
\definecolor{tu64}{rgb}{0.8510, 0.8980, 0.9176}

\definecolor{tu7}{rgb}{0.0314, 0.0314, 0.0314}
\definecolor{tu71}{rgb}{0.3725, 0.3725, 0.3725}
\definecolor{tu72}{rgb}{0.5882, 0.5882, 0.5882}
\definecolor{tu73}{rgb}{0.7529, 0.7529, 0.7529}
\definecolor{tu74}{rgb}{0.8667, 0.8667, 0.8667}

\definecolor{tu8}{rgb}{0.7765, 0.9333, 0.0000}
\definecolor{tu81}{rgb}{0.8431, 0.9529, 0.3020}
\definecolor{tu82}{rgb}{0.8863, 0.9647, 0.4980}
\definecolor{tu83}{rgb}{0.9333, 0.9804, 0.6980}
\definecolor{tu84}{rgb}{0.9569, 0.9882, 0.8000}

\definecolor{tu9}{rgb}{0.5373, 0.6431, 0.0000}
\definecolor{tu91}{rgb}{0.6784, 0.7490, 0.3020}
\definecolor{tu92}{rgb}{0.7686, 0.8196, 0.4980}
\definecolor{tu93}{rgb}{0.8588, 0.8941, 0.6980}
\definecolor{tu94}{rgb}{0.9059, 0.9294, 0.8000}

\definecolor{tu10}{rgb}{0.0000, 0.4431, 0.3373}
\definecolor{tu101}{rgb}{0.3020, 0.6118, 0.5373}
\definecolor{tu102}{rgb}{0.5490, 0.7490, 0.7020}
\definecolor{tu103}{rgb}{0.7490, 0.8588, 0.8353}
\definecolor{tu104}{rgb}{0.8549, 0.9176, 0.9059}

\definecolor{tu110}{rgb}{0.8000, 0.0000, 0.6000}
\definecolor{tu111}{rgb}{0.8706, 0.3490, 0.7412}
\definecolor{tu112}{rgb}{0.9216, 0.6000, 0.8392}
\definecolor{tu113}{rgb}{0.9608, 0.8000, 0.9216}
\definecolor{tu114}{rgb}{0.9804, 0.8980, 0.9608}

\definecolor{tu120}{rgb}{0.4627, 0.0000, 0.4627}
\definecolor{tu121}{rgb}{0.5961, 0.2510, 0.5961}
\definecolor{tu122}{rgb}{0.7294, 0.4980, 0.7294}
\definecolor{tu123}{rgb}{0.8392, 0.6980, 0.8392}
\definecolor{tu124}{rgb}{0.9216, 0.8510, 0.9216}

\definecolor{tu130}{rgb}{0.4627, 0.0000, 0.3294}
\definecolor{tu131}{rgb}{0.6118, 0.3020, 0.5333}
\definecolor{tu132}{rgb}{0.7569, 0.5490, 0.6980}
\definecolor{tu133}{rgb}{0.8667, 0.7490, 0.8314}
\definecolor{tu134}{rgb}{0.9216, 0.8510, 0.9020}
\DeclareMathOperator*{\argmax}{arg\,max}
\newcommand{\src}[0]{{\mathcal{D}^S}}
\newcommand{\tgt}[0]{{\mathcal{D}^T}}

\newcommand{\network}[1]{{{#1}}}

\newcommand{\encoder}[0]{{\mathbf{E}}}
\newcommand{\decoder}[0]{{\mathbf{G}}}
\newcommand{\classifier}[0]{{\mathbf{C}}}
\newcommand{\discriminator}[0]{{\mathbf{D}}}

\newcolumntype{C}[1]{>{\centering\arraybackslash}p{#1}}
\def\BibTeX{{\rm B\kern-.05em{\sc i\kern-.025em b}\kern-.08em
    T\kern-.1667em\lower.7ex\hbox{E}\kern-.125emX}}

\begin{document}
\history{}
\doi{}

\newcommand{\question}[1]{{\color{blue}#1}}
\newcommand{\comment}[1]{{\color{green}#1}}
\newcommand{\todo}[1]{{\textit{\color{red}#1}}}

\hypersetup{ 	
pdfsubject = {Domain Adaptation},
pdftitle = {Survey on Domain Adaptation},
pdfauthor = {Schwonberg et al.}
}

\title{Survey on Unsupervised Domain Adaptation for Semantic Segmentation for Visual Perception in Automated Driving}
\author{\uppercase{Manuel Schwonberg}\authorrefmark{1*,4},
\uppercase{Joshua Niemeijer\authorrefmark{2*}, Jan-Aike Termöhlen}\authorrefmark{3*}
\IEEEmembership{Student Member, IEEE}, \uppercase{Jörg P. Schäfer}\authorrefmark{2}, \uppercase{Nico M. Schmidt}\authorrefmark{1}, \uppercase{Hanno Gottschalk}\authorrefmark{4}, \uppercase{Tim Fingscheidt}\authorrefmark{3}\IEEEmembership{Senior Member, IEEE}}
\address[1]{CARIAD SE (e-mail: \{manuel.schwonberg, nico.schmidt\}@cariad.technology)}
\address[2]{Deutsches Zentrum für Luft- und Raumfahrt e.V (e-mail: \{joshua.niemeijer, joerg.schaefer\}@dlr.de)}
\address[3]{Institute for Communications Technology, Technische Universität Braunschweig, 38106 Braunschweig, Germany (e-mail: \{j.termoehlen,t.fingscheidt\}@tu-bs.de)}
\address[4]{Institute of Mathematics, Technical University Berlin, Germany (e-mail: gottschalk@math.tu-berlin.de)}
\tfootnote{The research leading to these results is funded by the German Federal Ministry for Economic Affairs and Energy within the project “KI Delta Learning" (Förderkennzeichen 19A19013C and 19A19013K). The authors would like to thank the consortium for the successful cooperation.}

\markboth
{Schwonberg \headeretal: Survey on Unsupervised Domain Adaptation for Semantic Segmentation for Visual Perception in Automated Driving}
{Schwonberg \headeretal: Survey on Unsupervised Domain Adaptation for Semantic Segmentation for Visual Perception in Automated Driving}

\corresp{* indicates an equal contribution. Corresponding author: Jan-Aike Termöhlen (e-mail: j.termoehlen@tu-bs.de).}

\begin{abstract}
Deep neural networks (DNNs) have proven their capabilities in many areas in the past years, such as medical applications, robotics, or automated driving, enabling technological breakthroughs.
DNNs play a significant role in environment perception for the challenging application of automated driving and are employed for tasks such as detection, semantic segmentation, and sensor fusion.
Despite this progress and tremendous research efforts, several issues still need to be addressed that limit the applicability of DNNs in automated driving.
The bad generalization of DNNs to new, unseen domains is a major problem on the way to a safe, large-scale application, because manual annotation of new domains is costly, particularly for semantic segmentation.
For this reason, methods are required to adapt DNNs to new domains without labeling effort.
The task, which these methods aim to solve is termed unsupervised domain adaptation (UDA). 
While several different domain shifts can challenge DNNs, the shift between synthetic and real data is of particular importance for automated driving, as it allows the use of simulation environments for DNN training.
In this work, we present an overview of the current state of the art in this field of research. 
We categorize and explain the different approaches for UDA. 
The number of considered publications is larger than any other survey on this topic. The scope of this survey goes far beyond the description of the UDA state-of-the-art. 
Based on our large data and knowledge base, we present a quantitative comparison of the approaches and use the observations to point out the latest trends in this field. 
In the following, we conduct a critical analysis of the state-of-the-art and highlight promising future research directions. 
With this survey, we aim to facilitate UDA research further and encourage scientists to exploit novel research directions to generalize DNNs better.

\end{abstract}


\titlepgskip=-15pt

\maketitle

\section{Introduction}
\label{sec:introduction}
\PARstart{P}{erception} of the environment using a variety of sensors is an essential component of modern autonomous systems, e.g., automated vehicles~\cite{fingscheidt_dnndataautomateddriving, Houben2022}.
\begin{figure}[t]
    \centering
    \includegraphics[width=1.0\linewidth]{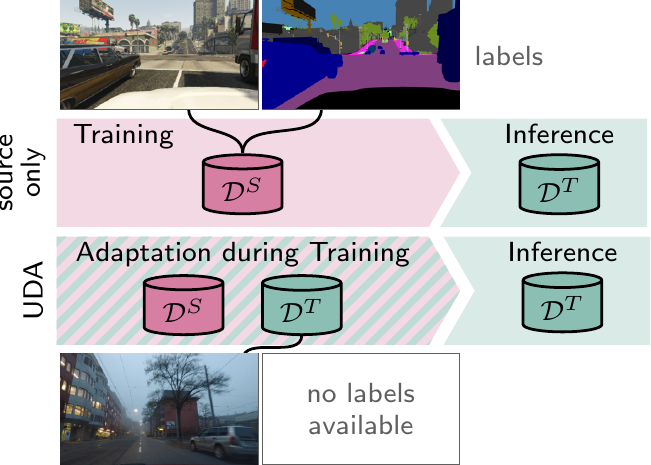}
    \caption{\textbf{Unsupervised domain adaptation} scheme illustrating the difference between source-only training and unsupervised domain adaptation (UDA). \textcolor{tu31}{Red} indicates (the use of) labeled data from the source domain and \textcolor{tu101}{green} indicates that unlabeled data from the target domain is utilized.}
    \label{fig:overview}
\end{figure}
Visual perception using camera sensors is of particular interest, as this type of sensor is inexpensive and provides diverse information not only on the geometry of the environment but also on the surfaces, e.g., colors, textures, and text on traffic signs.
The perception that converts sensory to semantic information utilizes machine learning methods such as deep neural networks (DNNs) that are trained on a large amount of human-labeled training data. 
This principle led to remarkable results \cite{krizhevsky2017imagenet} and a newly emerging field of machine learning research. 
Nowadays, DNNs are employed for a large variety of tasks and areas, but we will focus on DNNs for semantic segmentation in this survey.\par
Despite the capabilities of DNNs to learn complex relations, there are still several challenges to solve. 
First, adversarial attacks \cite{akhtar2021advances} that are designed to cause wrong predictions are a severe threat to DNNs in safety-critical areas such as automated driving. 
Second, the large amounts of human-labeled data that are required for training are costly, and the training of large models on thousands or millions of images also. 
It is worth mentioning that self-supervised learning substantially reduced the need for labels\cite{jaiswal2020survey}.
One of the most severe issues is the need to generalize DNNs to samples outside their training distribution.
Domain shifts, i.e., when training and inference distribution differ, occur often and can have multiple causes, such as changes in illumination, location, weather conditions, or sensor noise.
Significant performance degradation of the model is usually the consequence. 
A domain shift also occurs when training on synthetic data and employing the model on real data. 
Utilizing synthetic data for training is of particular interest, because it offers labels without human annotation effort and allows critical situations to be simulated that would be too dangerous or too rare in real data recordings, e.g., accidents and children running onto the street.
For semantic segmentation, manual labeling effort by a human annotator is very high and can take up to 90 minutes per image \cite{Cordts2016}. 
However, since synthetic data can substantially differ from real data, significant performance drops are caused when a model trained on synthetic data receives real data.\par
To counteract the problems introduced by domain shifts, unsupervised domain adaptation (UDA) methods have emerged that require only unlabeled samples of the target domain to adapt the network to it.
Figure~\ref{fig:overview} illustrates the basic principle.
Instead of training only on labeled images from the source domain $\src$, for UDA there are also images from the target domain $\tgt$ available for adaptation, but without any labels. 
In both cases, the inference is supposed to be performed on data from $\tgt$.\par
One of the most important and challenging applications is automated driving, where the autonomous system has to handle a broad range of driving scenarios. 
Domain shifts that cause a perception performance drop can seriously threaten human lives.
For this reason, UDA methods are essential for automated driving.
Consequently, this survey focuses on UDA methods for environment perception in autonomous driving, particularly on semantic segmentation of camera images.
Semantic segmentation not only provides important information about objects but also about the environment surrounding the vehicle (the background classes), which can serve as the basis of a local grid map~\cite{Meyer2018mapless} or for map verification~\cite{Plachetka2020,Plachetka2021,Plachetka2022}. 
This makes segmentation a crucial part of the perception system. It is also one of the most commonly used applications in scientific publications on unsupervised domain adaptation methods for visual perceptions. 
UDA for object detection is also an active research area \cite{oza2021unsupervised} but is out of the scope of this survey.\par
The UDA works included in this survey focus on the synthetic-to-real domain shift. 
As described, this shift is of special importance for automated driving, and the prioritization is valuable to assess how useful synthetic data can be for real applications. 
Because the synthetic-to-real domain shift benchmarking dominates UDA research, it is a reasonable choice for the focus of the survey to provide an extensive, valuable large-scale comparison of different approaches. 
However, since many more domain shifts are relevant for automated driving, we critically discuss this question in Section \ref{subsec:methodical_reflection}.\par
The evolvement of UDA publications for semantic segmentation for synthetic-to-real adaptation supports the importance of UDA for automated driving in research. 
This work focuses on UDA methods proposed for semantic segmentation since 2017.
The research area of domain adaptation for machine learning already emerged early before the development of deep learning. 
However, only in 2017, the very first work \cite{hoffman2016fcns} for UDA for semantic segmentation utilizing modern DNNs was proposed. 
For this reason, this year is taken as the initial year of research for UDA for semantic segmentation.\par
Several survey papers have attempted to provide a structured overview of this research field in recent years. 
Here we have to distinguish between surveys on domain adaptation in general \cite{wilson2020survey, zhang2021survey, wang2018deep} and surveys specifically for \textit{unsupervised} domain adaptation for semantic segmentation \cite{toldo2020review, csurka2021unsupervised}. 
The general domain adaptation surveys do not focus on domain adaptation for semantic segmentation and therefore provide not the same depth and quantity as our survey. Toldo et al.\ \cite{toldo2020review} and Csurka et al.\ \cite{csurka2021unsupervised} published specialized surveys about domain adaptation for semantic segmentation and are closest to our survey. Both cover a significantly smaller amount of papers (factor of three). 
Toldo et al.\ \cite{toldo2020review} include a performance comparison clustered according to the backbones, but with its publication in 2020, it is too old to cover recent trends. 
The clustering of UDA approaches is similar for both Toldo et al.\ \cite{toldo2020review} and Csurka et al.\ \cite{csurka2021unsupervised}. Our clustering also has commonalities with theirs. However, we extend the known taxonomy by the area of hybrid domain adaptation and provide more fine-grained sub-groupings for each area.

In summary, with this survey, our key contributions are: 
\begin{itemize}
  \item We propose a simple taxonomy that helps to group the different works. 
  We also cover vision transformer networks for UDA as the very first survey.
  \item We provide the most complete literature study and comparison up to date by surveying three times more papers than previous studies.
  \item We provide a quantitative comparison of the different methods, showing methodical and performance trends of the last years. 
  \item We take a critical look at the current approaches to overcome domain shifts and highlight common problems in the training process and evaluation of the adaptation approaches.
  \item We point out promising future research questions in this research area.
 \end{itemize}
 
\noindent In addition, we have created an interactive project website that allows a more detailed comparison of all methods than in written form alone.
The raw data of our quantitative comparison is also provided for further utilization by other researchers. 
The website can be accessed at: \href{https://uda-survey.github.io/survey/leaderboard}{\color{tu61}{https://uda-survey.github.io/survey/}}\footnote{We update the website with the latest approaches on a bi-monthly basis so that it can serve as a data hub for UDA researchers and keeping track of the current state-of-the-art.}. \par%
Our survey is expected to be valuable for three groups of readers: beginners, experts, and lecturers. 
This survey provides a structured introduction for readers without any prior knowledge of the topic (unsupervised) domain adaptation.
We also discuss the most important benchmarking methods, and together with the quantitative comparison, our survey provides entrance for future UDA researchers.
We also hope that expert-level readers find this survey helpful because of the more complete field coverage compared with prior works.
In the dynamic research environment of UDA, expert-level readers will potentially find our overview of the most recent developments useful. For lecturers, the taxonomy and comparison of the individual methods provide interesting information. In addition, we compare the task of unsupervised domain adaptation with other related methods.\par %
The survey is structured as follows. 
In Section~\ref{sec:research_context}, we provide context for related research topics and introduce mathematical definitions and principles of domain adaptation. 
In Section~\ref{sec:methods}, we present our taxonomy and explain the common methods for each part of the taxonomy. 
Section~\ref{sec:evaluation} describes the employed metrics and provides a quantitative comparison of the approaches.
Finally, we discuss the current research, presenting proposals for refinements and best practices as well as promising research directions in Section~\ref{sec:discussion}, before finishing with our conclusions in Section~\ref{sec:conclusions}. 
Overall, we categorize, analyze and compare more than 140 approaches methodologically and quantitatively.

\section{Research Context and Definitions}
\label{sec:research_context}%
\begin{figure}[t]
    \centering
    \includegraphics[width=1.0\linewidth]{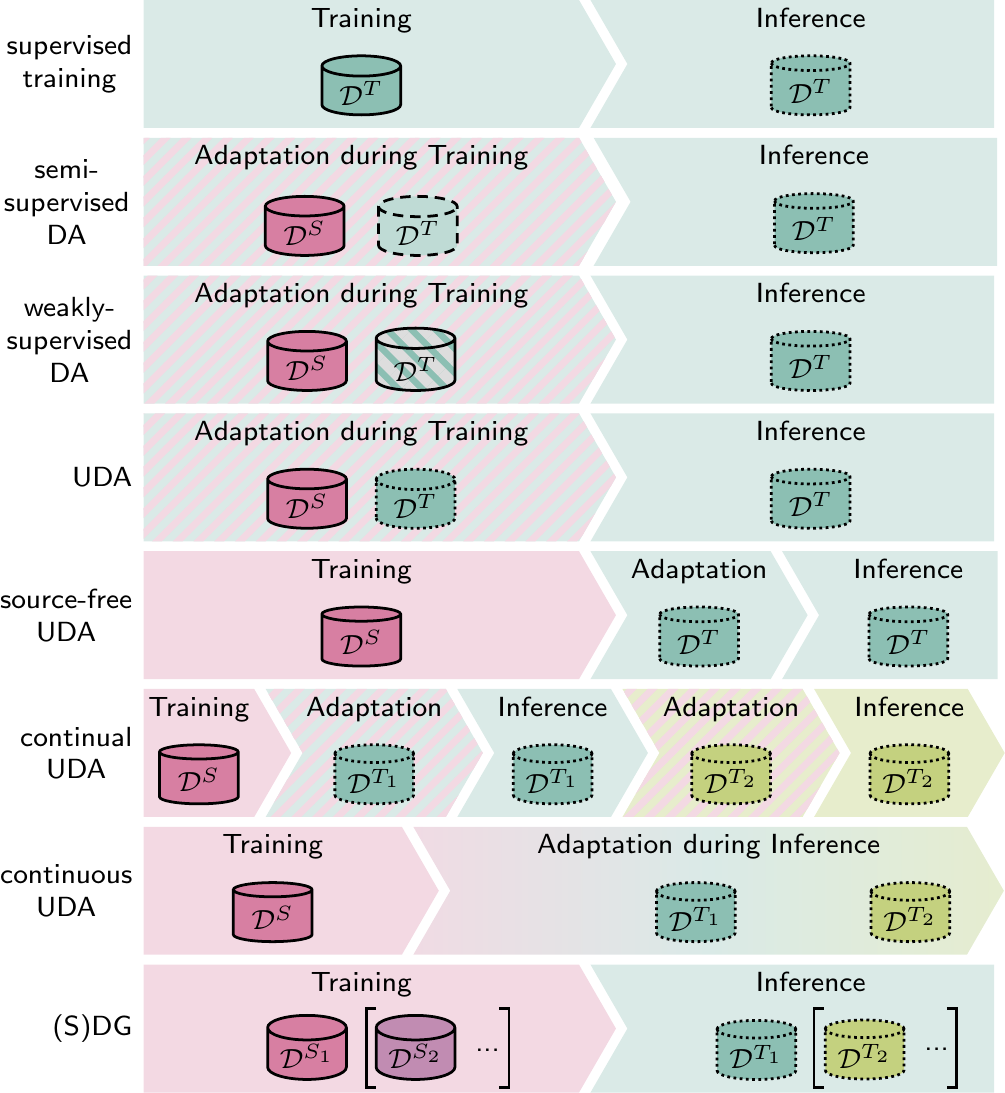}
    \caption{\textbf{Overview of adaptation paradigms.} Simplified schematic visualization. Red and green colors represent the source and target domains, respectively. Dotted lines indicate that no labels are available and dashed lines indicates a subset of labels. For weakly-supervised DA noisy labels are available.}
    \label{fig:adaptation_paradigmas}
\end{figure}
As mentioned before, UDA assumes that no labels are available for the target domain (cf.\ Fig.~\ref{fig:overview}). 
Some tasks are closely related to domain adaptation.
Figure \ref{fig:adaptation_paradigmas} shows an overview of the most important adaptation paradigms.
In the following section, we first give a brief overview of these related methods. 
Next, we introduce a mathematical notation to facilitate understanding of the various UDA methods.
Then, we define domain shifts and describe the most commonly used benchmarks.
\subsection{Related Methods}\label{subsec:related_methods}
In the following, we briefly explain other approaches for domain adaptation, which either use some form of labels or consider other constraints that apply to the availability of source domain data. 
Figure~\ref{fig:adaptation_paradigmas} provides an overview that categorizes the existing methods for domain adaptation found in the literature.
We also differentiate the topic from domain generalization and explain the difference between closed- and open-set adaptation. \\
\textbf{Supervised training}:
As visible in Figure \ref{fig:adaptation_paradigmas} (first row), the setting with full labels available in the target domain is called supervised training. 
In this case, no adaptation methods are required. 
However, this training paradigm is infeasible since manual annotation is costly and the number of domains is high. 
This way of adaptation is not scalable and will not be considered further. \\
\textbf{Semi-supervised DA}:
In semi-supervised domain adaptation (cf.\ Figure \ref{fig:adaptation_paradigmas}, second row), only a subset of the target domain data has labels. 
In contrast to self-training, these labels are not generated by the segmentation network itself but originate from human annotations.
Only a few works study methods for semi-supervised domain adaptation \cite{wang2020alleviating, mutze2022semi, chen2021semi, chen2021semi2}, but this research area is small compared to UDA and not the focus of this survey.  \\
\textbf{Weakly-supervised DA}:
For weakly-supervised domain adaptation (cf.\ Figure \ref{fig:adaptation_paradigmas}, third row), samples with noisy labels are available where, e.g., only bounding box labels are given as a weak label for the task of semantic or instance segmentation~\cite{Hanselmann2021weakly}.
Another option is image-level labels, e.g., to predict the presence of classes, as done in WDA \cite{paul2020domain}.
\\\textbf{Source-free UDA}: This is the task of UDA when there is a given model but no access to the data of the source domain (cf.\ Figure \ref{fig:adaptation_paradigmas}, fifth row). 
For standard UDA, the adaptation process primarily utilizes source and target domains in parallel. 
In source-free UDA, in contrast, the adaptation process to the new domains must occur without forgetting essential information from the source domain. 
In this case, the only information that may be used from the source domain is the implicit information in the network weights from the pre-training on the source domain, which includes normalization parameters~\cite{Klingner2020c,Klingner2020d}. 
\\\textbf{Continual/continuous UDA}:
The main idea behind continual and continuous domain adaptation is that a segmentation network is first trained on the labeled source domain and only afterward adapted to a target domain. 
During the adaption process, the source domain is unavailable.
It is, therefore, a decoupling of training and adaptation, which happen simultaneously in standard UDA.
Here, we want to propose a distinction between continual and continuous UDA (cf.\ Figure \ref{fig:adaptation_paradigmas}, sixth and seventh row). 
While continual means that something happens at (regular) intervals, continuous means that the adaptation happens without interruption, e.g., on a single-frame basis~\cite{Klingner2020d,Termoehlen2021}.
Usually, continual and continuous UDA does not prohibit the usage of some form of source domain representation, e.g., a generator network that creates samples~\cite{Wulfmeier2018}.
However, continual/continuous UDA methods can also be source-free when no source information is available during the adaptation process~\cite{Klingner2020d}.
\\\textbf{Domain generalization}: 
This is the task of training networks to perform better in unseen domains without using any data from the target domain for adaptation (cf.\ Figure \ref{fig:adaptation_paradigmas}, last row).
The model should generalize well, and the performance is usually evaluated on multiple unseen domains. 
The task is called single-domain generalization (SDG) if only one source domain is used.
It is simply called domain generalization (DG) if multiple source domains or additional real auxiliary domains are employed.
Compared to UDA, only a few works exploit the potential of domain generalization for semantic segmentation \cite{yue2019domain, lee2022wildnet, choi2021robustnet}. 
If it were possible to train networks that generalize perfectly, then adaptation to a target domain would no longer be necessary. 
On the other hand, it has been shown that even UDA approaches, e.g., DLOW~\cite{gong2019dlow}, can also ensure that the trained network generalizes better. However, since this is not the focus of the UDA task, these results are often not reported in the papers.  
\\\textbf{Closed- and open-set adaptation}:
Concerning the labeled classes in the source and target domains, a distinction is made between closed-set and open-set adaptation.
Closed-set adaptation refers to the more commonly employed method of domain adaptation, where the set of labeled classes is the same in the source and target domain. 
Thus it is assumed that only the visual domain changes, but the number of pre-defined semantic classes remains unchanged. 
In contrast, open-set adaptation assumes that the sets of labels do not have to be identical and that, e.g., new classes that have to be learned can occur in the new domain.
Even though this field has not yet been subject to intensive research, there are first approaches that enable class- \cite{uhlemeyer2022towards} and domain-incremental learning~\cite{Kalb2021classdomain}.
Although the open-set scenario is realistic, it is seldom considered in the academic context; see Tada et al.\  \cite{gong2021tada}, however, for first steps. 
This survey only covers UDA methods that perform closed-set adaptation.

\subsection{Mathematical Notation}
So far, papers on domain adaptation do not use a harmonized mathematical notation. 
We first present a unified mathematical notation to improve understanding and allow a more straightforward comparison of different methods.\\
As input to the segmentation network, we define the image $\mathbf{x} \in \mathbb{G}^{H \times W \times C}$, where $\mathbb{G}$ denotes the set of integer color intensity values, $H$ and $W$ the image height and width in pixels, and $C\!=\!3$ the number of color channels, respectively. 
A semantic segmentation network transforms an image into an output $\mathbf{y} = (y_{i,s})\in \mathcal{B}^{H \times W \times S}$ with posterior probability (score) $y_{i,s} = P(s|i,\mathbf{x})$ for each class $s \in \mathcal{S}$ at pixel index $i \in \mathcal{I} = \{1,2,...,H\cdot W\}$. 
Here, $\mathcal{S} = \{1,2,...,S\}$ denotes the set of $S$ classes and $\mathcal{B} = [0,1]$. 
The final segmentation map $\mathbf{m}=(m_{i})\in \mathcal{S}^{H \times W}$ is obtained with $\argmax$ operating on each pixel $i$ of the network output $\VEC{y} = (\VEC{y}_i)$ individually so that ${m}_i = \argmax_{s\in\mathcal{S}} {y}_{i,s}$. 
Note that $\mathbf{y}_i = (y_{i,s})$ is the vector of class posteriors at a pixel with index $i$. 
Superscripts ``S'' and ``T'' on $\mathbf{x}, \mathbf{y}$, and $\mathbf{m}$ denote the domain from which the variables stem, with, e.g., ${\src}$ being the source domain and ${\tgt}$ being the target domain.
\subsection{Definition of Domain Shifts}
The domain adaptation problem can be viewed as overcoming the dataset shift between the source and target domain distributions: $p_S(a,b)\neq p_T(a,b)$.
Where $p_S$ and $p_T$ represent the source and target distribution, and $a$ and $b$ are the feature and class variables, respectively, where both $a$ and $b$ are defined and used separately only for this explanation of domain shifts. 
We can distinguish between three distribution shifts to describe how the domains differ:
the prior, covariate, and concept shift \cite{kouw2018introduction}.\par
The prior shift occurs when $p_S(a|b)=p_T(a|b)$ but $p_S(b)\neq p_T(b)$.
The prior shift describes a change in class distribution.
An example of this shift can be found in the distribution of classes that may differ between domains. 
In a synthetic source domain an abundance of pedestrians might be rendered, while they are rare in the real-world target domain.\\
For the covariate shift, in contrast, $p_S(b|a)=p_T(b|a)$ but $p_S(a)\neq p_T(a)$ which means the input distribution changes. 
An example of the covariate domain shift is the difference in styles of the two domains, which can differ concerning, e.g., brightness, contrast, saturation, and hue.
Similarly, distributions can differ because objects or textures look different.\par
The concept shift refers to the case when $p_S(a)= p_T(a)$ but $p_S(b|a)\neq p_T(b|a)$ so that the conditional distribution differs and, therefore, the relations between $a$ and $b$ are different.
The same features in the source and target domain describe different classes. 
An example can be found in the synthetic-to-real domain shift case.
If a car in the synthetic world has a similar shape or texture as a truck in the real world, a concept shift has occurred. \par
In many practically relevant domain adaptation settings, the overall domain shift is caused by a mixture of prior, covariate, and concept distribution shifts. 
There are several such domain shifts relevant to computer vision systems. 
Training models on synthetic data for the application to real-world images introduces the synthetic-to-real domain shift.
Several real-to-real domain gaps exist, too.
Different sensors, locations, weather, day and night times, etc., can cause them.
Further domain gaps occur when a new generation of sensors is implemented in an autonomous vehicle, or the same sensor is mounted at different positions on a different car type.
Slight differences in illumination, resolution, noise, etc., can also lead to significant domain shifts. %
Since each domain gap would require retraining of models with new data and thus collecting and labeling this data is required, this can become very costly for large-scale applications. 
For this reason, domain adaptation or domain generalization methods are desired to overcome this issue and provide autonomous driving functions without needing a large-scale data selection and the corresponding data labeling effort.
\subsection{Benchmarks}
In this seection we will discuss commonly employed datasets, network architectures, and experimental setups in the field of UDA.
\subsubsection{Datasets}
\begin{table}[t]
    \centering
    
    \setlength{\tabcolsep}{.5em}
    \begin{tabular}{cccc}
        \hline
        \stz \textbf{Syn/Real} & \textbf{Dataset Name} & \textbf{\# Labeled Images} & \textbf{Resolution} \\
        \hline\hline
        \stz syn & GTA5~\cite{Richter2016} & 24,966 & $1914\!\times\!1052\phantom{^*}$ \\
        \stz syn & SYNTHIA~\cite{ros2016synthia} & 9,400 & $1280\!\times\!\phantom{1}760\phantom{^*}$ \\
        \hline
        \stz real & Cityscapes~\cite{Cordts2016} & 5,000 & $2048\!\times\!1024\phantom{^*}$ \\
        \stz real & NTHU~\cite{chen2017no} & 400 & $1280\!\times\!647\phantom{^*}$ \\
        \stz real & ACDC~\cite{Sakaridis2021} & 4,006 & $1920\!\times\!1080\phantom{^*}$ \\
    \end{tabular}
    \caption{\textbf{List of datasets} that are typically employed for UDA research. Shown are the total numbers of available labeled images, as well as the resolution of the images.}
    \label{tab:datasets} 
\end{table}
The selection of datasets used for UDA research is small and contains only two major synthetic and three real dataset, as shown in Table \ref{tab:datasets}, where only the Cityscapes~\cite{Cordts2016} dataset is commonly employed as a target domain. 
This simplifies the quantitative comparison. 
For the synthetic datasets, usually, GTA5~\cite{Richter2016}, extracted from the video game with the same name, and SYNTHIA~\cite{ros2016synthia}, specifically rendered for autonomous driving research, are utilized. 
It is worth mentioning that while GTA5 provides synthetic images from an ego-vehicle view perspective, SYNTHIA contains perspectives from both street-level and bird-eye-level views.
CARLA \cite{dosovitskiy2017carla} is another popular driving simulator that can be used to generate synthetic datasets. 
However, since no established dataset from CARLA exists, it rarely appears in UDA research for semantic segmentation\cite{mutze2022semi}.\par  
For most works, the target domain dataset is the established Cityscapes~\cite{Cordts2016} dataset.
We also include NTHU~\cite{chen2017no} and ACDC~\cite{Sakaridis2021} in this list. 
NTHU is rarely applied as a real-to-real domain shift evaluation benchmark in UDA papers.
This contrasts with the additional value from NTHU since it shows how the approaches perform on real city-to-city data.
ACDC does not appear in UDA synthetic-to-real works but is an often used benchmark for real-to-real adaptation from Cityscapes to ACDC with direct scene correspondences under adverse weather conditions.\\ 
SYNTHIA, GTA5, and Cityscapes in Table~\ref{tab:datasets} represent the de-facto standard in UDA research for semantic segmentation. See Csurka et al.\ \cite{csurka2021unsupervised} for a more extensive overview, including the number of classes, conditions, etc.
\subsubsection{DNN Architectures}
The most used segmentation networks in UDA are the \network{VGG16-FCN8} \cite{Long2015} and the \network{DeepLabv2} with a \network{ResNet-101} backbone \cite{Chen2018}. 
The more modern \network{ResNet}-based architecture dominates in more recent works. 
The variety of network architectures for UDA research is small, simplifying this survey's quantitative comparison.\par
Other architectures that were used in UDA works are the \network{MobileNetv2} as a backbone \cite{chao2021rethinking, toldo2021unsupervised}, \network{DRN-26} \cite{ye2020light, lee2021dranet, Chen2019crdoco}, \network{DRN-105} \cite{Saito2018a}, and smaller versions of the \network{ResNet} like \network{ResNet-18} \cite{shan2020semantic}, \network{ResNet-38} \cite{Iqbal2020, Lian2019, Zou2018} and \network{ResNet-50} \cite{Dong2019a, li2020variational}. However, all these architectures appear only rarely in UDA research.\par
The number of vision transformer architectures (see Section \ref{subsec:visiontransformers}) used for UDA is small since the research on such architectures for UDA only started recently. \network{DAFormer} \cite{Hoyer2022DAFormer}, which is based on \network{SegFormer} \cite{xie2021segformer} was the first work; \network{HRDA} has \network{DAFormer} as the basis. TransDA \cite{chen2022smoothing} uses \network{SwinFormer} \cite{liu2021swin} as its transformer architecture.    
\subsubsection{Standard Experimental Settings}
There exists a consensus in the research community for benchmarking UDA approaches because both datasets and architectures are the same in many works or at least very similar. 
Therefore, they provide a basic experimental setting for UDA benchmarks. 
However, in many training details the approaches differ significantly (like resolution, hyperparameters, dataset splits, etc.), so there are no unified benchmarking settings. 
A detailed discussion of these aspects is part of our discussion in Section~\ref{sec:discussion}.
\section{Unsupervised Domain Adaptation Approaches and Methods for Semantic Segmentation}
\label{sec:methods}
We discuss the methods developed for unsupervised domain adaptation in this chapter.
First, we explain our taxonomy. 
Afterward, we present the UDA methods of each part of our taxonomy in detail.
Finally, we review the latest UDA approaches using vision transformer networks.\par
We must fix two terms that we clearly distinguish throughout the remaining survey.
An \textit{approach} or \textit{method} is an entire paper that may include several different standalone \textit{techniques}. 
For instance, the Fourier domain adaptation (FDA) approach contains the techniques of Fourier-based style transfer and self-training, so FDA is an approach with two different techniques.
\subsection{Adaptation Spaces}
\begin{figure}[t]
    \centering
    \includegraphics[width=1.0\linewidth]{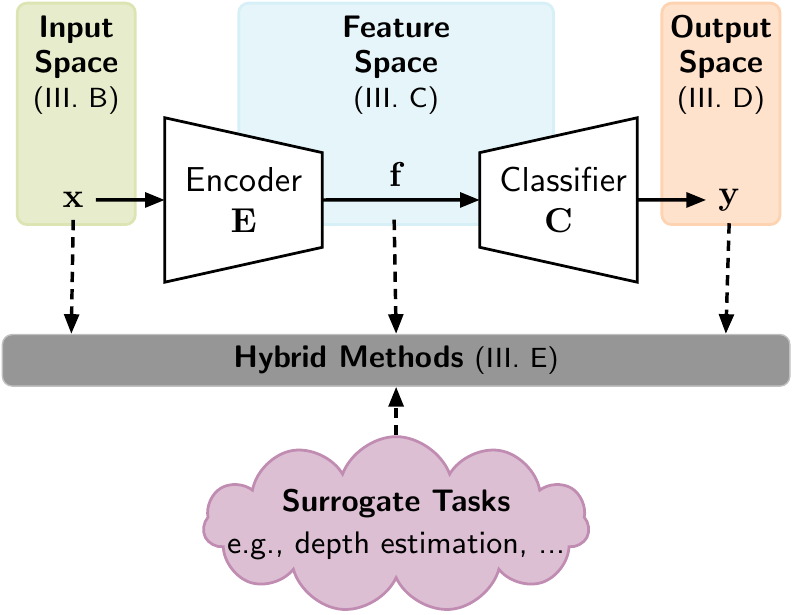}
    \caption{\textbf{Overview of adaptation spaces} that are covered in this survey. The subchapters dealing with the respective space are indicated. Hybrid methods perform adaptation in at least two spaces or utilize surrogate tasks.}
    \label{fig:adaptation_spaces}
\end{figure}
In order to approach the problem in a structured way, we have categorized the approaches. 
In deep learning, there are three typical spaces, i.e., the \textit{input space}, the \textit{latent representations} (\textit{features}) within the network, and the \textit{output} of the network. 
These spaces make up our three main adaptation categories illustrated in our taxonomy in Figure~\ref{fig:adaptation_spaces}. 
Approaches can combine methods in different spaces, which we view as approaches belonging to the category of \textit{hybrid} approaches. 
Additional surrogate tasks can help with adaptation but cannot be seen as a standalone category.
Toldo et al.\ \cite{toldo2020unsupervised} and Csurka et al.\ \cite{csurka2021unsupervised} propose a similar clustering by using the three spaces. 
Different from them and as one of our contributions, we introduce and analyze the category of hybrid domain adaptation, where we group approaches that combine two or more spaces.\par
The description of hybrid adaptation approaches is two-fold. 
First, the different techniques employed in the hybrid approaches are described in detail according to the standalone category of input, feature, or output space adaptation.
FDA as a hybrid approach, for instance, will be mentioned in the input and output space subsection since it contains techniques of both spaces.
Secondly, how the techniques of the different spaces interact and a more approach-level description will be provided in the section describing the hybrid approaches, where FDA will also appear. 
The advantage of this two-fold description is that the individual techniques of the hybrid approaches are contextually embedded within their category. 
In the hybrid section, the focus is purely on the different interactions between the spaces.\par
Unlike the previous surveys, we provide a fine-grained sub-grouping for each of the four categories. 
We provide a table that shows the sub-categorization of the approaches.
The idea behind these tables is that readers interested in a particular topic, e.g., output-level adversarial adaptation, can find a compact collection of all approaches employing this technique in the table. 
Approaches will appear multiple times if one approach consists of multiple techniques. 

\subsection{Input Space Domain Adaptation}
\label{sec:inputSpaceAlign}
\begin{figure}[t!]
  \centering
  \includegraphics[width=0.8\linewidth]{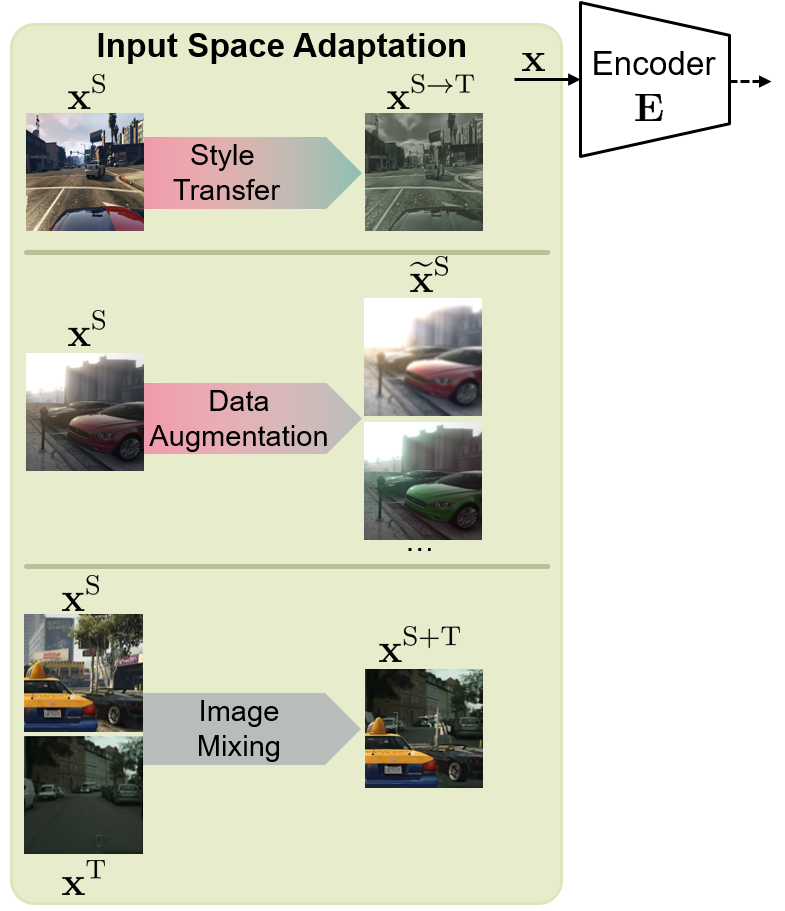}
  \caption{The three \textbf{main input space adaptation} methods are illustrated in this figure. Style transfer and data augmentation usually change the full appearance of the image, but style transfer does it in a more target domain directed manner. Image mixing generates images consisting of source and target domain pixels.}
  \label{figure:input_space}
\end{figure}
Unsupervised domain adaptation in the input space often refers to a change in the style of the images. The three main techniques employed in this adaptation space are depcited in Figure~\ref{figure:input_space}. \par
Typically, a distinction is made between the style and content of images. 
The semantic structure is often considered the same as the content.
Usually, it is described by well-defined low-level properties of images, such as hue, saturation, contrast, brightness, image-noise, depth of field, etc. 
However, the appearance of cars, e.g., their shape or texture, can also be counted as a style, even if the previously mentioned properties cannot express this.
The general idea with style transfer is to align the source and target domain distributions on a pixel-level in the input space.
This can happen, in the simplest form, at low-level image properties, such as hue, saturation, brightness, etc., e.g., by histogram
matching algorithms.
More complex methods, e.g., GAN-based methods, can vary the style even more and can change textures, depth of field, etc. 
However, a limitation of many approaches is that the semantics of the source images must remain the same so that the labels can still be used.
It is still hard to find a style transfer that maps, e.g., northern central-European vegetation, cars, traffic signs, etc., to, e.g., middle-eastern vegetation, different cars, and traffic signs.
The adaptation techniques that include object shape must be modeled by feature space adaptation. \par
When not only the style of images, but also the early feature maps are modified by the method, we would refer to it as a feature-level adaptation approach (cf. Section~\ref{sec:featSpaceAlign}).
During adaptation in the input space, the data samples $\VEC{x}$ used as the input for the method are modified. 
Multiple methods can be used to alter the input images and improve the performance in the target domain.
Usually, style transfer, content mixing, or data augmentation methods are employed. 
Style transfer methods try to match the target and source domains' style while not changing the samples' semantics.
For unsupervised domain adaptation, the style of the source domain, which is usually synthetic in scientific benchmarking, is transferred to be similar to the anticipated target domain, which is usually a real-world domain. \par
Domain adaptation, in most cases, is performed by more than style transfer methods.
These methods are often only part of the domain adaptation process and facilitate a more rapid convergence for subsequent methods. 
Nevertheless, some approaches solely rely on style transfer \cite{dundar2018a,gkitsas2019restyling,li2020generating}. 
Some methods perform the style transfer the other way around so that the target domain images are more similar to the source domain.
By doing this, the target domain does not have to be anticipated beforehand, but the style transfer method must be a part of the inference. 
These approaches can even be used for continuous domain adaptation under changing domains \cite{Termoehlen2021}.
So far, no simple style transfer technique has been able to achieve a state of the art performance for UDA on its own. 
The style transfer methods in the input space are typically detached from the further training process and are often combined with other methods. \par
The style transfer is the most popular approach and is usually performed utilizing feature transforms~\cite{dundar2018a, gkitsas2019restyling, lin2019adapting}, GAN-based networks, e.g., CycleGANs~\cite{Zhu2017}, normalization techniques, such as AdaIN~\cite{Huang2017b}, histogram matching~\cite{huang2021cross, ma2021coarse, lu2022bidirectional}, or image processing in the frequency domain \cite{yang2020fda, zhang2021spectral}. 
In recent years several image content mixing methods were proposed that mix the source and target directly in the image domain~\cite{tranheden2021dacs, Gao2021DSP,melas2021pixmatch, liu2021bapa, zhou2021domain, zhou2021context,wang2021domain,huo2022domain}. 
Also, several data augmentation methods~\cite{kim2020learning, huang2021rda, zhang2022unsupervised, zhou2020uncertainty} were proposed for input space UDA. \par %
In the following, we will briefly discuss the methods. An overview of these references is given in Table~\ref{tab:input_space_methods}.
Since the methods listed under Others do not specify the style transfer type used, they are not discussed in detail in the remainder of this chapter.
\begin{table}[t!]
    \centering
    \setlength{\tabcolsep}{.5em}
    \begin{tabular}{ccC{3.0cm}}
        \toprule
        \textbf{Technique} &\textbf{Sub-Cluster} & \textbf{Approach} \\
        \midrule
        \multirow{14}{*}{\textbf{Style Transfer}}&{Feature Transforms} & 
        \cite{dundar2018a}, %
        \cite{gkitsas2019restyling}, %
        \cite{lin2019adapting} %
        \\ \cmidrule{2-3}
        &{Normalization} &   
        \cite{musto2020semantically},
        \cite{luo2020adversarial},%
        \cite{choi2019self},%
        \cite{wang2020consistency}, %
        \cite{tang2021unsupervised}, %
        \cite{Ulyanov2016},%
        \cite{Huang2017b}, %
        \cite{wu2018dcan}%
        \\  \cmidrule{2-3}
         &{GAN-Based} & 
        \cite{gong2019dlow},
        \cite{ye2020light},
        \cite{lee2021dranet},
        \cite{Chen2019crdoco},
        \cite{toldo2020unsupervised},
        \cite{musto2020semantically},
        \cite{choi2019self},
        \cite{chen2019learning},
        \cite{sankaranarayanan2018learning},
        \cite{li2020generating}, %
        \cite{chang2019all},
        \cite{yang2020phase}, %
        \cite{song2020learning}, %
        \cite{hoffman2018cycada},
        \cite{yang2021context},
        \cite{ramirez2018exploiting}, %
        \cite{li2019bidirectional},
        \cite{kim2020cross},
        \cite{zhou2020affinity},
        \cite{wang2020differential}, %
        \cite{yang2020label}, %
        \cite{lee2018spigan}, %
        \cite{lee2020unsupervised}, %
        \cite{chung2021maximizing}, %
        \cite{cheng2021dual}, %
        \cite{xu2021self},
        \cite{gao2021addressing}, %
        \cite{dong2021and},
        \cite{saporta2020esl}, %
        \cite{li2020unsupervised},
        \cite{cai2020exploiting},
        \cite{li2020simplified}, %
        \cite{chiou2022beyond} %
        \\  \cmidrule{2-3}
         &Frequency Domain & 
        \cite{yang2020fda},
        \cite{zhang2021spectral}
        \\  \cmidrule{2-3}
         &Histogram Matching & 
        \cite{huang2021cross}, %
        \cite{ma2021coarse},%
        \cite{lu2022bidirectional}%
        \\\midrule
         \multirow{2}{*}{\textbf{Data Augmentation}}& & 
        \cite{kim2020learning},
        \cite{huang2021rda}, %
        \cite{zhang2022unsupervised},
        \cite{zhou2020uncertainty},
        \cite{araslanov2021self}%
        \\
        \midrule
        \multirow{2}{*}{\textbf{Image Mixing}}& & 
        \cite{Gao2021DSP},%
        \cite{melas2021pixmatch}, %
        \cite{liu2021bapa}, %
        \cite{zhou2021domain}, %
        \cite{zhou2021context},  %
        \cite{wang2021domain},  %
        \cite{huo2022domain}
        \\\midrule
       {\textbf{Others}}& & 
         \cite{paul2020domain},
        \cite{huang2021mlan} \\
        \bottomrule
    \end{tabular}
    
    \caption{Adaptation techniques in the \textbf{input space}. The papers are clustered and sub-clustered according to similar methodology.}
    \label{tab:input_space_methods} 
\end{table}
\noindent
\subsubsection{Style Transfer}
As already mentioned, style transfer is the primary input space adaptation technique. 
We will discuss style transfer using feature transforms, normalization techniques, image processing in the frequency domain, histogram matching, and GANs. 
Usually, style transfer is applied in one of two ways. First, the source images can be transferred to match the target domain during training (cf.\ Figure~\ref{figure:input_space}, upper part). 
In this case, during inference, no style transfer is needed.
Second, the target domain images can be transferred to match the source domain. 
With this setting, style transfer is also needed during inference. \\
\textbf{Feature transforms}:
Style transfer using feature transforms must be distinguished from feature space domain adaptation. 
The feature transforms presented here are methods that convert the images of the source domain into the style of the target domain with the help of a style transfer network. 
The features of the original segmentation network are not adapted during this process.
Instead, an additional network (usually an autoencoder) is trained on source and target images to transfer the style of the source images in their bottleneck features. \par 
Early works that employed a style transfer for unsupervised domain adaptation used simple feature transforms such as FastPhotoStyle~\cite{Li_2018_ECCV}, which comprises a two-step stylization and smoothing process. 
At first, the style of a content image is stylized in the  style of a style image from the target domain using an enhanced whitening and coloring transform (WCT)~\cite{Li2017c}, which is called PhotoWCT. In PhotoWCT, the upsampling layers of the style transfer network are replaced by unpooling layers. 
Afterward, smoothing is performed to ensure that semantically similar regions are stylized consistently. 
The FastPhotoStyle method~\cite{Li_2018_ECCV} was utilized by the domain stylization (DS)~\cite{dundar2018a} and the mask-aware gated discriminator (MAGD)~\cite{lin2019adapting} methods, which both randomly match source and target domain samples. 
Restyling data (RD)~\cite{gkitsas2019restyling} also employs FastPhotoStyle~\cite{Li_2018_ECCV} as a style transfer method and improves the sample matching by computing so-called perceptual hashes in the frequency domain of the images. These hashes are then used to match samples for the style transfer, and it is based on the Hamming distance of the respective hashes.\\ \noindent
\textbf{Normalization methods}:
The efficacy of normalization methods for style transfer has been known for some time~\cite{Ulyanov2016}.
Adaptive instance normalization (AdaIN)~\cite{Huang2017b} is particularly relevant in this context. The style transfer with AdaIN uses an encoder-decoder structure (usually based on a \network{VGG-19}~\cite{Simonyan2015} architecture), where the AdaIN layer receives the features of a content image (in the case of domain adaptation, usually an image from $\src$) and a style image (from $\tgt$ respectively). AdaIN then performs the style transfer by transferring the channel-wise mean and variance statistics of the features. AdaIN allows as many different style transfers to be learned for a (content) image as there are style images.\par
Methods such as DCAN~\cite{wu2018dcan} employ AdaIn and assume that the mean and standard deviation of the feature maps in an image generator encodes an image's style information.
They hence follow the idea to train an autoencoder in a way that it reconstructs images from the source domain $\src$. 
However, simultaneously, the mean and standard deviations are aligned between the source image that is to be reconstructed and a randomly selected image from the target domain $\tgt$.
Given that the feature statistics are matched, the generator will produce the source image in the target domain style.
As we shall see later in Section \ref{sec:featSpaceAlign}, this idea is also significant for the distribution alignment in the feature space.
The bi-directional style-induced domain adaptation (BiSIDA) \cite{wang2020consistency} employs a source-to-target style transfer for supervised training and a target-to-source style transfer for the unsupervised learning branch of the framework. The style transfer is performed  using the standard AdaIN method. 
Also, the CFContra method by Tang et al.~\cite{tang2021unsupervised} employs an encoder-decoder network with standard AdaIN layers for style transfer.\par
The adversarial style mining (ASM) method~\cite{luo2020adversarial} uses a newly proposed random AdaIN (RAIN) module for style transfer. 
RAIN adds a style variational autoencoder (VAE) in the latent space to encode the features' channel-wise mean and variance statistics into a Gaussian distribution that can be sampled from the latter.
During training, the RAIN module is trained to iteratively generate harder stylized images around the initial target sample according to the current learning state. 
That way, the segmentation model learns more potential styles in the target domain.\par  %
The target-guided and cycle-free data augmentation (TGCF-DA) method~\cite{choi2019self} employs a cycle-free generator network that is based on multimodal unsupervised image-to-image translation (MUNIT) \cite{Huang2018multimodal}. The generator is extended by AdaIN layers, which enable several style transfers (as many as there are style images) to be learned. 
The network is trained by a discriminator (distinguishing whether the image stems from the source or the target domain) and a semantic loss, ensuring that the semantics between the original source image and the style-transferred source image remains unchanged.\par
\noindent
\textbf{Frequency domain}:
Domain adaptation in the frequency domain is a relatively new field. 
Yang et al.~\cite{yang2020fda} proposed a new form of style transfer by implanting low-frequency information from the target images into the source images. 
This Fourier domain adaptation (FDA) is performed in the frequency domain. Only parts of the amplitude spectrum are exchanged, as these are assumed to contain the general style of the images.
Similar to FDA, the authors of SUDA~\cite{zhang2021spectral} employ a style transfer in the frequency domain. They decompose the input image into multiple frequency components and train a transformer network to recompose a newly stylized image from these frequency components. 
The  transformer network learns to suppress domain-variant contents and  enhance domain-invariant contents.  
\\ \noindent
\textbf{Histogram matching}:
Histogram matching is a long-established method~\cite{Pizer1987histogram} to match the style of images. However, only recently has there been research for its use for domain adaptation.  
Huang et al.~\cite{huang2021cross} tackle the task of panoptic segmentation, but the technique can also be employed for classical semantic segmentation. They propose an inter-style consistency, where the input images get stylized, and the segmentation masks between different styles, e.g., illumination or weather conditions, are learned to be equal. 
This is then combined with an inter-task consistency, which enforces consistent labels between a semantic segmentation and an instance segmentation network.
They employ a histogram-matching algorithm~\cite{Pizer1987histogram} for the stylization.
Ma et al.~\cite{ma2021coarse} propose a global photometric alignment for style transfer. 
They align the source and target images style by histogram matching in the three channels of the L*a*b* color space.
The same global photometric alignment is also employed by BiSMAP~\cite{lu2022bidirectional}.
\\ 
\label{subsec:ganbasedmethods}
\noindent\textbf{GAN-based methods}:
Generative adversarial networks (GANs) currently dominate the field of input space adaptation methods. 
GANs modify an image by a generator network so that a subsequent discriminator network can no longer distinguish from which domain the image originates.
By training a discriminator network, high-quality style transfers can be performed. 
In particular, CycleGAN \cite{Zhu2017} has proven to be a successful choice. It provides a photorealistic transformation between different image styles and mostly prevents semantic changes in the image due to the cycle consistency.
The goal is to learn a mapping function $\mathbf{G}\!:\!\src\!\mapsto\!\tgt$ as  well as an inverse mapping function $\mathbf{G}^{-1}\!:\!\tgt\!\mapsto\!\src$ and employ the cycle consistency to enforce that an image remains semantically the same after mapping and inverse mapping. However, most GAN-based methods are limited in terms of the variability of the stylized images. \par 
Methods such as MUNIT~\cite{Huang2018multimodal} that combine GANs with, e.g., AdaIN, try to overcome this limitation. 
They use AdaIN in their generator network to generate more specific style transfers.
The LSD method by Sankaranarayanan et al.~\cite{sankaranarayanan2018learning} was about the first to employ a standard GAN-based style transfer for domain adaptation. 
Also, Chen et al.~\cite{chen2019learning} employed a GAN for style transfer.
The domain invariant structure extraction (DISE) method~\cite{chang2019all} tries to disentangle the images' structure and texture during style transfer. 
This way, the structure and the texture of different source or target images can be combined. The method employs a least squares GAN (LSGAN)~\cite{Mao2016} and can be used in both directions. \par
Li et al.~\cite{li2020generating} follow a slightly different strategy as they do not employ a style transfer directly on the image level.
Instead, they propose a label-to-image domain adaptation (L2I-DA) transfer where they generate image-label pairs in the target domain style. They also employ a standard GAN for the image translation process.\par
DRANet~\cite{lee2021dranet} improves the style transfers from the generator network by searching the target features whose content component is most similar to the source features. The domain transfer is performed by incorporating style information from more suitable target features. \par
SPIGAN~\cite{lee2018spigan} simplifies the CycleGAN architecture by only using a single sim-to-real generator (no cycle consistency) and a downscaled generator network.
The light-weight calibrator (LWC) method~\cite{ye2020light} employs the ResNet generator proposed by Johnson et al.\ \cite{Johnson2016Perceptual} as a data calibrator. The calibrator can be seen as the generator. 
Two discriminators are employed, one on pixel level, and one on feature maps from the feature extractor.
The translation process is based on an adversarial distribution alignment of the feature space and a pixel-wise calibration network in the input space.  The pixel-wise calibration is based on an encoder-decoder architecture and is applied during inference, too.\par
Cai et al.~\cite{cai2020exploiting} propose a condition-guided style transfer by employing a standard conditional GAN~\cite{Mirza2014} that is trained with a semantic consistency loss. They also utilize concepts from StarGAN~\cite{Choi2018stargan} and BicycleGAN~\cite{Zhu2017bicyclegan}. This way, preferred styles like \texttt{foggy} or \texttt{cloudy} can be added to the images as needed.  \par
SUIT~\cite{li2020simplified} allows an improved style transfer by designing a novel semantic-content loss that focuses on label- and content-consistency between original and stylized images to guide the style transfer. The content-consistency is employed by comparing features of a pre-trained network for the stylized and the normal input images. \par
The stochastic image translation method by Chiou et al.~\cite{chiou2022beyond} is based on MUNIT~\cite{Huang2018multimodal}. 
The authors propose not performing an image-based but stochastic-style translation.
A source encoder encodes the content of the source image, and a target generator generates stylized versions of this image by sampling from a style distribution of the target domain.\par
The CycleGAN architecture, in particular, has been used and expanded by many papers as their style transfer network of choice. The CyCADA method~\cite{hoffman2018cycada} was among the first to perform a style transfer with a CycleGAN. It also explicitly encourages high semantic consistency before and after image translation for the source domain samples with a pre-trained source segmentation network. 
Also, CrDoCo~\cite{Chen2019crdoco}, MSS~\cite{toldo2020unsupervised}, and CADA~\cite{yang2021context} employ a standard CycleGAN for their image translation. 
Zhou et al.~\cite{zhou2020affinity} show that their ASANet+ is complementary to style transfer by combining their method with the image translation module from CyCADA~\cite{hoffman2018cycada}.\par
The SE-GAN method~\cite{xu2021self} makes adversarial training more stable and employs a simple CycleGAN for style transfer.
Yang et al.~\cite{yang2020phase} utilize a CycleGAN that uses both a cycle consistency and a phase consistency loss.
They show that the semantic information is mostly encoded in the phase from the complex spectrum of the image and enforce its similarity for the transformation and inverse transformation of the CycleGAN. \par
In DLOW~\cite{gong2019dlow}, the authors generate a sequence of intermediate domains between the source and target. They define a domainness factor $z$ that affects the generator and the discriminator. They also employ a cycle consistency loss and build their method upon CyCADA~\cite{hoffman2018cycada}. \par
 Another idea is to use a content invariant representation (CIR) \cite{gao2021addressing}, which can be seen as an intermediate domain between the source and target with the same content as the source domain and the same style distribution as the target domain.
They use a vanilla CycleGAN to generate this CIR. \par
A popular approach on which many works build is the bi-directional learning (BDL) method~\cite{li2019bidirectional}, which improves the image-to-image translation model by iteratively improving the translation model with feedback from the subsequent semantic segmentation model. 
This way, the image-to-image translation is not fixed but improves during training and adaptation.
The authors also published their style-transferred images from the GTA5~\cite{Richter2016} and SYNTHIA~\cite{ros2016synthia} datasets, which were used by many subsequent methods. For example, CDGA~\cite{kim2020cross}, SIM~\cite{wang2020differential}, MCSSF~\cite{chung2021maximizing}, and BDL+ESL~\cite{saporta2020esl} use this method or the already transferred images. \par
In contrast to previous works, the authors of LDR~\cite{yang2020label} train a style translation model that transfers the target domain images in order to make them look like source domain images. They employ the general translation model of BDL~\cite{li2019bidirectional} but add a cycle-reconstruction loss to enforce semantic consistency between the image and the image reconstructed from the labels.
The active pseudo-labeling (APL) method~\cite{song2020learning} first adapts the target domain images to the source domain using a style transfer. Afterward, the style-transferred images are used to create pseudo labels that are later used for self-supervised training in the target domain (cf.\ \ref{subsubsec:Hybrid_Domain_Alignment}). The style transfer is similar  to that of LDR\cite{yang2020label}, but it replaces the transposed convolutions with bilinear upsampling and convolutions. \par
Ramirez et al.~\cite{ramirez2018exploiting} employ a CycleGAN for style transfer from the source to the target domain in their image-level domain adaptation (ILDA) method. 
They enforce the similarity of segmentation masks based on style-transferred images and unaltered synthetic images using a discriminator in the generation process to avoid artifacts and guide the synthesis.
The DISE-CT method~\cite{lee2020unsupervised} is based on DISE \cite{chang2019all} but adds a cycle consistency to the generator training. It also adapts the zero loss~\cite{Benaim2019} to a zero-style loss. 
A content transfer is employed for long-tail classes of the target domain to incorporate more of these classes into the training samples. \par
Dual path learning (DPL)~\cite{cheng2021dual} employs two pipelines, where images are transferred from the source to the target domain or from the target to the source domain. Both pipelines are trained interactively with a so-called dual path adaptive segmentation. \par
With KATPAN, Dong et al.~\cite{dong2021and} employ a modified CycleGAN for the image translation process. They extend the standard CycleGAN with a transferability-aware information bottleneck that guides the encoder to encode only discriminative features. \par
Musto et al.\ propose a new semantically adaptive image-to-image (SA-ITI) translation~\cite{musto2020semantically}. They utilize the segmentation maps from the source image provided by the segmentation network to guide the style transfer of the source domain images to the target domain. 
As their style transfer network, they design two coupled GANs similar to a CycleGAN and adaptively denormalize each pixel based on the semantic information. 
The translated image is then fed to the segmentation network again. Consistency is enforced between the two output posteriors using a new symmetric cross-entropy loss. \par
However, there are also further enhancements of the CycleGAN architecture, e.g., the symmetric adaptation consistency (SAC) method uses a StarGAN~\cite{Choi2018stargan} for image-to-image translation.
\\ \noindent
\subsubsection{Data Augmentation}
An additional idea for domain adaptation in the input space is data augmentation. With data augmentation, the styles of the images are changed in a less targeted manner than with a style transfer (cf.\ Figure~\ref{figure:input_space}, middle part). 
Thus, no attempt is made to represent the target domain as precisely as possible. 
Instead, the images are changed as diversely as possible to train a network that is as robust as possible against various domain shifts. This is related to domain randomization, which is often used for domain generalization.\par
Zhou et al.~\cite{zhou2020uncertainty} perform a \textit{class out strategy} in the input space by employing a ClassDrop mask generation algorithm that provides class-wise perturbations. 
The learning texture invariant representation (LTIR) method~\cite{kim2020learning} generates a stylized version of the commonly used GTA5~\cite{Richter2016} and SYNTHIA~\cite{ros2016synthia} datasets to force the model to learn texture invariant representations, which are usually not learned from style-transferred images. \par
 Huang et al.~\cite{huang2021rda} train a more robust network against domain shifts by learning Fourier domain adversarial attacks and iteratively learning to defend against these attacks. 
These attacks are some form of style augmentation.
Araslanov et al.~\cite{araslanov2021self} perform heavy data augmentation and then calculate output consistency using differently augmented images.
The unsupervised contrastive domain adaptation (UCDA) method~\cite{zhang2022unsupervised} also employs multiple augmentation techniques on source and target domain images.
\\ \noindent
\subsubsection{Image Mixing}
Similar to data augmentation techniques, more and more methods have recently been developed that mix source domain images with portions of target domain images (cf.\ Figure~\ref{figure:input_space}, lower part).
One popular method is domain adaptation via cross-domain mixed sampling (DACS)~\cite{tranheden2021dacs}, on which many other methods have been built since. 
DACS mixes samples from the two domains along with the corresponding source labels and target pseudo-labels. 
The labeled source domain images and the mixed images are used for training.
It also applies color augmentation and Gaussian blurring to the training samples. \par
The RCCR approach~\cite{zhou2021domain} employs ClassMix~\cite{Olsson2021classmix} and CutMix~\cite{yun2019cutmix} as proposed by DACS~\cite{tranheden2021dacs}. Also BAPA-Net~\cite{liu2021bapa} solely employs CutMiX~\cite{yun2019cutmix}. 
Likewise, the CorDA method by Wang et al.~\cite{wang2021domain} is based on DACS~\cite{tranheden2021dacs} and utilizes all of its input space adaptations. 
Zhou et al.\ propose a new image mixing method termed CAMix~\cite{zhou2021context}, where they leverage contextual information on relationships to guide the image mixing. It can be seen as an improved version of DACS. \par
DBST~\cite{cardace2022plugging} adds depth guidance to DACS and the authors explicitly propose their method as a module that can be combined with any other UDA method like, e.g., ProDA \cite{zhang2021prototypical}.
The dual soft-paste (DSP) method~\cite{Gao2021DSP} improves on DACS~\cite{tranheden2021dacs} by pasting mainly long-tail classes from the source domain in source and target domain images. 
It creates two intermediate domains, which serve as a bridge between the domains.
They preserve the original domain information by keeping objects, layout, and general structure the same. \par
PixMatch~\cite{melas2021pixmatch} employs a consistency training with two different perturbations added to the images in its best working model. The authors show that Fourier domain and CutMix~\cite{yun2019cutmix} perturbations yield the best results.
\\ \noindent

\subsection{Feature Space Domain Adaptation}
\label{sec:featSpaceAlign}
\begin{figure}
  \centering
  \includegraphics[width=0.8\linewidth]{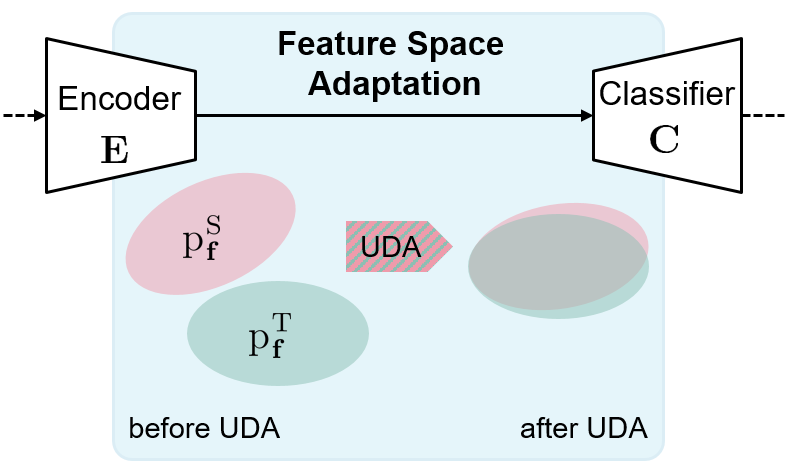}
  \caption{\textbf{Feature space adaptation methods:} The encoder of a CNN projects the input image to the feature space. Here before domain adaptation, the source and target distributions are not aligned. Hence the source domain-trained classifier does not generalize to the target domain. After feature space adaptation methods are used, the feature distributions are generally aligned much better, which improves performance on the target domain.}
  \label{figure:stages}
\end{figure}

\begin{table}[t]
    \centering
    \setlength{\tabcolsep}{.5em}
    \begin{tabular}{ccC{2.5cm}}
        \hline
        \multicolumn{2}{c}{\textbf{Technique}} & \textbf{Approach} \\
        \hline\hline
        \multirow{16}{*}{\makecell{\textbf{Distribution}\\\textbf{Divergence}}} 
        & Adversarial Training &
        \cite{paul2020domain},
        \cite{chen2017no},
        \cite{ye2020light},%
        \cite{Chen2019crdoco}, 
        \cite{Dong2019a}, 
       \cite{toldo2020unsupervised}, 
        \cite{hoffman2018cycada},%
        \cite{yang2021context}, 
        \cite{cheng2021dual}, 
         \cite{huang2021mlan}, 
        \cite{Hoffman2016}, 
        \cite{du2019ssf}, 
        \cite{li2020spatial},
        \cite{luo2019significance}, 
        \cite{luo2019taking}, 
        \cite{Hu2020SemanticDA}, 
        \cite{tsai2019domain}, 
        \cite{zheng2019deep}, 
        \cite{yu2021dast}, 
        \cite{wang2020classes}, 
        \cite{luo2021away}, 
        \cite{dong2020cscl}, 
        \cite{huang2018a}, 
        \cite{wang2019class}, 
        \cite{chen2018road}, 
        \cite{chen2021unsupervised}, 
        \cite{chen2020classification}, 
        \cite{zhang2020transferring} , 
        \cite{saha2021learning}, \cite{Bolte2019a}\\ \cmidrule{2-3}
        & Instance Norm/ Gaussian & 
        \cite{huang2018a}\\ \cmidrule{2-3}%
        & L2 Distance & 
        \cite{chen2018road}\\   \cmidrule{2-3}     
        & Max Classifier Discrepancy& 
        \cite{Saito2018a}, 
        \cite{li2020bi}, 
        \cite{lee2019sliced}\\\cmidrule{2-3}
        & Max Mean Discr. (MMD) & 
        \cite{gao2021dual}\\\cmidrule{2-3}
        & Shared Style GAN & 
        \cite{sankaranarayanan2018learning},
         \cite{lee2020unsupervised},%
        \cite{Hong2018}, %
        \cite{zhu2018penalizing}, 
        \cite{ruan2019category}, 
        \\
        \hline
         \multirow{2}{*}{\textbf{Instance Norm}} & & 
         \cite{Klingner2020c},
         \cite{Lian2019}, 
        \cite{wu2018dcan},
        \cite{Romijnders2019}, 
        \cite{Ioffe2017}, 
        \cite{Li2018d}\\
        \hline
         \multirow{8}{*}{\textbf{Self-Supervised}}
        & Contrastive & 
         \cite{zhang2022unsupervised}, 
         \cite{zhou2021domain}, 
        \cite{marsden2021contrastive}, 
        \cite{xie2021spcl}, 
        \cite{shim2021learning},
        \cite{liu2021domain},
        \cite{kang2020pixel}\\
        & Semantic Clustering & 
        \cite{toldo2021unsupervised}, 
        \cite{Dong2019a},
        \cite{tang2021unsupervised},
        \cite{kim2020cross},
        \cite{chung2021maximizing},  
        \cite{liu2021bapa}, 
         \cite{huang2021category},
        \cite{niemeijer2021combining}, 
        \cite{wang2021separable}, 
        \cite{li2021semantic}\\
        & Depth \& Ego Motion & 
        \cite{wang2021domain},
        \cite{saha2021learning}, 
        \cite{guizilini2021geometric}\\
        & Augmentation & 
        \cite{sun2019unsupervised},
        \cite{xu2019self2} \\
        & Weak Supervision & 
        \cite{paul2020domain}
        \cite{lv2020cross}\\
        \hline
    \end{tabular}
    
    \caption{Adaptation methods in the \textbf{feature space}. The papers clustered and sub-clustered according to similar methodology.}
    \label{tab:featurespace} 
\end{table}

As identified in the previous sections, the distribution shift between the source and the target domain leads to decreased performance.
Since the pre-logit feature space (the output of the last layer before the classifier) distributions of the source and target domain differ, a classifier trained on one cannot generalize well to the other (see Figure \ref{figure:stages}).
Hence, we discuss approaches that try to adapt the model from the source to the target domain, trying to align the distributions in the feature space. 
In this case, the alignment of the distributions depends on the learned encoder-decoder function that maps the input to the (pre-logit) feature space.
Therefore, distribution alignment between input data from the source and the target domain means learning the encoder-decoder function in a way that maps input from both domains that semantically represent the same things to a similar point in the feature space.
In this case, the classification hyperplane learned on the source domain will generalize well to the target domain.

One can identify different subclusters in the methods for distribution alignment in the feature space (see Table \ref{tab:featurespace}) that we will discuss in the following subsections.

\subsubsection{Distribution Divergence}
Methods that fall in this cluster try to minimize a divergence measure describing the distance between the source and the target domain.\\
\textbf{Adversarial adaptation}:
Ganin et al.\ \cite{Ganin2015} introduced the first and most prominent among these methods.
Although this paper deals with image classification, this work significantly impacted unsupervised domain adaptation for segmentation.
The authors define the distance between the source and the target domain as the so-called H-divergence. 
The H-divergence is computed based on a classifier, classifying whether the feature representation of an image is from the source or the target domain. 
Given such an optimal domain classifier, the H-divergence is minimal if the error of the optimal classifier is maximized. 
Minimizing the H-divergence poses a min-max problem. 
The H-divergence is minimal if the encoder is learned, so the domain classifier yields a maximum error rate. 
In contrast one has to minimize the error to obtain the optimal domain classifier. 
Ganin et al.\ \cite{Ganin2015}, propose a gradient-reversal layer (GRL) between the domain classifier and the feature space to solve this problem. 
During the backward pass, the gradients of the domain classification loss are applied to the domain classification head but inverted for the encoder-decoder function and thereby minimizes the H-divergence. \par
There is a multitude of methods (see Table \ref{tab:featurespace}) that make use of this idea for the task of semantic segmentation. 
\textit{FCNs in the Wild}~\cite{Hoffman2016} is the earliest example of adversarial learning in the feature space.
The proposed method applies the adversarial loss function on the pre-logit feature map (the last representation before the classifier). 
Hoffman et al.\cite{Hoffman2016}, however, implement the adversarial principle  in a different way. 
They define a domain classification loss that is used to optimize the domain classifier and the encoder-decoder weights of the segmentation network and use the inverse of this loss to update only the encoder-decoder weights.
Both losses are applied in an alternating way. 
This way of implementing the adversarial principle of the H-divergence, without a gradient reversal layer, is applied by the majority of UDA approaches. \par
Building on \textit{FCNs in the Wild}~\cite{Hoffman2016}, many approaches use adversarial training as an additional tool in their domain adaptation strategy, e.g., WDC \cite{zheng2019deep}, RPT \cite{zhang2020transferring}, DPL \cite{cheng2021dual}, CAA-Net \cite{ruan2019category}, and SWLS \cite{Dong2019a}. 
There are, however, many works that introduce new strategies to improve adversarial learning.\\ 
\textbf{Sub-distribution alignment}:
The alignment of the global distributions of the source and target domain can lead to issues.
A possible consequence of global distribution alignment is that sub-distributions of source and target domain that are closely aligned even before the adaptation are affected negatively by the global alignment (see Luo et al.\cite{luo2019taking}). 
Sub-distributions in our context denote parts of the source or target domain feature distributions that depend on, e.g., the classes or the spatial position. 
Wang et al.~\cite{wang2020classes} argue that parts of the class-wise sub-distributions might get mixed up through the global distribution alignment.
They further point out that the different frequencies of classes lead to the situation that the sub-distributions of frequent classes are aligned better than rare classes' sub-distributions.
Finally, Chen et al.\ \cite{chen2020classification} speculated that another issue with global distribution alignment through GANs are non-contributing ambiguous features. \par
Chen et al.\ \cite{chen2017no} and Du et al.\ \cite{du2019ssf} introduced early approaches to align the class-conditional sub-distributions. 
Their idea is to introduce a class-wise adversarial training.
The discriminator classifies between the source and target domain  only for feature representations of the same class.
The approaches use self-inferred pseudo labels on the target domain to implement the class-dependent domain classifier. 
Additionally, the approach weighs the adversarial loss higher for classes with low average confidence.
The work by Du et al.\ \cite{du2019ssf} improves the approach of Chen et al.\ \cite{chen2017no} by addressing the inconsistent adaptation issue.
 CCDA ~\cite{wang2019class} addresses the alignment of the class-conditional sub-distributions of classes with different frequencies.  
Wang et al.~\cite{wang2019class} employ two discriminator networks, one for coarse-level alignment and one for pixel-level alignment.
For the coarse-level alignment, the discriminator network predicts
the domain label of every coarse feature representation element and the classes present in the receptive field.
The second discriminator %
computes the adversarial loss pixelwise. 
The influence of each class is normalized with its frequency, giving
frequent and rare classes a similar weight. 
Additionally, they weigh spatial elements higher which possess a high classification uncertainty. 
FADA~\cite{wang2020classes} and CCDA~\cite{wang2019class} follow a very similar idea.

CCD~\cite{chen2020classification} tackles the problem that non-productive ambiguous features are learned during global distribution alignment through GANs.
To prevent this, they also train a segmentation loss on the sub-network in addition to the discriminator. 
However, the segmentation on this network is not backpropagated to the shared backbone.

Finally, ROAD ~\cite{chen2018road} assumes that similar classes occur at similar spatial positions in an image and uses the adversarial loss dependent on the spatial position. 
Their domain classification loss is computed for predefined regions (grid elements) in the image.
Given that similar classes and objects appear at similar spatial positions, the source and target domain distributions of grid elements match well a priori. 
The adversarial distribution alignment hence matches sub-distributions that are similar. 
\\\textbf{Adversarial training with attention}:
The following approaches aim at guiding the adversarial adaptation process to the most relevant regions. 
For this purpose, the approaches utilize an implicit or an explicit way of guiding the attention of the adaptation process. \par
Li et al.\ \cite{li2020spatial} use spatial and channel attention to achieve this goal.
They create a so-called highly embedded feature vector representing information about the feature space, the network prediction, and spatial and channel-wise attention maps.
The adversarial training is done based on this feature vector so the goal goal is to align the distributions of the source and target domain of the vector. 

DAST ~\cite{yu2021dast} uses discriminator confidences to measure the alignment of the source and target domain. 
After an initial adversarial alignment, the authors weight the feature map of the target domain with the domain classification output.
A high confidence score of the discriminator indicates that a feature representation is easily identifiable as part of the target domain. 
Hence such a feature representation still has to be aligned to the source domain and is given a high weight.

Chen et al.\ \cite{chen2021unsupervised} do not directly model the attention via a measure for the distribution alignment or a spatial attention module.
They instead assume that the semantic edges or boundaries between classes are significant for predicting semantic segmentation.
Thus the network comprises semantic and edge (class boundaries) segmentation branches.
In order to make the edge predictions domain-invariant, adversarial training in the edge branch feature space and feature fusion between the semantic and edge branch is applied.
The edge feature distribution alignment guides the attention implicitly to the class boundaries.
\\\textbf{Adversarial training on style}:
As described in Section \ref{sec:inputSpaceAlign}, style transfer enables supervised training on source domain images with the style of target domain images. 
Apart from that, several approaches utilize style transfer for adversarial adaptation in the feature space.

CyCaDa presented by Hoffman et al.~\cite{hoffman2018cycada} and as described in Section \ref{sec:inputSpaceAlign} trains a CycleGAN network to transform source domain images into the target domain and vice versa. 
Apart from this main contribution, the approach applies adversarial learning between the stylized source domain images and the not transformed target domain images. 
The discriminator distinguishes between the respective feature representation of stylized source domain and the target domain images.

CrDoCo \cite{Chen2019crdoco} trains two segmentation networks on the source domain labels, a source, and a target domain network. 
The target domain network is trained with the source domain labels but with style-transferred images.
Two separate discriminators enable the adversarial training in the feature space of the two networks.
The discriminator for the source domain network takes feature representations of source domain images and target domain images that were transferred to the source domain. 
Vice versa, the target domain network is trained. %
A consistency loss between the outputs of the two networks for the target domain images and the transferred target domain images is applied.
The fact that the source and target domain only differ in style but not in content facilitates the distribution alignment.
The authors of MSS \cite{toldo2020unsupervised} follow a similar approach as in CrDoCo \cite{Chen2019crdoco}.
The main difference to CrDoCo \cite{Chen2019crdoco} is that the encoder computing the feature representations is shared between the domains. 

LWC \cite{ye2020light} is different from the previous works
because it tries to align the distributions of the source and target domain not by altering the encoder but by transforming the input image. 
The authors of LWC \cite{ye2020light} present a calibrator strategy for domain adaptation. 
Given a model trained on the source domain, the aim is to train a calibrator network that transforms the input image in such a way that the distributions of the source and target domain feature representations are aligned in the feature space of the source-trained segmentation model. 
\\\textbf{Shared style GAN}:
Most approaches that use a shared style GAN rely on the original GAN principle presented by Goodfellow et al.~\cite{Goodfellow2014}. 
These approaches usually have four elements: A shared encoder $\encoder$ that generates a shared feature space; a segmentation classifier $\classifier$ that computes the semantic segmentation from the feature map produced by $\encoder$; a decoder $\decoder$ that reconstructs the input image (mostly trained by $L1$ loss); And finally a discriminator $\discriminator$  that tries to classify the image output of G into either being "fake", or "real".

The approach presented in LSD \cite{sankaranarayanan2018learning} 
has all the architectural elements described.
The discriminator $\discriminator$ distinguishes between fake and real source domain images and fake and real target domain images. 
The decoder $\decoder$  generates fake target and source domain images by adding dropout noise to the feature embeddings generated by $\encoder$.
An adversarial loss is computed between real and fake images inside each domain and cross-domain.
This way, the encoder $\encoder$ is trained to output similar feature space embeddings for the source and target domain. 
$\classifier$ takes this domain-aligned feature map to compute the segmentation. 
CAA-Net presented by Ruan et al.\ \cite{ruan2019category} follows a similar direction as LSD \cite{sankaranarayanan2018learning}.

The architecture in PTP \cite{zhu2018penalizing} consists of the elements and again follows a similar principle as CAA-Net \cite{ruan2019category} and LSD \cite{sankaranarayanan2018learning}. 
Similar to LSD \cite{sankaranarayanan2018learning}, the final objective is to achieve similar feature embeddings for both domains by applying an adversarial loss. 
The biggest difference is that, e.g., on the source domain "real" would be the reconstruction of the source domain image and "fake" would be the reconstruction of the same image in a target domain style.

CLADA \cite{Hong2018} computes a transformation that is added to the pre-logit feature space of a segmentation network to transform the source domain features to the target domain. 
The classifier is trained on a target domain feature distribution given such a transformed source domain feature space. 
The conditional generator G takes in a noise channel and a low-level source domain feature map of encoder E. 
The two inputs are concatenated and passed through a \network{ResNet} architecture, computing the transformation. 
The discriminator distinguishes between transformed and non-transformed source domain feature maps. 

The network architecture of Lee et al.\ \cite{lee2020unsupervised} has three encoders that share the first convolution layers. 
One encoder extracts the content, and two other encoders extract the source and target domain style information, respectively.
The decoder computes the segmentation, and the other two decoders compute the image reconstruction in the source or target domain style.
The authors use a zero loss function, which minimizes the $L1$ norm so that the two encoders capture unique information that only exist in the source domain and target domain. According to the authors, the source encoder only learns the style-independent content features. 
\\\textbf{Maximum classifier discrepancy (MCD)}:
Next to the H-divergence, other distribution discrepancy metrics are used for distribution alignment in the feature space. 
Saito et al.\ \cite{Saito2018a} introduce an approach based on the maximum classifier discrepancy (MCD).
The MCD is computed by first training two classifiers for the source domain, mapping the same feature map into a segmentation.
In the second step,  the discrepancy of the probability output of these two classifiers is maximized for the target domain. 
The discrepancy between the class probabilities is computed using the $L1$ norm.
The segmentation loss is trained on the source domain in parallel to keep the source domain performance from degrading.
The result is two classifiers that agree with each other for samples with support from the source domain distribution and disagree with each other for samples that are not represented well in the source domain. 
The latter case characterizes most of the target domain samples.
In the third step, the feature extractor is then optimized to minimize the MCD for the target domain.
This causes those samples from the target domain far away from the source domain distribution to move closer to the source domain distribution (here, the support of the source domain is given, and the two classifiers agree).
The three steps are iterated, which results in an adversarial optimization.  \par
Lee et al.\cite{lee2019sliced} advance this work by introducing an improved way of computing the discrepancy between the classifier probability outputs.
The authors propose the sliced Wasserstein discrepancy, which considers the properties of the underlying geometry of probability space and thus improves upon the $L1$ norm used in \cite{Saito2018a}.
Further follow-up work is presented in Li et al.\cite{li2020bi} where two classifiers are trained on the source domain while also updating the feature generator. 
Then they maximize the classifier discrepancy on the target domain while ensuring the source domain classification stays the same. 
In the final stage, they train the feature generator to minimize the classifier discrepancy,  pushing the target domain data to the statistical support of the source domain.
\\\textbf{Other methods for distribution divergence minimization}:
MMD \cite{gao2021dual} uses the soft paste algorithm, combining two images by a weighted overlay (see Section \ref{sec:inputSpaceAlign}). 
A reference source domain image is pasted into a target and source domain image.
This is done based on a mask containing relevant classes in the reference image. 
The authors try to align the feature space representation of the source and target image in the region of the mask.
The minimization of the squared difference of the kernel-mean-embedding of the feature representations in the mask regions in the reproducing kernel Hilbert space introduces the alignment.
In addition to the alignment in the mask region, 
the authors apply a global alignment using the same method (MMD) without filtering with the reference image mask.
The general activation matching (GAM) \cite{huang2018domain} approach trains two networks, 
one for the target and one for the source domain. 
The authors apply an $L2$ minimization of the difference of the weights between the source and target domain networks and Jensen-Shannon divergence matching between the output of source and target domain.
The latter is optimized in an adversarial manner. 
Additionally, the feature maps of the target domain are scaled to the source domain mean and variance. 
These adaptation methods are applied in each layer of the network. 
PFR \cite{zhang2020towards} approach utilizes the $L2$ distance of the feature representations of source and target domain images. 
The style features and content features are computed using the method presented by Gatys et al.\ \cite{gatys2016image}.
The $L2$ distance is minimized at different feature levels. 
\subsubsection{Self-Supervised Learning}
Self-supervised learning is based on so-called pretext tasks that can be annotated automatically without human effort. 
The assumption is that the training on pretext tasks results in an encoder that produces features that are relevant to the actual task that should be solved, i.e., semantic segmentation.
Since self-supervised learning can be used for unsupervised feature learning, it is often applied for pre-training. More importantly, in our case, it is an essential method for unsupervised training on the unlabeled target domain, too.\\
\textbf{Augmentation and depth}:
SSL-UDA \cite{SelfSuperSemSeg} and SSDA \cite{xu2019self2} introduce a process to make use of self-supervised learning for an implicit alignment of the source and target domain. 
In addition to the main task of semantic segmentation, they employ the pretext tasks image rotation, image flipping, and location prediction of image crops.
These tasks are trained on the source and the target domain jointly. 
The idea is that by training the encoder to produce relevant features on the source and target domain, the distributions of the source and target domain will also align in the feature space.
The authors of SSL-UDA \cite{SelfSuperSemSeg} show that the centroids of both distributions get closer over the training epochs. 
However, the quality of such approaches is dependent on the pretext task. 
The approaches presented in GUDA \cite{guizilini2021geometric}, CTRL \cite{saha2021learning}, and CorDA \cite{wang2021domain},  show that depth prediction and ego-motion estimation are meaningful pretext tasks. 
GUDA \cite{guizilini2021geometric} makes use of recent advances in the domain of unsupervised depth estimation.
In addition to the semantic segmentation on the source, domain the authors train an unsupervised depth estimation on the target domain that implicitly predicts the ego-motion. 
Since the prediction of pixel-wise depth maps requires similar features as semantic segmentation, utilizing depth for the pretext task trains the encoder to extract relevant features even on the target domain. 
CorDA \cite{wang2021domain} and CTRL \cite{saha2021learning}, in contrast to GUDA \cite{guizilini2021geometric}, do not train the unsupervised monocular depth estimation as a pretext task.
Instead, the authors assume fixed depth labels, which unsupervised monocular depth estimation approaches can compute, too.
In Wang et al.\ \cite{wang2021domain}, another difference can be found in how depth estimation is incorporated via spatial attention into semantic segmentation.
For further details of how approaches incorporate depth and ego-motion pre-text tasks for self-supervised feature learning, refer to Section \ref{sec:outputSpaceAlign}.
\\\textbf{Contrastive learning}:
\label{subsec:contrastivelearning}
Unlike the implicit alignment of the feature distributions, self-supervised approaches are also used for direct alignment. 
DACL \cite{shim2021learning}, SPCL \cite{xie2021spcl}, UCDA \cite{zhang2022unsupervised}, PWCL \cite{liu2021domain}, RCCR \cite{zhou2021domain}, and CLST \cite{marsden2021contrastive} make use of contrastive self-supervised learning. \par
Contrastive self-supervised methods are based on so-called positive and negative pairs. 
In general, such methods aim to make the feature representation of positive pairs more similar and those of negative pairs more different. 
It depends on the task and method how positive and negative pairs are constructed. 
A positive pair are, e.g., two instances of the same class or two augmented versions of the same object. 
In the case of domain adaptation, the construction of positive and negative pairs is not trivial because no labels are available in the target domain. \par
The approach presented by Shim et al.\ \cite{shim2021learning} uses a CycleGAN-generated style transfer from the source to the target domain. 
The resulting pseudo-target domain images with ground truth labels yield pixelwise class information. 
The contrastive loss can be computed based on the so-constructed positive and negative pairs. \par
The authors of CLST \cite{marsden2021contrastive} follow a different approach. 
The idea is to construct high-quality pseudo labels for the target domain.
Given such pseudo labels, one can construct positive and negative pairs across the source and target domain.
The positive and negative pairs are constructed between the source domain class centroids and the target domain class centroids for each target domain image.
The feature representations of the target domain are clustered towards their respective source domain centroids and moved away from the wrong source domain class centroids.
The approach of SPCL \cite{xie2021spcl} is similar to CLST \cite{marsden2021contrastive}. 
The authors compute the average feature representations of each class on the source domain and update them in a moving average way. 
The contrastive loss is computed from the feature representation of each pixel to the centroids. 
In the target domain, the assignment is done by the pseudo-labels of the network. \par
RCCR \cite{zhou2021domain} combines contrastive learning with knowledge distillation and introduces both a teacher and a student for the projection head. 
Differently from other works, the projection head consists of convolution layers. 
The positive and negative pairs are constructed by the student and teacher network based on a source-target mixed image and a regular target image. 
As the only UDA approach, RCCR utilizes a memory bank to include negative samples from previous batches to increase the variety of the negative pairs and thereby improve the discriminability of the learned representations. 
UCDA \cite{zhang2022unsupervised} follows SimCLR \cite{chen2020simple}. 
It adds two MLP layers to transform the feature representations into a 128-dimensional vector representation. 
Class prototypes are computed per batch. 
Each feature vector contributes to each class prototype according to the softmax probability of the teacher network. 
Then anchor features are chosen within the same domain, and the contrastive loss is computed. 
Additionally, they choose anchor features in the source domain and assign them to the corresponding target domain centroid. \par
PWCL \cite{liu2021domain} determines positive and negative pairs between source and target domain image patches. 
The use of image patches is a distinct feature and is done through multi-level spatial pyramid matching.
Their contrastive approach is close to the idea of MoCo\cite{chen2020simple} and utilizes the cosine similarity.\par 
SCDA \cite{li2021semantic} differs from the previously described contrastive approaches because it does not create positive and negative pairs based on concrete feature representation, but rather operates on class distributions.  
The authors estimate the distributions of each class in the feature space based on source domain statistics using the mean and covariance.  
It is computationally infeasible to compute the contrastive loss for multiple positive and negative pairs. 
To resolve this limitation Li et al.\ derive a loss that directly utilizes the gaussian distributions of the positive and negative classes.
\\\textbf{Semantic clustering}:
Apart from the implicit adaptation through self-supervised learning and the construction of semantic pairs in the source and target domain, 
one can identify a third class of self-supervised domain adaptation approaches. 
Semantic self-supervised approaches as presented in DANCE \cite{saito2020universal} 
CAM \cite{wang2021separable}, 
CFContra \cite{tang2021unsupervised},
SCDA \cite{li2021semantic},
BAPA-Net \cite{liu2021bapa},
SWLS\cite{Dong2019a} %
, and SSS+ST \cite{niemeijer2021combining} 
which all aim to cluster the pre-logit feature space towards so-called class prototypes directly.
These class prototypes are vectors that represent the pre-logit feature representations of their respective class.

By advancing the method proposed by DANCE \cite{saito2020universal} Niemeijer et al.\ with SSS+ST \cite{niemeijer2021combining} present an approach for semantic self-supervised learning for semantic segmentation.
The class prototypes are computed as the moving average of source domain feature representations of the respective class during the training.

The authors of CFContra \cite{tang2021unsupervised} compute the average feature representations on the source and on the target domain. 
The target domain centroid is computed by assigning pseudo labels based on the distance of a feature representation to the source domain class centroids. 
Based on that, the two closest centroids are computed. 
The authors compute the contrastive loss between each combination of source domain features, target domain features, source domain centroids, and target domain centroids.

The authors of CAM \cite{wang2021separable} apply prototype clustering on both source and target domains. 
For each class, a single target domain feature representation is selected to serve as the prototype for the class.
This prototype feature is computed by determining the feature representation that has the 
maximum cosine similarity to all the other feature representations of the same class.
The similarity matrix and the entropy minimization are computed similarly to SSS+ST \cite{niemeijer2021combining}.
Distinct from this paper, the authors propose a contrastive clustering loss. 
This loss takes  normalized first-order statistics (mean representation) of each class cluster from the source and target domain and uses the euclidean distance as a distance metric for the clustering of the mean representations. 

The authors of OCE \cite{toldo2021unsupervised} apply feature clustering in the source and target domain, aiming to group feature vectors of the same class together and those of different classes away from each other.
Notably, OCE differs in the computation of the cluster centroids and the distance metric compared to the previous approaches.
The class centroids are computed based on the current batch, both on the source and target domain.
The distance metric that is used to define the similarity is the $L1$ norm.
During optimization, the $L1$ norm between the current feature representation is minimized to centroids of the predicted class and maximized to centroids of the other classes. 
Additionally, they introduce an orthogonality requirement meaning that feature vectors of different classes are forced to be orthogonal in the feature space.
The orthogonality requirement is based on the cosine similarity between the current feature representation and the class centroids. 

The method introduced in MCSSF \cite{chung2021maximizing} is also based on clustering. 
The authors introduce a dictionary containing the correctly classified feature representations in the source domain is defined.
The target domain feature representations of the current batch are stored in a dictionary, also. 
A cosine similarity matrix is computed between the target and source domain features of the same class.
Elements of this matrix representing low similarities are eliminated by thresholding and the cosine similarity of the remaining elements is maximized. Therefore, this approach does not optimize the feature representations of different classes to be dissimilar. 
This is also the case for LSR \cite{barbato2021latent} and BAPA-Net \cite{liu2021bapa}.

Similarly, the authors of LSR \cite{barbato2021latent} apply non-contrastive clustering to prototypes.  
The prototypes are computed for source and target domain and updated via a moving average.
The authors minimize the $L2$ norm of each feature representation to its corresponding class prototype. 
The correspondence to a class is determined based on the prediction probability.
Additionally, they enforce perpendicularity between prototypes of different classes (as in OCE \cite{toldo2021unsupervised}) and the norm of the target and source domain features to be the same.
They assume, according to recent research by Xu et al.\ \cite{Xu2019ICCV}, that target domain feature vectors have a smaller norm. Enforcing the norm to be the same in the target and source domains introduces domain alignment.  

SIM \cite{wang2020differential} is also based on non-contrastive clustering but distinct from the previous approaches the clustering is done differently for stuff classes like road or sky and thing or instance classes like car or pedestrian.
For the stuff classes, the authors compute multiple average feature representations per class by averaging the feature representations. 
For a given target domain centroid, the $L1$ norm is minimized towards the closest source domain centroid of the predicted class. 
For a given target domain instance centroid, the $L1$ norm is minimized towards the closest source domain instance centroid of the predicted class. 

Li et al.\ \cite{liu2021bapa} (BAPA-Net) assume that near-boundary pixels are hard to classify and propose a special handling of the boundary regions, different from the previous approaches. 
The authors employ the CutMix \cite{yun2019cutmix} operator to paste source pixels and labels to the target domain, artificially creating more boundary pixels that are assigned a higher weight. 
They employ a prototype clustering algorithm between the source and mixed target domain images.
The prototypes of the mixed target domain in the current batch are computed by assigning the feature vectors to the predicted class and filtering out those feature vectors for the centroid computation that are too close to a boundary. 
The class-wise centroids of the mixed images are optimized to minimize the $L1$ norm to the closest source centroid of the same class.\par
As we have seen, the above clustering approaches often use the classification of feature representations to determine to which centroid the current (target domain) feature representation should be clustered. 
Hence a good classification is necessary. 
Based on this, CaCo \cite{huang2021category} shows that existing domain adaptation methods can profit from an additional feature space clustering, given that they provide a good classifier to determine the clustering target centroid.

\subsection{Output Space Domain Adaptation}
\label{sec:outputSpaceAlign}
\begin{table}
    \centering
    \setlength{\tabcolsep}{.5em}
    \begin{tabular}{C{1.6cm}C{1.8cm}C{4.2cm}}
        \toprule
        \multicolumn{2}{c}{\textbf{Technique}} & \textbf{Approach} \\
        \midrule
        \multirow{17}{*}{\textbf{Self-Training}} & Global Thresholding &  
        \cite{musto2020semantically},
        \cite{yang2021context},
        \cite{li2019bidirectional},
        \cite{zhou2020affinity}, 
        \cite{lee2020unsupervised},  
         \cite{cheng2021dual},  \cite{xu2021self},
         \cite{kim2020learning}, 
        \cite{tranheden2021dacs},
        \cite{wang2020classes},
        \cite{saha2021learning}, 
        \cite{liu2021domain}, 
        \cite{stan2020unsupervised}, 
        \cite{chen2019domain},  \cite{xie2022sepico}\\ \cmidrule{2-3}
        & Adaptive Training &  
        \cite{Iqbal2020},\cite{Zou2018}, \cite{Dong2019a}, \cite{li2020variational}, 
        \cite{song2020learning},\cite{wang2020differential}, \cite{lee2020unsupervised}, \cite{chung2021maximizing},\cite{dong2021and},  \cite{huang2021cross},\cite{ma2021coarse}, 
        \cite{araslanov2021self},
        \cite{zhang2021prototypical},
        \cite{yu2021dast},
        \cite{dong2020cscl},
        \cite{xie2021spcl},
        \cite{wang2021separable},
        \cite{li2021semantic},
         \cite{zou2019confidence}, \cite{mei2020instance},   \cite{li2020content}, \cite{zheng2019unsupervised},   \cite{chung2021exploiting} \\\cmidrule{2-3}
        &Image-Level Self-Training&\cite{paul2020domain}, \cite{Iqbal2020}, \cite{Lian2019}, \cite{wang2019class}, \cite{lv2020cross}, \cite{huang2020contextual}, \cite{zhang2017a}, \cite{liu2021adversarial}, \cite{guan2021scale}\\ \cmidrule{2-3}
        &Entropy-Based&\cite{tang2021unsupervised}, \cite{huang2021cross}, \cite{zhou2021context}, \cite{huang2021mlan}, \cite{chen2021unsupervised},  \cite{niemeijer2021combining}, \cite{zou2019confidence}, \cite{li2020content}, \cite{Subhani2020}, \cite{wang2021cross},  \cite{wang2021uncertainty}, \cite{truong2021bimal},    \cite{pan2020unsupervised}, \cite{yang2020adversarial}\\ \cmidrule{2-3}
        &Ensemble Learning& \cite{Iqbal2020}, \cite{li2020unsupervised}, \cite{yang2020fda}, \cite{li2020bi}, \cite{barbato2021latent}, \cite{chen2019domain},   \cite{zheng2021rectifying}, \cite{xu2022unsupervised}\\ \cmidrule{2-3}
        &Discriminator Confidence&\cite{zheng2019deep}, \cite{michieli2020adversarial}, \cite{spadotto2020unsupervised}, \cite{shen2019regularizing}\\ \cmidrule{2-3}
        &Others&\cite{lu2022bidirectional}, \cite{wang2021domain}, \cite{huang2021mlan}\\
        \midrule
        \multirow{15}{*}{\textbf{Adversarial}} & Basic&
        \cite{li2020variational}, 
        \cite{musto2020semantically}, 
        \cite{chang2019all}, 
        \cite{song2020learning},  
        \cite{li2019bidirectional}, 
        \cite{kim2020cross}, 
        \cite{wang2020differential}, \cite{yang2020label}, 
        \cite{lee2020unsupervised},
        \cite{cheng2021dual},
        \cite{xu2021self}, 
        \cite{gao2021addressing},
        \cite{li2020unsupervised}, 
        \cite{kim2020learning},
        \cite{yu2021dast}, 
        \cite{dong2020cscl}, 
         \cite{mei2020instance}, 
         \cite{guan2021scale}, 
         \cite{wang2021uncertainty},
         \cite{yang2020adversarial},
         \cite{zheng2021rectifying},  \cite{xu2022unsupervised},
        \cite{michieli2020adversarial},
         \cite{shen2019regularizing},
         \cite{tsai2018learning},\cite{liu2021source}, \cite{zhang2020joint},  \cite{vu2019a},  \cite{yan2021pixel}  \\ \cmidrule{2-3}
        & Depth & 
        \cite{Dong2019a}, \cite{chen2019learning}, 
        \cite{zheng2019deep}, \cite{saha2021learning}, 
        \cite{zhang2020towards},
        \cite{vu2019a} \\ \cmidrule{2-3}
        & Entropy-Based &\cite{chiou2022beyond},
        \cite{chen2021unsupervised},
        \cite{wang2021cross}, \cite{wang2021uncertainty},
        \cite{pan2020unsupervised},
        \cite{cicek2020spatial}, \cite{tang2020towards}, \cite{vu2019advent}    \\ \cmidrule{2-3}
        & Modified Input & 
        \cite{zhou2020affinity}, \cite{tsai2019domain} \\ \cmidrule{2-3}
        & Bi-Classifier &\cite{Saito2018a},\cite{luo2019taking}, \cite{li2020bi},  \cite{lee2019sliced}\\ \cmidrule{2-3}
        & Intra-Domain & \cite{wang2021cross}, \cite{pan2020unsupervised}, \cite{yan2021pixel}\\ \cmidrule{2-3}
        & Multi-Level &
        \cite{shan2020semantic}, \cite{huang2021mlan}\\ \cmidrule{2-3}
        & Multi Discriminator & \cite{lin2019adapting}, \cite{yang2021context}, \cite{spadotto2020unsupervised}, \cite{chen2018road}\\ \cmidrule{2-3}
        & Class-Wise & \cite{du2019ssf}, \cite{wang2020classes}, \cite{luo2021away}, 
        \cite{wang2019class},
        \cite{ruan2019category},  \cite{liu2021adversarial}\\ 
        \midrule
        \multirow{5}{*}{\textbf{Consistency}} & Augmentations&\cite{zhang2021spectral}, 
        \cite{ma2021coarse}, 
        \cite{melas2021pixmatch},
        \cite{tranheden2021dacs},   \cite{guan2021scale}, \cite{Subhani2020}\\ \cmidrule{2-3}
        & Style Transfer & \cite{Chen2019crdoco},
        \cite{toldo2020unsupervised},
        \cite{gao2021addressing}, \cite{li2020unsupervised}, \cite{li2020simplified}, \cite{yang2020adversarial}\\ \cmidrule{2-3}
        & Knowledge Distillation& \cite{choi2019self},
        \cite{xu2021self},
        \cite{lu2022bidirectional}, 
        \cite{zhou2020uncertainty}, \cite{araslanov2021self}, \cite{Gao2021DSP}, \cite{zhou2021context}, \cite{zhang2021prototypical}, \cite{zheng2021rectifying}, \cite{zhang2021multiple}, \cite{xu2019self},\\ \cmidrule{2-3}
        & Others & \cite{kim2020cross}, \cite{zhou2020affinity}, \cite{zhang2020transferring}, \cite{lv2020cross},  \cite{chung2021exploiting}\\
        \midrule
        \textbf{Depth-Based} &&\cite{chen2019learning}, \cite{lee2018spigan}, \cite{cardace2022plugging}, \cite{saha2021learning}, \cite{guizilini2021geometric}, \cite{Vu2019dada}\\
        \midrule
        \textbf{Contrastive Learning}&&\cite{zhang2022unsupervised}, \cite{zhou2021domain}, \cite{marsden2021contrastive}, \cite{liu2021domain}, \cite{kang2020pixel}, \cite{li2021semantic}\\
        \midrule
        \textbf{Others}&&\cite{kim2020cross}, \cite{liu2021bapa}, \cite{chen2021unsupervised}\\
        \bottomrule
    \end{tabular}
    
    \caption{Adaptation methods in the \textbf{output space}. The papers are clustered and sub-clustered according to similar methodology.}
    \label{tab:outputspace} 
\end{table}

Output space adaptation methods can be formally defined by the distinctive property that the pixel-wise logits or softmax probability outputs $y_{i,s} = P(s|i,\mathbf{x})$ of the network are utilized for the adaptation.\par
Output space adaptation methods can be subdivided into different subcategories, which are shown in Table \ref{tab:outputspace}. 
The two most popular and commonly employed output space adaptation methods are self-training and adversarial learning, while several other methods have also been utilized for adaptation, such as entropy-based methods and consistency or contrastive learning.

\subsubsection{Self-Training}
\begin{figure}[t]
    \centering
    \includegraphics[width=1.0\linewidth]{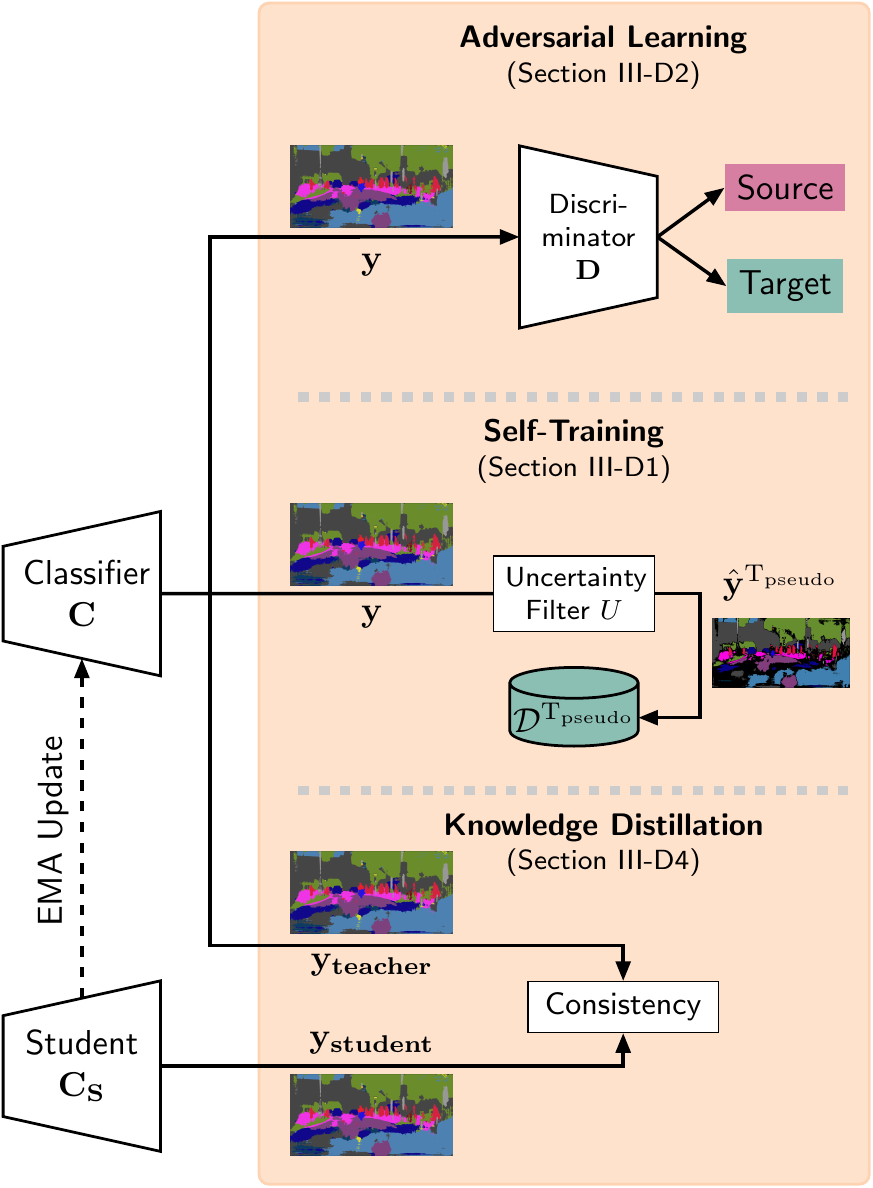}
    \caption{\textbf{Overview of output space adaptation methods}: Simplified and exemplary schematic visualization. Deviations of certain approaches from this basic scheme are possible.}
    \label{fig:output_adaptation_methods}
\end{figure}

The general idea is to retrain the network on labels that are generated by itself (see Figure \ref{fig:output_adaptation_methods}). 
In the unsupervised domain adaptation setting, the network $\mathbf{F}(\cdot)$ is trained in the source domain $\src$ in a supervised manner as the first step. 
In the second step, the trained network generates the raw predictions by running inference in the target domain $\tgt$ delivering $\mathbf{y}=\mathbf{F}(x^{\mathrm{T}}; \VECG{\theta})$. 
Because of the domain shift, $\mathbf{y}$ is noisy and contains wrong labels, so a direct utilization as pseudo-ground truth is not optimal. Instead, methods are required to discriminate between reliable and non-reliable predictions.\par
For this reason, the distinctive operation of self-training methods is mostly the filter operation $\overline{\VEC{y}}^{\mathrm{T}_\mathrm{pseudo}} = \VEC{U}(\mathbf{y})$,  which removes predictions with low confidence. A small sub-taxonomy of self-training methods is given in the second column of Table ~\ref{tab:outputspace}. These methods are particularly often used in hybrid approaches and only rarely as stand-alone adaptation methods (cf. Section~\ref{subsubsec:Hybrid_Domain_Alignment}).\par 
Some methodological characteristics are shared among self-training methods. 
At the beginning of UDA research, one of these was a so-called warm-up step \cite{mei2020instance, marsden2021contrastive, zhang2019category}, where a different adaptation method is employed to obtain an initial adaptation of the network and a better start performance for the pseudo-labels. 
However, with the rise of hybrid methods, dedicated warm-up steps became obsolete. 
Also, often multiple iterations or stages of self-training are performed to iteratively increase the performance of the pseudo-labels \cite{Zou2018, zou2019confidence, mei2020instance, zhang2021prototypical}. \par
\noindent\textbf{Global thresholding}: Global maximum softmax thresholding is the simplest method employed by several approaches (see Table ~\ref{tab:outputspace}). 
These approaches take the softmax probability distribution $P(s|i,\VEC{x})$ as a pixel-wise confidence estimation of the network and filter out every pixel whose maximum softmax probability is below a certain class-independent threshold, often $0.9$ or $0.95$.\\ 
\textbf{Adaptive training}: Several other researchers propose extended softmax thresholding mechanisms belonging to the adaptive softmax thresholding category. 
One important motivation for this group of methods is not to treat all classes in the same way but employ different adaptive thresholds to the classes since not all classes have similar output probability distributions due to the domain shift.
Class-balanced self-training (CBST)~\cite{Zou2018} introduced these class-wise thresholds as one of the first works. 
It combines output normalization and class-specific quantile-guided thresholding.
The best $p_i$ percent of the pixels per class $i$ are chosen as a pseudo-label, and $p_i$ is increased over the self-training iterations. 
Several self-training methods are directly based on CBST, e.g., CVRN~\cite{huang2021cross}, MRNet~\cite{zheng2019unsupervised}, MLSL~\cite{Iqbal2020}, and CSCL~\cite{dong2020cscl}. 
Other approaches only use the class-wise quantile-guided thresholding as a self-training method, where the top $p \%$ pixels of each class is selected, e.g., CCM~\cite{li2020content}, CSCL~\cite{dong2020cscl}, APL~\cite{song2020learning}, PA+CCR~\cite{ma2021coarse}, and SCDA~\cite{li2021semantic}. 
Additionally, regularization methods for CBST were proposed. 
The cross-view regularized network (CVRN)~\cite{huang2021cross} extends the CBST method by an inter-task and inter-style regularized multi-task self-training, which enforces consistency between instance and semantic segmentation and two different styles.
Confidence-regularized self-training (CRST)~\cite{zou2019confidence} introduces label- and model-regularized self-training. 
In addition to CBST three different regularization techniques are utilized to penalize over-confident labels to output a more uniform probability distribution. 
All three regularization methods $L2$, entropy regularization, and KL divergence regularization, achieve similar performance.\par
Stuff and instance matching (SIM)~\cite{wang2020differential} utilizes a partially adaptive class-wise thresholding by computing the class-wise median of the maximum softmax probabilities across the entire target dataset. The median is used as a threshold if being smaller than a fixed threshold of $0.9$.\par
Contrary to previous works, ProDA~\cite{zhang2021prototypical} proposes not iteratively to change the hard pseudo-labels but to keep them fixed and weigh them instead.
The authors show that both stage-wise self-training and the parallel update of network and pseudo-labels (trivial solution) lead to sub-optimal results. 
This is notably different from popular self-training strategies such as CBST, where a self-paced scheme with an increasing amount of labels is used.
The ProDA weights are computed based on prototypical features (see Section \ref{sec:featSpaceAlign}), and ProDA outperformed previous self-training approaches by a large margin. 
Note that also additional methods such as knowledge distillation (see Section \ref{sec:consistency_output}) contribute to the performance increase.\par
The predictions from past iterations can be complementary utilized to refine the pseudo-labels. 
IAST~\cite{mei2020instance} extends the quantile-guided class-wise thresholding to instance-adaptive thresholding, where the applied threshold is an exponential moving average of the quantile-based image-specific threshold and the history of thresholds from previous samples. 
In this way, information from both other instances of the dataset and the current instance are considered for thresholding. 
Similar to that, SSAC~\cite{araslanov2021self} utilizes an exponential moving average of class-specific prior probabilities (certain pixel belonging to a certain class) and introduces two hyperparameters so that the threshold for often-occurring classes, such as road, etc. remains unaffected but decreases for rare classes such as traffic light, etc.  
CLST~\cite{marsden2021contrastive} uses a temporal ensemble by storing predictions from past iterations in a weighted manner. 
The pseudo-labels are obtained by a majority vote of the stored predictions, and a class weighting based on the class frequencies is also employed.\par
Next to the pixel-wise self-training methods, a smaller cluster of methods emerged, which can be seen as image-level self-training. 
It can be distinguished by its characteristic to employ self-training for patch- or image-level predictions. 
Often approaches of this cluster make use of multi-scale domain alignment to obtain both local and global alignment. A popular representative of this cluster is PyCDA~\cite{li2018pyramid}.\\
\textbf{Image-level self-training}: Next to the standard pixel-wise self-training with softmax thresholding, self-training on larger patches with patch-level pseudo-labels is conducted in a curriculum manner. 
The patch-level labels are obtained by average pooling on the pixel-level labels and thresholding. 
This method's highest level of abstraction is the prediction of the global label
distribution of the entire image.
CDA~\cite{zhang2017a} shares significant similarities with PyCDA since it also utilizes the prediction of global image class distribution and superpixel class distribution. 
In contrast to PyCDA, logistic regression, and a support vector machine are used for the corresponding tasks and no pixel-level pseudo-labels are utilized. 
The authors argue that estimating global class distributions is easier than pixel-wise pseudo-label prediction. Similarly, pivot interaction transfer (PIT) \cite{lv2020cross} utilizes multi-nomial logistic regression. 
It is trained on the source domain with the image-level class distribution to train multiple region expansion units. 
These units consist of a convolution, and an up-sampling operation and with different smoothing parameters in the aggregation layer afterward, the different units focus on different object sizes. 
This is combined with a knowledge transfer to the pixel-level source segmentation branch.\par
WDA \cite{paul2020domain} simplifies the training task further by training a network with a class-wise binary cross-entropy loss on the image-level to predict whether a class exists in a training sample. 
This is done by training class-wise classification networks with class-wise features, so each class has its classifier and feature representation.
CCDA \cite{wang2019class} developed a different method for the same aim. 
Instead of class-specific classifiers like WDA, this approach utilizes two discriminator branches to predict which classes are present in a particular patch using binary classification loss and an adversarial loss.
For the target domain, the existing classes of a patch are predicted using the pixel-wise thresholded pseudo-labels. This coarse patch-level is accompanied by a fine-grained pixel-level discriminator, reaching a multi-scale domain alignment.\\
\textbf{Entropy-based}: The output probability distribution's entropy is a popular uncertainty estimation tool.
It is utilized for several UDA approaches since it provides more information about the network output than the maximum softmax prediction. Just like maximum softmax-based self-training, entropy-based self-training can be easily applied complementary to other UDA methods \cite{saporta2020esl, wang2021cross} and often used for adversarial learning (see Section \ref{sec:adversarial_output}). One of the first works using entropy information was ADVENT \cite{vu2019advent}. 
This approach employs entropy minimization, which can be seen as entropy-based self-training without a thresholding mechanism and, thus, without one-hot encoded labels. 
FDA \cite{yang2020fda} followed a similar idea and included a Charbonnier penalty weighting \cite{bruhn2005towards} for the entropy minimization, which assigns a higher loss to high entropy regions. 
Next, the entropy can replace the maximum softmax value as the uncertainty measure and be utilized analogously with thresholding. Several works are employing this method with different thresholding techniques: Niemeijer et al.\ \cite{niemeijer2021combining} use a global quantile-based threshold, ESL \cite{saporta2020esl} and LSE+FL \cite{Subhani2020} both apply a class-based quantile-like threshold for the entropy taking the most confident $p$\% pixels as pseudo-labels. CRA \cite{wang2021cross} employs a manually chosen threshold which is halved for rare classes.\par 
In contrast, other works avoid the usage of thresholds. SEDA \cite{chen2021unsupervised} does not utilize thresholding but takes the inverse of the entropy values as weights for the unfiltered target pseudo-labels. UncerDA \cite{wang2021uncertainty} fits a Gaussian mixture model to obtain a class-wise positive and negative distribution which is employed to assess if predictions are used as pseudo-labels.\par
Two approaches aim to utilize the entropy by developing new loss functions next to the entropy minimization described previously. MSL~\cite{chen2019domain} argues that the entropy minimization is not optimal for rare classes due to a higher gradient for higher probabilities.
Consequently, the maximum squares loss is introduced, where the loss is the square of the predicted probabilities, which leads to a linearly increasing gradient and makes both easy and hard classes better transferable since the gradient grows not exponentially for easy classes. 
PTP~\cite{zhu2018penalizing} also modifies the standard cross-entropy loss so that the
loss becomes symmetric, and very high predicted probabilities are penalized for discouraging the network from overfitting.
BiMaL \cite{truong2021bimal} introduces a new loss function that is a generalized form of the entropy minimization of ADVENT \cite{vu2019advent}. 
The bijective maximum likelihood loss uses a sequence of bijective mapping functions to map the segmentation output to the latent space, where the log-likelihood loss is computed. 
This approach should better capture image-level characteristics than entropy minimization since not every pixel is treated individually.\\
\textbf{Ensemble learning}: Ensemble learning is a commonly applied method for various applications, e.g., uncertainty quantification \cite{abdar2021review}. 
Ensemble learning refers to a group of methods where two or more predictions of different DNNs or different (segmentation) heads are included to obtain the final prediction.
SAC~\cite{li2020unsupervised} proposes a classic ensemble learning method by training two segmentation networks with correspondingly translated source and target images.
The predictions on the target image are averaged and filtered by a softmax threshold to obtain the pseudo-labels. 
The MRNet~\cite{zheng2021rectifying} method is different by only using a second classifier head which takes features from a different layer for weighted prediction summation.
EPS-UDA~\cite{xu2022unsupervised} employs three semantic segmentation heads with a shared encoder but differs from other ensemble learning methods in two ways. 
First, each head is trained with the outputs from the two other branches. 
The valid pseudo-labels are only assigned when both outputs agree, making this the only method where this hard constraint is employed instead of, e.g., averaging.
Second, the agreement between the heads is measured using KL divergence and then used as a weighting factor. \par
For UDA approaches, ensemble learning is often closely connected to the group of multi-inference self-training methods, where different versions of one image are fed into the network, and multiple predictions are combined to get a more robust final prediction. These methods can be seen as an extension of the standard ensemble learning methods with Monte Carlo uncertainty estimation. 
In FDA \cite{yang2020fda}, three independently trained networks are utilized, and each network receives differently style-transferred input images. 
The predictions of all three networks are averaged, and an argmax operation obtains the prediction.
SIT \cite{chiou2022beyond}] follows a similar scheme. 
However, instead, it uses a stochastic style transfer (see Section \ref{subsec:ganbasedmethods}) to vary the style of the translated images and trains a triplet of networks with a large style variety of the translated input images.
One of these networks is trained with ten different style transfers per image, which can be seen as a Monte-Carlo-like uncertainty reduction by averaging over the predictions. 
The outputs are averaged across the three networks, and class-balanced self-training is applied to further filter the pseudo-labels.
DPL~\cite{cheng2021dual} employs a similar method having a target and a source network. 
With both target images and source style translated target images, for the source network, two different predictions are obtained, averaged like in FDA, and the known maximum softmax thresholding is applied.\par
A crucial design choice for ensemble learning architectures is the number of different networks or network heads. 
STAR \cite{lu2020stochastic} introduced an alternative to a fixed number of classifiers. 
Instead, the authors propose to model a distribution of classifiers as a multivariate Gaussian and randomly sample the model weights from this learned distribution.
During training, the weights themselves are not optimized, but the distribution parameters from which the weights are then sampled. 
In practice, the method employs two different classifiers, which should lead to similar results as training with an infinite number of classifiers. It can therefore be seen as a stochastic variant of CLAN \cite{luo2019taking}.\par
Most methods do not explicitly estimate or utilize the source and target domain shift.
However, only a few works aim to utilize the domain shift estimation for self-training. 
The authors of CorDA~\cite{wang2021domain} argue that depth prediction can be used as a proxy for the actual domain shift estimation. 
The target images are processed through a source and a target depth prediction network, and the difference between these predictions is computed. 
If the depth prediction difference is high, the according pixel gets a low weight in the self-training loss assigned, and vice versa.\\
\textbf{Discriminator confidence}: Using discriminator confidence is a less popular self-training method and is only possible adjacent to an adversarial learning framework. 
The underlying assumption of these works is that those pixels where the discriminator has high confidence are also good pseudo-labels.
AL+ST \cite{spadotto2020unsupervised} exploits a pixel-wise domain discriminator to use these outputs as a confidence estimate for the target predictions. The thresholding of the discriminator confidence values is the same as the class-wise quantile-based softmax thresholding. 
MADA \cite{shen2019regularizing} applies a very similar principle for its two pixel-wise discriminator confidence maps but combines it with the softmax probabilities, so only pixels where both confidence estimates are above the threshold get qualified for the pseudo-label.\par
DMLC \cite{guo2021metacorrection} is significantly distinct from the other self-training methods since it utilizes meta-learning for UDA. 
It aims to correct wrong pseudo-labels with a noise transition matrix (NTM). 
This matrix contains class transition probabilities and is jointly optimized with the segmentation network in a three-step scheme that includes a meta-optimization step. 
In this step, the source data with a domain predictor is used to serve as a so-called metadata set, estimate the network's generalization, and update the NTM accordingly.
 
\subsubsection{Adversarial Output Adaptation}
\label{sec:adversarial_output}
Adversarial output adaptation is one of the most often applied methods in UDA research.\\
\textbf{Basic}: 
It introduces the basic idea of attaching a discriminator after the segmentation probability output and trains it to predict if a certain prediction sample originates from a source or target input on the image-level.
A simple visualization of this idea is shown in Figure \ref{fig:output_adaptation_methods} (top). 
By backpropagating this adversarial loss, the segmentation networks should output similar segmentation distributions for both the source and target domain since the segmentation network will try to fool the discriminator with similar outputs.
The target predictions will become more similar to the source, and the network gets adapted to the target domain.
This principle also works for two different discriminators as proposed by AdaptSegNet. 
One receives the standard high-level softmax output, and the second discriminator ensures a low-level adaptation by getting segmentation predictions only based on lower-level features. 
Several other works employ adversarial output adaptation as proposed by AdaptSegNet in addition to other methods \cite{lin2019adapting, zhang2020joint, kim2020learning, wang2020differential, yang2020label, chang2019all}. 
Some approaches employ a reduced adversarial learning method by leaving out the utilization of low-level features \cite{li2019bidirectional, dong2020cscl}. 
A minor change to the original adversarial learning method is to replace the source input with a style-transferred source image whose style is more similar to the target domain \cite{musto2020semantically, song2020learning, xu2021self}. 
Similarly, DPL \cite{cheng2021dual} uses two adversarial losses. 
One loss for translated source and real target images and the second for translated target and real source images. 
Another minor change to the original scheme is proposed in UncerDA \cite{wang2021uncertainty}. 
A sophisticated sampling strategy for the source images in the adversarial adaptation process is proposed to show rare or hard classes according to their entropy uncertainty. Classes with high uncertainty are shown more often and vice versa.\par
However, there are a lot of extensions and improvements of this original adversarial method which can be clustered as shown in Table \ref{tab:outputspace}. 
A straightforward extension is multi-level adversarial learning, which distinctive characteristic of AdaptSegNet is the utilization of features in multiple different network layers for the adversarial adaptation.
Therefore, these works closely correlate to approaches operating in the feature space.
SASP \cite{shan2020semantic} applies two types of adversarial adaptation. 
Next to the known output adversarial learning, it concatenates multiple latent layers before sending the concatenated result to a classification layer and applying the adversarial loss. 
The authors reason that also the earlier layers receive a strong learning signal from this multi-layer fusion. 
MLAN \cite{huang2021mlan} also proposes multi-level adversarial adaptation but has a different argumentation and approach. 
In global alignment, no local distributions can be adapted, so MLAN introduces region-level adversarial learning where relations between small patches are utilized to reach fine-grained region-level adaptation. 
In addition to the connection between local and global alignment, consistency maps on multiple levels are calculated. \\
\textbf{Multi-discriminator}: AdaptSegNet employs two discriminators for different levels but with the same objective. 
A straightforward extension is two discriminators with different objectives, as proposed by MDD \cite{spadotto2020unsupervised}. 
Here a second discriminator is trained to distinguish between the source predictions and the source ground truth. CADA \cite{yang2021context} trains three different discriminators. 
They all align the source and target domain. 
However, since two discriminators receive their prediction based on the output of a feature attention mechanism, CADA can also be seen in the group of multi-discriminator approaches.
Next to the standard discriminator, MAGD \cite{lin2019adapting} proposes a gated discriminator next to the standard discriminator that additionally takes foreground-background segmentation masks as the input since both areas have different adaptation difficulties.
The gated convolutions make it possible to utilize the masks without hard thresholding, and the additional input for the discriminator makes distribution alignment easier for fore- and background.\par
The basic adversarial output adaptation does not distinguish between classes, so a group of works addresses class-dependent adversarial learning. 
CCDA \cite{wang2019class} introduces two extensions to the adversarial adaptation from AdaptSegNet \cite{tsai2018learning}. First, it has one coarse- and one fine-grained branch enforcing adaptation at different levels of granularity.\\ 
\textbf{Class-wise adversarial learning}: Second, a special focus lies on class-dependent adaptation, and class-conditional loss functions are used.
The adversarial part of the coarse branch is trained to distinguish the image-level class predictions of the two domains. The pixel-wise predictions of the fine-grained branch to discriminate between source and target domain are computed by a class-conditional binary cross-entropy loss. \par
SSF-DAN \cite{du2019ssf} computes class-wise semantic features by separate convolutional layers for each class and calculates the sum across all classes before computing the loss. 
SSF-DAN introduces a pixel-wise weighting based on the softmax prediction confidences to let the adversarial loss focus more on difficult classes. 
CGDA \cite{luo2021away} utilizes a so-called cross-domain grouping network that performs clustering of the segmentation output to better align the particular classes. 
The discriminator receives an input conditioned on the learned sub-clusters of the grouping network and, therefore, can better learn to distinguish classes. 
FADA \cite{wang2020classes} proposes a significantly different approach. Instead of only binary domain labels, the discriminator is trained to output the entire softmax probability prediction from the segmentation network. 
This additional knowledge is expected to enable a better class-adaptive alignment since the same network predicts class distribution and domain labels.
CLS \cite{liu2021adversarial} employs a similar method and proposes a shared classifier and discriminator for better conditional alignment.
The shared decoder has one more output class in addition to the predefined semantic classes to support the discriminator.
CAA-Net \cite{ruan2019category} follows a similar idea but applies class-wise masks to the prediction maps so that the discriminator predicts a domain label for each separate class map.\\
\textbf{Entropy-based}: Several approaches use entropy to replace the softmax predictions for adversarial training.
ADVENT \cite{vu2019advent} was one of the early works training the discriminator to distinguish between domains based on the entropy maps of source and target output probability maps. 
SCDS \cite{cicek2020spatial} slightly modifies this method by using random patches for adversarial training since they argue that the distribution of an entire image differs too much. Some works simply integrate ADVENT into their framework as an additional method \cite{chiou2022beyond, tang2020towards}. 
IntraDA \cite{pan2020unsupervised} employs the same method and uses entropy to split the target domain into hard and easy samples for an adversarial intra-domain adaptation.
SEDA \cite{chen2021unsupervised} extends this approach and introduces a weighting factor for the adversarial loss based on the entropy of the images so that hard samples with a high entropy can have a higher impact on the loss. CRA \cite{wang2021cross} generates binary labels for confident and non-confident regions based on entropy thresholding and trains a discriminator to distinguish between these regions within the target domain. 
In UncerDA \cite{wang2021uncertainty}, the entropy is used to resample the input of the adversarial learning according to the class-wise uncertainties on the target domain.\par
Next to the entropy, the basic softmax probability input for the discriminators can also be replaced with other inputs aiming for more meaningful alignment. ASANet \cite{zhou2020affinity} introduces affinity maps for adversarial adaptation. These maps are computed by also taking the predictions of adjacent pixels into account and therefore enforcing the discriminator to focus on the structural properties of the domains. 
As the only work for semantic segmentation, APODA \cite{yang2020adversarial} uses an adversarial attack to compute adversarial features and to attack both the discriminator and the classifier. It trains the attacked discriminator to output the same domain label for both the clean and the perturbed prediction. DRP \cite{tsai2019domain} combines adversarial learning and patch-wise clustering. It trains an additional network on the source domain to cluster output patches and uses adversarial learning for alignment with the source domain. MLAN \cite{huang2021mlan} is mostly similar to that but uses DBSCAN clustering \cite{ester1996density} instead of training a separate network for clustering.\par
Adversarial learning is usually employed between the source and target domain for aligning the two different domain distributions. However, a group of works aims to obtain intra-domain alignment with adversarial learning, which means alignment between different distributions within one of the domains. While this is technically no domain adaptation, this group of methods helps to learn target representations. IntraDA \cite{pan2020unsupervised} combines both inter- and intra-adaptation. 
First, the known entropy-based adversarial alignment is applied, and second, the image-level entropy is used to divide the target domain into easy and difficult samples for adversarial training. PixIntraDA \cite{yan2021pixel} extends IntraDA to the pixel-level since they argue that the image-level does not capture the local differences of the prediction. 
However, no entropy is involved there, but simple softmax thresholding is used to distinguish between easy and hard samples. CRA \cite{wang2021cross} can be seen as a combination of IntraDA and PixIntraDA since it combines entropy with pixel-level adversarial learning. It also classifies so-called trusted and untrusted regions of the target domain by their corresponding entropy and trains a discriminator in an adversarial manner to align both regions. 
This method should help to transfer the knowledge from the well-segmented parts of the target domain to the less-confident areas.\par
CSCL \cite{dong2020cscl} focuses on separating transferable knowledge as one of only a few works since the authors argue that global adversarial adaptation may cause misleading knowledge transfer.
For that purpose, a transferability quantizer and critic are introduced to distinguish between these different types of knowledge. 
The critic is necessary to inform the quantizer where the transferability estimates are not accurate. The critic is trained using a reward based on how good the segmentation predictions are.\par
Depth information as an additional modality is popular in UDA research and adversarial output adaptation several approaches make use of it. GGIO \cite{chen2019learning} is one of the simplest combinations of depth and segmentation information. It concatenates the two predictions and provides this as the input to the discriminator. DADA \cite{Vu2019dada} includes the depth modality in a more sophisticated way. It fuses the segmentation entropy with the depth prediction as the element-wise product and forwards the fused results to the discriminator. The discriminator is trained to distinguish between the source and the target domain. 
The authors argue that depth and geometric information are similar between the domains and, therefore, beneficial for better domain alignment. 
CTRL \cite{saha2021learning} follows a similar idea but on task-level. 
Depth prediction is incorporated here to improve the adaptation. A dedicated cross-task relation layer is employed, where the entropy of the semantic prediction, the semantics refinement head prediction and the depth prediction are concatenated and forwarded to the domain discriminator. The authors reason that an adversarial adaptation on both semantic and depth entropy maps makes the adaptation easier to difficult to transfer classes.\\
\textbf{Bi-classifier}: Maximum classifier discrepancy (MCD) \cite{Saito2018a}, SWD \cite{lee2019sliced}, and BCDM \cite{li2020bi} belong to a group of adversarial methods that are use two different classifiers instead of or in addition to a discriminator. These methods obtain an adversarial alignment in the feature space, so a detailed description is provided in Section \ref{sec:featSpaceAlign}. However, there is a mutual connection to the output space alignment because the output discrepancy between their two classifiers is iteratively maximized to construct the adversarial learning process.
The bi-classifier method proposed in CLAN \cite{luo2019taking} differs from the previous settings. 
First, an additional discriminator is employed after the two classifiers, receiving the summed output, and the cosine similarity of the classifiers weights the adversarial loss.
The cosine similarity is used as the discrepancy loss between the two classifiers. No alternating optimization with the same loss function as in the three other works is applied \cite{Saito2018a, lee2019sliced, li2020bi}.\par
The growing research area of knowledge distillation/transfer is combined with adversarial domain adaptation in the SE-GAN approach\cite{xu2021self}. A student-teacher network architecture with an exponential moving average update for the student replaces the standard generator to stabilize the adversarial training.  

\subsubsection{Contrastive Output Adaptation}
Contrastive learning is mainly applied directly in the feature space as described in Section \ref{sec:featSpaceAlign}, but some approaches exploit it for the output space. 
The basic principle here is the same: the network is trained to output similar representations for similar inputs or classes and vice versa.\par
The approaches PWCL \cite{liu2021domain}, CLST \cite{marsden2021contrastive}, SDCA \cite{li2021semantic}, RCCR  \cite{zhou2021domain}, and UCDA \cite{zhang2022unsupervised} all have in common that the contrastive adaptation operates in the feature space, and a detailed description is provided in \ref{subsec:contrastivelearning}. 
However, to compute positive and negative pairs, all access the output space to obtain the pseudo-labels, which directly correlates to output space alignment since reliable pseudo-labels are important for the adaptation process.\par
This also applies to PLCA \cite{kang2020pixel}but it is the only work that conducts multi-level contrastive learning on both the feature maps and the semantic predictions. 
 For the latter one in the output space, the authors choose a different metric to compute the similarity and their positive pairs between source and target prediction, namely the Kullback-Leibler divergence.

\subsubsection{Consistency Output Adaptation}
\label{sec:consistency_output}
The idea of consistency output adaptation is to enforce two or more different network outputs to be similar using a dedicated loss function. 
In UDA research, several approaches employ consistency learning in the output space. \par
RPT \cite{zhang2020transferring} proposes an entire consistency framework combining three different levels of consistency. 
For patch-wise consistency, superpixels are computed, and all pixels within these superpixels are enforced to have the same predicted class.
A similar strategy is conducted for cluster-wise consistency, where the superpixels are grouped into clusters and enforced to have the same predicted majority-voted class. 
On top, an LSTM is used to enforce source and target to have a similar spatial structure.  \\
\textbf{Augmentations}: It is a widely adopted method in consistency learning to use two or more different style versions of the same image and enforce the network to predict the same outputs since the semantic content, i.e., the classes, are the same. Generally,  one can distinguish two ways to generate different versions of the same image: rule-based like image augmentations and learnable such as GANs and CycleGANs. SUDA \cite{zhang2021spectral} creates two different spectral views of the same target images and applies an $L1$ consistency loss to obtain similar predictions. SVmin \cite{guan2021scale} utilizes the same loss for scale-invariance. 
The target images are downscaled and enforced to be similar to the original resolution's prediction. 
LSE+FL \cite{Subhani2020} applies the same but patch-wise with a cross-entropy loss for consistency. 
A popular and simple way are image augmentations which can severely change image characteristics such as sharpness, contrast, hue, etc. 
Similarly, PA+CCR \cite{ma2021coarse} augments the target images with color jitter and enforces the prediction to be similar to the clean prediction. The standard cross-entropy loss can be used since the clean prediction is treated as a one-hot encoded pseudo-label that the augmented prediction has to match. PixMatch \cite{melas2021pixmatch} applies more sophisticated and multiple different augmentations, including the discrete Fourier transformation. The consistency loss (cf.\ PA+CCR) is applied, making consistency learning and self-training very similar in this setting. DACS \cite{tranheden2021dacs} introduces cross-domain image-level mixing and blurs the distinctive boundaries between consistency learning and self-training. The training can be understood as both self-training on mixed labels and consistency learning to predict the same classes independently from added source content in the target image.\
\textbf{Style transfer}: Another line of work uses a GAN or CycleGAN to obtain different image styles. 
SUIT \cite{li2020simplified} employs a GAN to transfer source images to the target style and then enforces consistency between style-transferred and real source images by cross-entropy loss. SAC \cite{li2020unsupervised} follows a similar idea but trains two distinct networks and enforces consistency using an $L2$ loss. 
CrDoCo \cite{Chen2019crdoco} is similar to this approach but uses a CycleGAN and two domain-specific networks to enforce the output prediction consistency with a bi-directional KL divergence. 
MSS \cite{toldo2020unsupervised} follows a similar approach but applies the consistency loss for both the source and target domain and utilizes the cross-entropy as the consistency loss. 
APODA \cite{yang2020adversarial} employs a more sophisticated technique since the features are perturbed with an adversarial attack, and an $L2$ loss enforces the prediction of both the clean and the perturbed maps to be the same. \\
\textbf{Knowledge distillation}: A popular and straightforward application of consistency learning is knowledge distillation, where the knowledge should be transferred from a teacher network to a student network. SEAN \cite{xu2019self} proposes a typical UDA knowledge transfer framework. After being augmented, the target images are processed by both a student and a teacher network, and an $L2$ consistency loss enforces the two different target predictions to be similar. SE-GAN \cite{xu2021self}, TGCF-DA \cite{choi2019a}, and BiSMAP \cite{lu2022bidirectional} (with KL divergence consistency loss) follow very similar methods. 
Augmentations and teacher-student learning are expected to make consistency enforcement more effective in this setting.
UACR \cite{zhou2020uncertainty} extends this basic idea with an uncertainty module and a second consistency loss. 
Two uncertainty-weighted mean squared error losses (MSE) are applied as the consistency loss to enforce student and teacher to generate similar predictions. 
At the same time, a class-wise mask is used to enforce consistency between perturbed and non-perturbed images.
Notably, these losses are the only adaptation losses applied in this approach. 
MRNet \cite{zheng2021rectifying} is distinct from the other works since the authors argue that a second additional classifier with a shared encoder can also act as a teacher and regularize the main model; the KL divergence loss is used to obtain output consistency. \par
CAMix \cite{zhou2021context} uses the method from DACS and extends it by knowledge transfer and a so-called significance mask. 
This is computed based on the entropy of the target prediction and a contextual mask using spatial similarities between the source and target domain. 
The original domain mixture idea from DACS is further extended by DSP \cite{gao2021dual}. 
It pastes domain-specific content in both directions, so source and target domain images are modified with content from the other domain. The cross-entropy loss is a combination to enforce both the source and the target content-based predictions to be consistent with the corresponding unmixed labels. 
The clean predictions are obtained from the teacher, and the mixed prediction from the student model, so it also enforces consistency between these two models. 
SAC \cite{araslanov2021self} relies on strong image augmentations for the inputs for a momentum network that is updated as a moving average of the student network. In contrast to UACR, a single focal loss enforces the consistency between the momentum and segmentation network. Similar to UACR, predictions from multiple crops are averaged to obtain more confident pseudo-labels. BiSIDA \cite{wang2020consistency} combines knowledge distillation and style transfer. 
It processes the original target image and several different style transferred images of that original image through the network. 
The style transfer predictions are averaged and utilized as the pseudo-label in the consistency loss.\par
ProDA \cite{zhang2021prototypical} introduces two novel extensions worth mentioning. First, it initializes the student network with weights from a self-supervised pre-training on ImageNet, which provides a strong bias towards diverse real-data representations. Second, it performs multiple iterations of distillation, providing further performance improvement.\par
MFA \cite{zhang2021multiple} proposes the probably most complex UDA consistency framework by combining two knowledge distillation units each consisting of a teacher and a student, resulting in four networks overall. However, the consistency mechanism, embedded into the larger hybrid framework, is similar to other works enforcing the student and teacher networks prediction to be similar by optimizing for an $L1$ loss.\par
PIT \cite{lv2020cross} introduces a relaxed consistency loss between the fine- and coarse-grained network branches. 
The known $L2$ loss enforces the class activation maps of both branches to be similar. 
However, learnable weights for the coarse branch allow some adaptation between the branches and, therefore, relaxation of the consistency condition. 
CDGA \cite{kim2020cross} shows that consistent adaptation can also be conducted on the class distributions predicted from an additional network, which are enforced to be similar according to an $L2$ loss.
In contrast to the other works, ASAnet \cite{zhou2020affinity} enforces local region-wise affinity consistency within the same image for both the source and target domain. The goal is to obtain the same predicted class for all pixels in a certain region except at the semantic boundaries.\par
SAM \cite{chung2021exploiting} is the only work combining consistency and a self-attention learning mechanism.
A self-attention module receives the segmentation output, and an $L1$ loss then enforces the predicted output to be similar to two  self-attention maps. 
This should improve adaptation since the attention maps enforce a focus on inter-pixel correlations.

\subsubsection{Depth-Based Output Adaptation}
It is a straightforward idea to enrich the domain adaptation process with additional or surrogate information to simplify the adaptation process. 
A dominant modality is depth information because of its close relation to the actual semantic segmentation map and because it is possible to obtain ground truth without human labeling effort. 
Since the output of the depth estimation is mostly utilized, depth-enriched adaptation can be seen as another category of output space adaptation.
Depth-based adversarial output adaptation methods were already described in Section \ref{sec:adversarial_output}.\par
SPIGAN \cite{lee2018spigan} proposes a framework that may utilize multiple kinds of additional information from the source domain but evaluates on depth data. 
It trains a second decoder (with a shared encoder) network for depth estimation in the source domain with an $L1$ loss. 
GUDA \cite{guizilini2021geometric} builds upon a similar architecture as SPIGAN, but extends it with new components. Next to the depth estimation, the additional prediction of depth surface normals serves as a regularization for the depth prediction task. 
More importantly, domain adaptation may benefit in two ways from the depth information.
First, via the shared encoder, which additionally learns depth prediction for the source domain. Second, via an image synthesis task, where both the target depth prediction and the previous frames are required to predict the target image.\par
DBST \cite{cardace2022plugging} is the only approach that incorporates self-supervised depth estimation on the target domain to  obtain depth labels for this domain, which is different from CorDA, where the depth is not used as a label in the target domain. DBST contains two separable units that rely on depth information. The first unit trains one network on the depth labels in both domains and a second network on the segmentation labels of the source domain. A transfer network then predicts the semantic output from the depth network so that the depth knowledge of the target is utilized for the segmentation task. The second unit can be seen as a depth-guided version of DACS~\cite{tranheden2021dacs}. The depth information is leveraged to mix source and target content in a more meaningful way and to generate a more diverse dataset for self-training.

\subsubsection{Other Methods}
Next to the already described large methodological cluster with many different proposed methods, there are a few works in the line of UDA output space adaptation that cannot be assigned to any previous clusters.\par
Clustering-based adaptation methods are often applied in the feature space, as described in Section \ref{sec:featSpaceAlign}. However, CDGA \cite{kim2020cross} attaches an additional clustering network of two convolutional layers directly after the semantic prediction output. The clustering network is trained with two different losses. One loss enforces class distribution clustering consistency between source, and target and the second loss minimizes a cosine similarity across the predicted clustered class distributions. The original class distribution will be clustered into a fixed number of sub-clusters strengthening the inter-class adaptation.\par  
The accurate segmentation along the boundaries between objects is still a challenge for segmentation in general. 
Most  UDA works ignore the particular adaptation of object boundaries, but two approaches specifically aim to utilize the boundaries for adaptation purposes.\par
BAPA-Net \cite{liu2021bapa} builds upon DACS~\cite{tranheden2021dacs} and uses semantic boundaries in two ways. 
First, it weighs the standard cross-entropy loss by the distance of each pixel to the mixed boundary of the source and target mixed image. This weighting should enforce the network to focus on the domain mixed boundaries. 
Second, the opposite strategy is applied for prototype alignment in the feature space, and the mixed boundary pixels are excluded to not confuse the prototype alignment.
In contrast, SEDA \cite{chen2021unsupervised} proposes an entire semantic boundary prediction framework. A second network branch is trained to predict the semantic boundaries in the source domain, and a feature-level adversarial loss helps to obtain accurate semantic boundary predictions in the target domain. An $L1$ consistency loss between the predicted boundaries of the second network and the actual boundaries of the predicted semantic segmentation map makes it possible to transfer the knowledge of the boundary branch to the actual segmentation branch.

\subsection{Hybrid Domain Adaptation}
\label{subsubsec:Hybrid_Domain_Alignment}
\begin{table}[t]
    \centering
    \setlength{\tabcolsep}{.5em}
    \begin{tabular}{cccC{4cm}}
        \toprule
        \multicolumn{3}{c}{\textbf{Adaptation Space}} & \textbf{Approach} \\
        \cmidrule{1-3}
        Input & Feature & Output & \\
        \midrule
        \multirow{1}{*}{\checkmark}&\multirow{1}{*}{\checkmark}& & 
        \cite{gong2019dlow},
        \cite{ye2020light},
        \cite{luo2020adversarial}, 
        \cite{wu2018dcan}, %
        \cite{hoffman2018cycada},           
        \cite{shim2021learning} 
        \\
        \midrule
        \multirow{6}{*}{\checkmark}&&\multirow{6}{*}{\checkmark} &
        \cite{lin2019adapting},
        \cite{musto2020semantically}, 
        \cite{wang2020consistency},
        \cite{chen2019learning},
        \cite{chang2019all},
        \cite{yang2020phase}, 
        \cite{song2020learning},
        \cite{li2019bidirectional}, %
        \cite{kim2020cross}, %
        \cite{zhou2020affinity}, 
        \cite{yang2020label},
        \cite{lee2018spigan}, 
        \cite{cheng2021dual}, 
        \cite{xu2021self},
        \cite{saporta2020esl},
        \cite{li2020unsupervised}, 
        \cite{chiou2022beyond},
         \cite{yang2020fda}, %
         \cite{zhang2021spectral}, 
         \cite{huang2021cross},
          \cite{kim2020learning}, %
        \cite{zhou2020uncertainty}, 
        \cite{zhou2021context},
        \cite{huang2021mlan},
        \cite{tranheden2021dacs},
        \cite{tsai2019domain}, %
       \cite{zhu2018penalizing}, 
        \cite{ruan2019category},
        \cite{guizilini2021geometric}, 
        \cite{chung2021exploiting} %
        \cite{yan2021pixel},
        \cite{choi2019a}
        \\
        \midrule
        &\multirow{6}{*}{\checkmark}&\multirow{6}{*}{\checkmark} & 
          \cite{chao2021rethinking},%
           \cite{Dong2019a}, %
        \cite{li2020variational} %
        \cite{zhang2022unsupervised}, %
        \cite{huang2021mlan}, %
        \cite{zhang2021prototypical}, %
        \cite{yu2021dast}, %
        \cite{luo2021away}, %
        \cite{dong2020cscl}, %
        \cite{chen2021unsupervised}, %
        \cite{zhang2020transferring}, %
         \cite{marsden2021contrastive}, %
        \cite{xie2021spcl}, %
        \cite{kang2020pixel}, %
        \cite{huang2021category}, %
        \cite{niemeijer2021combining}, %
        \cite{wang2021separable}, %
        \cite{li2021semantic}, %
        \cite{xu2019self2}, %
        \cite{zhang2020towards}, %
        \cite{xie2022sepico}, %
        \cite{huang2020contextual}, %
        \cite{liu2021adversarial}, %
         \cite{wang2021cross}, %
          \cite{yang2020adversarial}, %
          \cite{zheng2021rectifying}, %
        \cite{xu2022unsupervised}, %
        \cite{shen2019regularizing}, %
        \cite{tsai2018learning}, %
        \cite{zhang2020joint},  %
        \cite{tang2020towards}, %
        \cite{xu2019self}, %
         \cite{Vu2019dada}, %
        \cite{zhang2019category}, %
        \cite{Niemeijer2022LowCoplex}
           \\
        \midrule
        \multirow{5}{*}{\checkmark}&\multirow{5}{*}{\checkmark}&\multirow{5}{*}{\checkmark} & 
        \cite{paul2020domain}, %
        \cite{Chen2019crdoco}, %
        \cite{toldo2020unsupervised}, %
        \cite{tang2021unsupervised}, %
        \cite{yang2021context}, %
         \cite{kim2020cross}, %
         \cite{wang2020differential}, %
          \cite{lee2020unsupervised}, %
          \cite{chung2021maximizing}, %
        \cite{gao2021addressing}, %
        \cite{dong2021and}, %
        \cite{cai2020exploiting}, %
        \cite{ma2021coarse}, %
        \cite{lu2022bidirectional}, %
        \cite{araslanov2021self}, %
        \cite{liu2021bapa}, %
        \cite{zhou2021domain}, %
        \cite{wang2021domain}, %
        \cite{huo2022domain}, %
        \cite{cardace2022plugging}, %
        \cite{gao2021dual} %
         \cite{liu2021domain}, %
        \cite{zhang2021multiple} %
        
        \\
        \bottomrule
    \end{tabular}
    
    \caption{\textbf{Hybrid adaptation} approaches employing techniques in \textbf{multiple adaptation spaces}. The papers are clustered and sub-clustered according to the employed adapation spaces.}
     \label{tab:hybrid_approaches}
\end{table}
It became evident early in the research that methods of the different adaptation spaces can be combined to increase performance. 
A large group of research works has emerged from this idea, and we refer to these approaches as \textit{hybrid} domain adaptation approaches. 
The complexity of different approaches and ways to combine techniques is large. Therefore, we provide a two-leveled grouping to ease the overview. The first-level grouping is done according to the variations of how the different spaces can be combined so that we obtain four different fields, as shown in Table \ref{tab:hybrid_approaches}.\par
For the second level grouping, we introduce the terms mutually independent and mutually dependent approaches. 
Mutually independent describes approaches where the different methods are combined independently so that the approach would still work without one of the spaces.
That, in turn, means that the methods from the different spaces do not directly rely on each other w.r.t.\ the information flow. A simple example would be, e.g., a style transfer method with multiple loss functions for input space alignment. 
To increase output alignment, softmax-based self-training can be "attached" so that both techniques build a framework but are still independent. 
Mutually dependent approaches combine techniques that closely interact with each other and are directly dependent on the other space, e.g., style transfer provides the input for output consistency learning.\par
The advantage of hybrid methods is that the performance increase in the target domain is, in most cases, significant.
A detailed analysis of the performance capabilities of the hybrid approaches will follow in Section \ref{sec:evaluation}. 
A critical analysis of the limitations and disadvantages of hybrid approaches follows in Section \ref{sec:discussion} as part of our meta-analysis. 
\subsubsection{Input and Feature Space Adaptation}
In this section we will discuss approaches that combine input and feature space adaptation techniques. We will start with with mutually independent followed by mutually dependent approaches.
\\ \textbf{Mutually independent approaches}:
CyCADA \cite{hoffman2018cycada} is one of the most popular approaches that combines input and feature-level techniques in a mutually independent manner. 
Next to a style transfer with a CycleGAN, adversarial learning on the feature-level is applied.
The approach DLOW \cite{gong2019dlow} works in the same way and only extends the style transfer by a domainness factor for higher style diversity. 
Closely related to that, GAM \cite{huang2018domain} utilizes CycleGAN-transferred images for pre-training and independently applies deep activation matching afterward.
Likewise, the idea of DACL \cite{shim2021learning} is similar but applies contrastive learning in the feature space.\\
\textbf{Mutually dependent approaches}:
LWC \cite{ye2020light} combines input-level and feature-level adversarial learning within one framework. 
However, both techniques interact, and the feature-level adversarial learning is enabled by the input style transfer forming a mutually dependent approach.\par
ASM \cite{luo2020adversarial}  is different because it utilizes an autoencoder-based style transfer to generate mini-batches with different stylized versions of the same image. 
This is necessary to enforce feature consistency across the mini-batch.

\subsubsection{Input \& Output Space Adaptation}
In this section we will discuss approaches that combine input and output space adaptation techniques. Again, we will start with with mutually independent followed by mutually dependent approaches. 
\\ \textbf{Mutually independent approaches}:
APL \cite{song2020learning} and DISE \cite{chang2019all} are exemplary approaches for this sub-cluster with a focus on input space adaptation.
APL consists of an input-level image reconstruction adaptation along with self-training. DISE employs a complex input adaptation module in combination with output space adversarial learning. 
Similarly, LTIR \cite{kim2020learning} first aims to learn texture-agnostic representations by both domain-randomized and translated images, followed by the second stage with adversarial learning and self-training.
Unlike these approaches, PCEDA \cite{yang2020phase} focuses on input and output adaptation by Fourier phase consistent style transfer and an additional network to encode the source segmentation priors in the output space.\par
A large group of approaches focuses on output space adaptation and where the input space adaptation is added as an independent sub-component.
The three methods PixIntraDA \cite{yan2021pixel}, MAGD \cite{lin2019adapting}, and MLAN \cite{huang2021mlan} have in common that they focus on output-level adversarial learning but additionally utilize a Cycle-GAN-based style transfer to increase the performance further. ASANet+ \cite{zhou2020affinity} focuses on output space structure learning but includes a style transfer to show the orthogonality of their method. 
In contrast to the other approaches, SPIGAN \cite{lee2018spigan} and GUDA \cite{guizilini2021geometric} include depth information in their adaptation methods, and both conduct image-level alignment. Additionally, SPIGAN attaches an adversarial-based technique to their multi-task depth and segmentation network. 
Different from that, GUDA combines depth prediction with a view synthesis module.\par
PTP \cite{zhu2018penalizing} and CAA-Net \cite{ruan2019category} are distinct from the other approaches by combining image reconstruction techniques with output space methods. 
PTP is special since it utilizes the so-called conservative loss in the output space.
\\ \textbf{Mutually dependent approaches}:
The mutually dependent combination of style transfer and self-training closely relates to ensemble-like learning.
FDA \cite{yang2020fda}, SAC \cite{li2020unsupervised}, and SIT \cite{chiou2022beyond} all share the same hybrid idea of generating multiple versions of the same image using style transfer and training multiple networks to obtain the pseudo labels. DPL \cite{cheng2021dual} employs two networks to process images in both translation directions. 
All three methods, style transfer, adversarial learning, and self-training, are applied for both translation streams.
This group of approaches obtains a better-aligned input space and directly utilizes that to increase the confidence of the pseudo labels for output space alignment.
In contrast to these approaches, the hybrid idea of DACS \cite{tranheden2021dacs} is more straightforward because it computes pseudo labels only based on mixed images from both domains. CVRN \cite{huang2021cross} and SUDA \cite{zhang2021spectral} both differ from the other approaches since they focus on consistency between different styles. 
CVRN combines inter-style and inter-task regularization loss, and SUDA combines input adversarial learning with a consistency loss for the different stylized image versions.\par
Several other methods integrate style transfer, self-training (or a different output space adaptation method), and adversarial output learning into adaptation frameworks. 
SA-ITI \cite{musto2020semantically} combines these three methods, while BDL \cite{li2019bidirectional} has to be highlighted because they propose a framework that utilizes more interaction between the two spaces. 
The learned segmentation model is utilized for the perceptual loss of the translated images. 
The framework uses an iterative interaction between input and output space next to self-training and adversarial learning.\par
Another combination of input and output adaptation as a mutually dependent framework is knowledge distillation with a teacher and student network. 
TGCF-DA \cite{choi2019a} proposes an exemplary framework where the source images are translated  to a target-like style and used as input for the student network. UACR \cite{zhou2020uncertainty} and CAMix \cite{zhou2021context} follow a similar scheme, but CAMix inputs domain-mixed images to the student network instead of a style transfer. 
In contrast, BiSIDA \cite{wang2020consistency}  employs style transfer in both directions; therefore, student and teacher networks receive images from both domains with a shared style.

\subsubsection{Feature and Output Space Adaptation}
In this section we will discuss approaches that combine feature and output space adaptation techniques. Again, we will start with with mutually independent followed by mutually dependent approaches. \\
\textbf{Mutually independent approaches}:
AdaptSegNet \cite{tsai2018learning}, SEDA \cite{chen2021unsupervised}, MLAN \cite{huang2021mlan}, CGDA \cite{luo2021away},  
and  CrCDA \cite{huang2020contextual} all utilize the adversarial learning for distribution alignment in the output and the feature space.
Similarly, CLS \cite{liu2021adversarial} and DAST \cite{yu2021dast} combine the adversarial alignment of distributions in the feature space with a self-training method in the output space.
This cluster contains self-supervised learning techniques introduced in Section \ref{sec:featSpaceAlign} and self-training approaches described in Section \ref{sec:outputSpaceAlign}. 
The approaches SSS+ST \cite{niemeijer2021combining} and SePiCo \cite{xie2022sepico} apply contrastive clustering as described in \ref{sec:featSpaceAlign} 
and self-training in the output space.  
 SWLS \cite{Dong2019a}  falls in a similar category, but they utilize an adversarial loss for output space alignment.
A common strategy for mutually independent feature and output space alignment is the utilization of feature-level adversarial learning \cite{tsai2018learning} in addition to another output technique. Several approaches follow this idea. RPT \cite{zhang2020transferring} proposes output patch consistency. SSDA \cite{xu2019self2} combines adversarial learning with self-supervised pretext tasks. 
 JAL \cite{zhang2020joint} adds a weight transfer, while CRA \cite{wang2021cross} is proposed as an additional technique to any UDA method and can be combined with adversarial learning.
VAE-UDA \cite{li2020variational} applies an autoencoder-based output space alignment and adversarial alignment. PFR \cite{zhang2020towards} and SRDC \cite{tang2020towards} are slightly distinct from these works because they utilize output adversarial learning in combination with style minimization and feature clustering, respectively. The authors of SEAN \cite{xu2019self} instead combine a self-attention mechanism in the feature space with an output consistency loss. \par
The approaches DADA \cite{Vu2019dada} and CTRL \cite{saha2021learning} apply an implicit distribution alignment in the feature space 
by training depth regression on the source and target domain and an adversarial alignment of the distributions in the output space. 
Similarly, GUDA \cite{guizilini2021geometric} and CorDA \cite{wang2021domain} utilize depth regression as self-supervised training but use self-training in the output space.
\\ \textbf{Mutually dependent approaches}:
The approaches MCD \cite{Saito2018a}, SWD \cite{lee2019sliced}, and BCDM \cite{li2020bi}, which utilize the maximum classifier discrepancy, fall into this category and have a very close and crucial interaction between feature and output space. 
 Their three-step iterative adversarial learning scheme (see Section \ref{sec:featSpaceAlign}) works because the feature extractor and two classifier heads are updated alternatingly, so that feature- and output-alignment directly support each other.\par
A crucial challenge for contrastive learning is the definition of semantically meaningful positive and negative pairs. 
Often class information is accessed to guide the selection of positive and negative pairs, which gives a close mutual dependence between feature and output space. Different approaches such as ProDA \cite{zhang2021prototypical}, CLST \cite{marsden2021contrastive}, SPCL \cite{xie2021spcl}, and EPS-UDA \cite{xu2022unsupervised} follow this principle. 
The actual adaptation happens in the feature space, but reliable pseudo labels in the target domain are required, so both spaces are strongly dependent. The additional application of self-training is widespread.\par  
Similar to this principle, another group of mutually dependent approaches directly utilizes the feature prototypes or anchors for assigning pseudo labels. This provides a strong dependency of both spaces since the quality of pseudo labels directly relies on the extracted prototypes. SCDA \cite{li2021semantic} is an exemplary work for this, and also UCDA \cite{zhang2022unsupervised}, CAM \cite{wang2021separable}, and CAG \cite{zhang2019category} utilize this idea. \par
The approach presented in MADA \cite{shen2019regularizing} presents an example of mutual dependency between feature and output space.
The authors apply adversarial training at low-level and high-level feature maps combined with self-training based on the classifier and discriminator confidences. CSCL \cite{dong2020cscl} utilizes a more complex mutual interaction. Next to self-training and adversarial learning, a critic function aims to distinguish between domain-specific and domain-invariant knowledge and closely interacts in the feature- and output space.

\subsubsection{Input, Feature \& Output Space Adaptation}
In this section we will discuss approaches that combine techniques from all three adapation spaces (input, feature, and output space). Again, we will start with with mutually independent followed by mutually dependent approaches. \\
\textbf{Mutually dependent adaptation}:
A notable pattern of mutually dependent adaptation is the utilization of input space based domain mixing, i.e. content from source images is pasted to target images and/or vice versa. All three approaches RCCR \cite{zhou2021domain}, DAP \cite{huo2022domain} and BAPA-Net \cite{liu2021bapa} are building upon this mechanism that was initially proposed by DACS; DISE-CT \cite{lee2020unsupervised} also employs source and target domain mixture. RCCR closely connects the three spaces by processing the mixed images with a student-teacher framework and a consistency loss in the output space. The latent features of the student and teacher network are used for contrastive learning. BAPA-Net \cite{liu2021bapa} instead uses the domain mixture to enforce the boundary consistency on feature and pseudo label-level. DAP \cite{huo2022domain} has to be highlighted in this context since it introduces a novel extension at the intersection of feature and output space also including input space alignment. As the only currently known UDA approach for semantic segmentation it introduces another modality by using word2vec embeddings \cite{mikolov2013distributed} as domain invariant priors and projects them together with the mixed semantic output to enforce similarity between the priors and actual network features.\par
Another group of mutually dependent approaches is formed by CrDoCo \cite{Chen2019crdoco} and MSS \cite{toldo2020unsupervised}. They both utilize feature-level adversarial learning and then connect input and output space by applying style transfer to compute consistency loss between the predictions of the different stylized images. DSP \cite{gao2021dual} and SSAC \cite{araslanov2021self} are very similar because they connect input and output space utilizing a teacher-student framework. 
For DSP, the student network receives domain mixed images, and a weighted CE-loss is computed for both source and target mixed images in the output space. 
Independently from this, a local and global MMD loss is applied in the feature space. Similarly, SSAC applies augmentations to obtain different versions of the same image. Additionally, target BatchNorm adaptation is conducted independently in the feature space.\par
Unlike from previously described works, another familiar pattern of mutually independent approaches is the close interaction between feature and output space. A popular representative of this idea is SIM \cite{wang2020differential}. Feature and output space are closely connected to compute class-wise feature representations in the latent space and minimize the distance between source and target features. 
Independently from that input space adaptation following BDL and adversarial feature adaptation is applied. 
DCAA \cite{cai2020exploiting} also has an independent input adaptation module. 
However, the attention-based feature adaptation and self-training output adaptation interact by using the attention weights for the pseudo labels and an attention discriminator. BiSMAP \cite{lu2022bidirectional} instead introduces a novel utilization of the three adaptation spaces. 
First, a gaussian mixture model in the feature space is used to assign the pseudo labels, which are used to train a student-teacher framework along with a consistency loss.\par
Distinct from the previous works, KATPAN \cite{dong2021and} employs three mostly independent domain adaptation modules in input, feature, and output space. The feature adaptation module has a connection to the style transfer module making, KATPAN a mutually dependent approach. The feature transferability information is used to weigh the style transfer bottleneck and improve the input-level transfer of well-transferable regions.\\
\textbf{Mutually independent adaptation}:
Some works mainly vary in feature space adaptation methods among these approaches.
An exemplary approach for this is CIR~\cite{gao2021addressing}, where the style transfer with a CycleGAN and adversarial discriminator acts independently from the attention mechanism in the feature space and output-level self-training. CADA \cite{yang2021context} is very similar to that in the input and output space; only the feature-level channel and spatial-wise attention mechanism are different. The same applies to MCSSF \cite{chung2021maximizing}, which employs standard input and output space methods but uses a cosine similarity-based feature centroid alignment. 
WDA \cite{paul2020domain} shares the same underlying idea with slightly different modules. 
It combines an attention mechanism in the feature space with class-wise discriminators and image-level class existence prediction on the output level.\par
There are two contrastive learning frameworks among the mutually independent approaches. 
CFContra \cite{tang2021unsupervised} and PWCL \cite{liu2021domain} share the idea of embedding feature-based contrastive learning methods into a larger framework with style transfer. 
Both apply entropy minimization in the output space.
 In PWCL, a patch-matching module is required to compute positive and negative pairs, and self-training is conducted.
CorDA \cite{wang2021domain} and DBST \cite{cardace2022plugging} utilize domain mixture techniques along with depth information in slightly different ways. Building upon DACS, CorDA uses depth information and a feature-level attention mechanism to enable knowledge exchange between the depth and semantic stream, followed by a depth disparity-based output alignment. DBST connects the three spaces in a different mutually independent manner because it first applies a feature-level transfer between two networks before a depth-extended DACS version is used.\par
In contrast to the other works, only the input space alignment of PA+CCR \cite{ma2021coarse} enforces inter-domain alignment by style transfer. 
The feature space centroid alignment and output space consistency alignment work independently and only within the source and target domain, respectively.

\label{subsec:visiontransformers}
\section{Vision Transformer Networks for UDA}
This section will describe the novel and recently emerging field of vision transformers. 
We aim to give the reader a better intuition of why vision transformers show promising results on domain adaptation benchmarks and why they could be an exciting research direction for domain adaptation. 
For this reason, we will first briefly overview this research field in general and focus on the novel properties of these networks.
The following will review the existing UDA works utilizing vision transformer networks. \par
Transformer networks with attention mechanisms were initially developed for language processing \cite{vaswani2017attention}. 
Starting with the foundational work from Dosovitskiy et al.\ with \network{ViT} \cite{dosovitskiy2020image}, the transformer networks recently gained much attention in computer vision and showed promising results on several benchmarks and applications \cite{khan2022transformers}. 
For semantic segmentation, several new architectures were developed, e.g., \network{Swin} transformer \cite{liu2021swin}, pyramid transformer \cite{wang2021pyramid}, \network{SegFormer} \cite{xie2021segformer}, and recently \network{HRViT} \cite{gu2022multi}.\par 
The self-attention mechanism is the major change compared to standard convolutional architectures such as the \network{ResNet} versions.
The self-attention mechanism, originating from language processing, learns the relations between the elements of a sequence of inputs and aims to capture how the sequence element influence each other.
In computer vision, the input often is not a sequence but a single image. 
For this reason, \network{ViT} \cite{dosovitskiy2020image} divides one image into image patches of $16\!\times\!16$ pixels replacing the sequence known from language processing. 
The self-attention mechanism learns the relations between the image patches. \network{SegFormer} \cite{xie2021segformer} uses smaller $4\!\times\!4$ patches for semantic segmentation.\par
Several works indicate that vision transformer networks have higher robustness against perturbations and better generalization capabilities than CNNs \cite{shao2021adversarial, naseer2021intriguing, mao2022towards}. 
In contrast, Bai et al.\ \cite{bai2021transformers} and Wang et al.\ \cite{wang2022can} show with their works that CNNs can achieve similar adversarial robustness when transferring certain elements except for the self-attention mechanism of vision transformer training to CNNs. 
Also, the recently proposed \network{ConvNeXt} \cite{liu2022convnet} outperforms vision transformer networks with modifications for a CNN-based architecture, thereby questioning the superiority of the vision transformers.
That indicates that research for vision transformers is still at its beginning, and more new research findings can be expected. 
Next to these partially contradicting results, there are some first findings on how the learned representations from vision transformer networks differ from CNNs and why that might impact their robustness and generalization capabilities. 
The first difference is related to the self-attention mechanism, which is supposed to learn the relations between different patches.
This causes a larger receptive field \cite{xie2021segformer} and enables the transformer networks to incorporate more global contextual information at the early layers \cite{raghu2021vision}, which might be one reason why occlusions cause a smaller performance drop than for CNNs \cite{naseer2021intriguing}. 
Second CNNs are considered sensitive to texture \cite{geirhos2018imagenet}, a crucial problem in domain adaptation. 
In contrast to that and following the current research, vision transformers focus more on the shape of objects, making them more robust to texture shifts \cite{naseer2021intriguing}.
Third, Raghu et al.\ \cite{raghu2021vision} observed that vision transformer networks could better propagate location information through the network than CNNs, which is a beneficial property for localization tasks such as detection and segmentation.\par
\network{DAFormer} \cite{Hoyer2022DAFormer} can be seen as the foundational work with vision transformers for unsupervised domain adaptation. 
It proposes novel contributions to both the method- and architecture-level. 
This combination caused a major step in the SOTA performance outperforming the previous best approach ProDA \cite{zhang2021prototypical}, by more than 10 \% mIoU. 
On the architecture-level \network{DAFormer} builds upon \network{SegFormer} \cite{xie2021segformer}, which is used as the encoder architecture. 
Two well-known methods from the segmentation DNNs are utilized. 
First \network{DAFormer} introduces skip-connections between the encoder and decoder to transfer low-level knowledge better. It then uses an ASPP-like~\cite{Chen2018a} fusion where the stacked encoder outputs from different levels are processed with different dilation rates, which should further increase the receptive field. 
On the method-level DAFormer partially adapts known UDA methods for CNNs. 
Self-training with a teacher-student framework, strong augmentations, and softmax-based confidence weighting is employed. 
In addition, rare class sampling on the source domain and a feature distance loss to the pre-trained ImageNet features are part of the DAFormer approach. 
An interesting side observation is that learning rate warm-up methods can be beneficial for UDA.\par
The second important vision transformer contribution and current SOTA work HRDA \cite{Hoyer2022HRDA} work directly builds upon \network{DAFormer}. 
Its major contribution is a scale attention mechanism. 
The network receives two inputs; one high and one low resolution input. 
The scale attention then learns to assign attention scores that decide whether low- or high-resolution input should get higher weighted. 
The idea behind that method is that different classes and objects are easier to learn on specific scales, and, e.g., contextual information can be better extracted from smaller crops. 
Self-training is applied using a sliding window to generate pseudo labels. 
This overall further improves the \network{DAFormer} performance but still leaves a performance gap.\par
TransDA \cite{chen2022smoothing} observes that the vision transformer can have a so-called high-frequency problem. 
Using the \network{Swin} Transformer \cite{liu2021swin} architecture, Liu et al.\ show that it generates target pseudo labels and features that change more significantly and with a higher frequency over the iterations than for a \network{ResNet-101}. 
Therefore they argue that the high-frequency problem only affects vision transformer networks. 
TransDA proposes feature and pseudo label smoothing using a momentum network to reduce the high-frequency flickering along with self-training and weighted adversarial output adaptation. 
This is similar to the teacher-student adaptation approaches known for CNNs, as described in Section \ref{sec:consistency_output}.\par
Next to these three approaches with methods tailored explicitly for vision transformer networks, a rising number of works evaluates on them. 
ProCST \cite{ettedgui2022procst} follows the idea of hybrid adaptation and applies style transfer in the input space in addition to DAFormer \cite{Hoyer2022DAFormer} and HRDA \cite{Hoyer2022HRDA}. 
Several other works, which are already described in the previous sections, combine their methods with the \network{DAFormer} framework and further improve the performance by small margins \cite{xie2022sepico, zhou2021context, vayyat2022cluda, du2022learning} but without beating HRDA. However, CLUDA \cite{vayyat2022cluda} also builds upon HRDA and further improves this performance.

\section{Quantitative Comparison of UDA Approaches}
\label{sec:evaluation}
After the extensive review of UDA techniques in the previous section, we will conduct a large-scale performance analysis, including all UDA approaches. 
Therefore, first, an overview of the most common UDA performance metrics and tools is given.
Afterward, the comparison method is described, and finally, new insights about the performance capabilities and development are revealed.
\subsection{Performance Metrics and Tools}\label{subsec:metrics}
The dominant quantitative evaluation metric is the mean intersection over union $\mathrm{mIoU}=\frac{TP}{TP+FP+FN}$, with $TP$ being the true positive predicted pixels, $FP$ the false positive, and $FN$ the false negative ones. 
It is a well-established metric for semantic segmentation to quantify the segmentation quality and is used by all included papers of our quantitative comparison. 
Depending on the complexity of the specific approach, the $\mathrm{mIoU}$ is utilized to assess and compare the performance of several different configurations. 
In addition, usually, the class-wise $\mathrm{IoUs}$ are reported. 
That is important since not all classes may benefit similarly from the adaptation, and the class-wise $\mathrm{IoUs}$ allow a more fine-grained assessment. \par
Some works utilize a t-SNE \cite{van2008visualizing} analysis, to visualize domain alignment: \cite{yan2021pixel, toldo2021unsupervised, zhu2018penalizing, huang2020contextual, guo2021metacorrection} 
Our analysis shows that it is the second most used method to assess domain alignment, even though it is a qualitative method and not a quantitative metric.
T-SNE can visualize high-dimensional feature distributions in the 2-dimensional space by applying nonlinear dimensionality reduction \cite{van2008visualizing}. 
With this technique, t-SNE is appropriate to visualize and assess the aligned feature distribution of UDA approaches. 
A typical change of the t-SNE plots after adaptation is a better alignment of the target feature centroids with the source centroids \cite{huang2020contextual}. 
Next to this, most of the papers present segmentation maps as a qualitative verification of the performance of their approach. 
However, this can only serve to highlight specific achievements exemplarily. \par
Next to these often applied methods, some metrics only appear in single works like the t-test for comparison with other approaches \cite{dong2021and} or similarity and sparsity scores \cite{toldo2021unsupervised}.\par
Overall it becomes clear that the set and variety of different metrics are limited; namely, only $\mathrm{mIoU}$  for quantitative and t-SNE for qualitative comparison are established, while only the first one allows large-scale comparisons.
That can hurt the kind of findings researchers can draw from their evaluation.
\subsection{Quantitative Performance Comparison}
\subsubsection{Method}
For our large-scale performance comparison across the UDA segmentation approaches, we utilize the $\mathrm{mIoU}$ to measure the segmentation performance. 
These values are reported by all papers of the comparison and can therefore serve as a comparison metric. 
Cityscapes is the de-facto standard benchmark for the target domain. 
Other datasets like NTHU \cite{chen2017no}, A2D2 \cite{geyer2020a2d2}, or BDD \cite{yu2020bdd100k} would be valuable additional evaluations for the real-to-real domain shift.
However, their appearance in the evaluations is too rare for a valid large-scale comparison.
The $\mathrm{mIoU}$ values which we are reporting in this survey are, in all cases, taken directly from the original papers without any modifications, so no individual experiments were conducted
That means that differences in the evaluation protocols of the different papers like resolution etc. (as discussed in Section \ref{discuss:train_issues}) can also have an impact on the reported values. 
The best performance is reported in the case of different reported performance values for different configurations of the same approach. \par
For the comparison, both the performance and the improvement over the source-only baseline are of interest, and we provide both values for each paper if possible.
In several cases, the source-only performance of the approaches is not reported. 
Therefore, it is impossible to provide values for the improvement of these methods.
For GTA5 and SYNTHIA source-only training often, $36.6\%$ and $38.6\%$ mIoU, respectively, are reported. 
However, we cannot assume these values for all papers due to several possible modifications in the source-only training. 
It has to be mentioned that this improvement has to be carefully considered since weaker baselines can cause a larger improvement while the performance is still low. Another highly interesting performance assessment for UDA approaches would be the performance gap to the oracle performance, which means supervised training with labels for the target domain. The number of papers reporting this performance is also limited, so no valid comparison is possible.\par
\begin{figure*}[t!]
    \centering
    \includegraphics[width=0.8\linewidth]{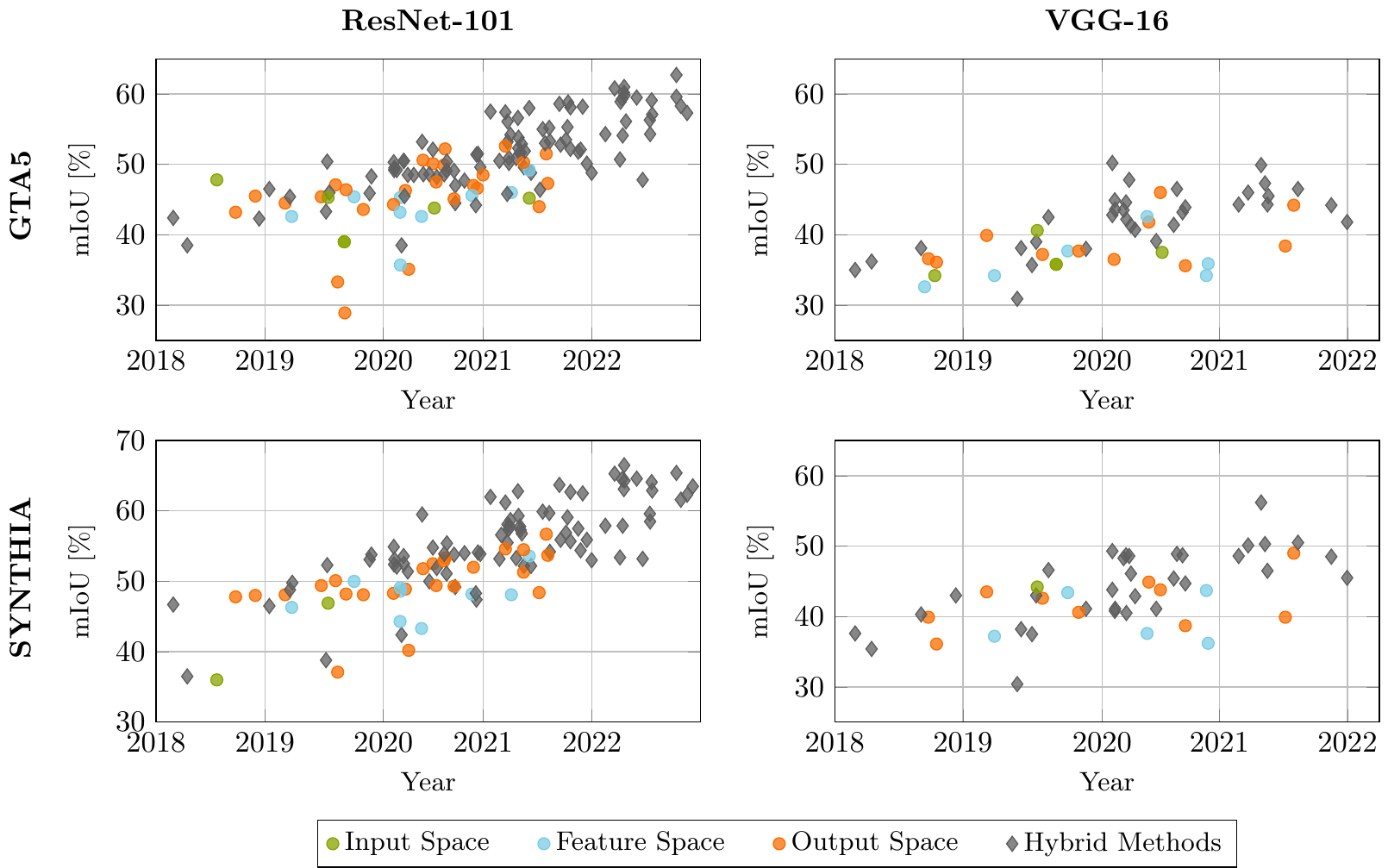}
     \caption{\textbf{Performance (mIoU (\%))} on the Cityscapes validation set after training on the source domains \textbf{GTA5} (top row) or \textbf{SYNTHIA} (bottom row) with simultaneous adaptation to Cityscapes. The results are shown for models based on a \network{ResNet-101} feature extractor (left column) or a \network{VGG-16} feature extractor (right column). The reported values are taken from the respective papers.}
    \label{fig:gta5_adaptation}
\end{figure*}
\begin{figure*}[t!]
    \centering
    \includegraphics[width=1.0\linewidth]{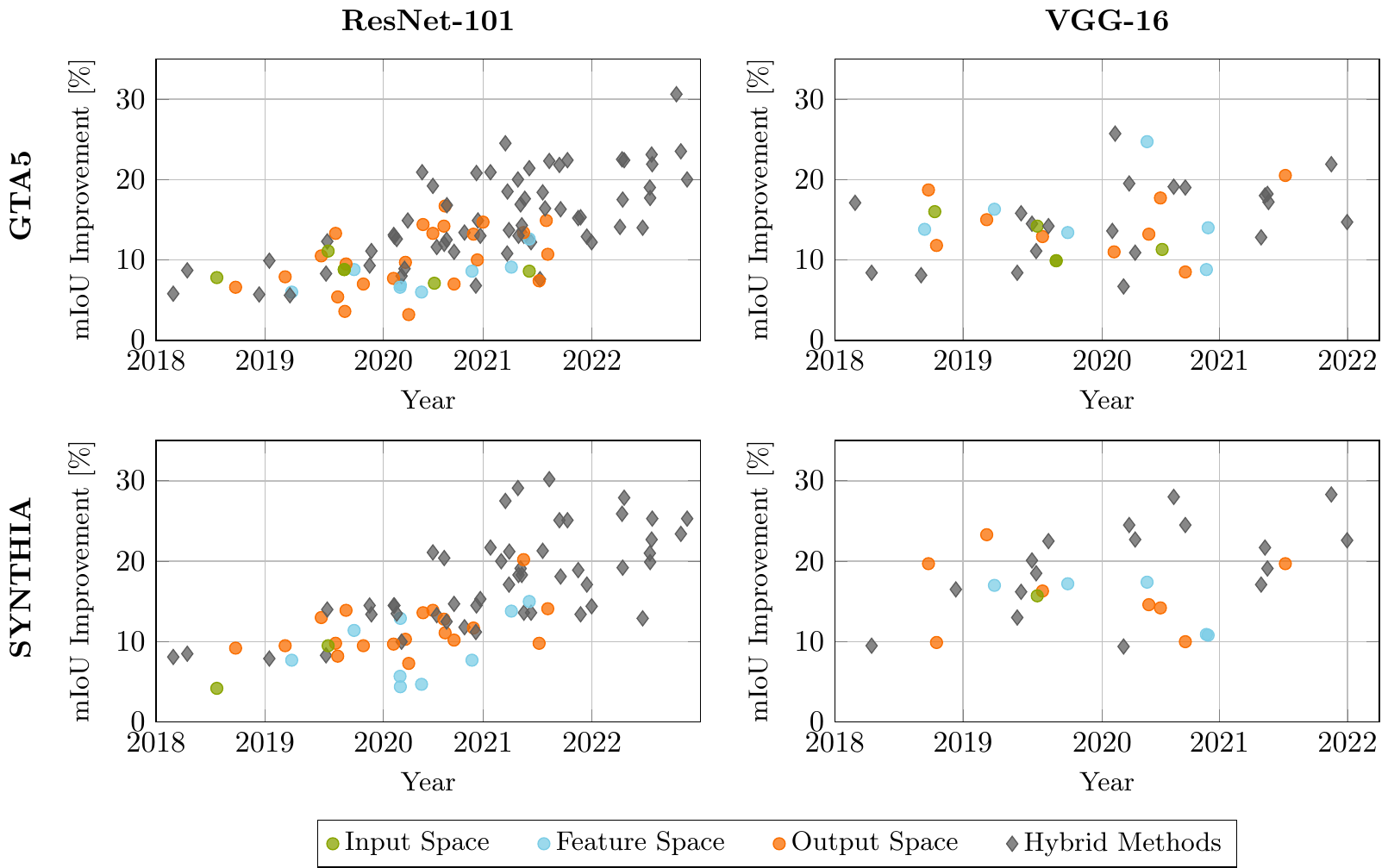}
     \caption{\textbf{Performance improvement (mIoU (\% absolute))} on the Cityscapes validation set after training on the source domains \textbf{GTA5} (top row) or \textbf{SYNTHIA} (bottom row) with simultaneous adaptation to Cityscapes. The results are shown for models based on a \network{ResNet-101} feature extractor (left column) or a \network{VGG-16} feature extractor (right column). The reported values are taken from the respective papers. Note that not all papers provide a baseline performance without adaptation.}
    \label{fig:synthia_adaptation}
\end{figure*}

We present the performance comparison as a plot over time with an assignment to input, feature, output, and hybrid space adaptation. This makes the quantitative progress in UDA research over the years directly observable and reveals new insights about the capabilities of specific methods. 
For better interaction and more detailed information, we also provide the plots as interactive graphics on our project website with links to the papers and more information.
To obtain a fair comparison w.r.t the time of publication, we took the earliest publication date we found for each paper.
Figure \ref{fig:gta5_adaptation} shows the performance, and Figure \ref{fig:synthia_adaptation} shows the performance improvement in \% absolute for both GTA5 and SYNTHIA for the VGG-16 and ResNet-101 backbone. 
As seen by the decreasing number of data points for the VGG-16 backbone, more recent works mainly utilize a ResNet-101 as the backbone.
However, we include the VGG-16 performance to verify that the observations are valid for both backbones. 
Vision transformer networks are not included in these plots because they use a different architecture and would not contribute to a fair comparison. Additionally, the number of approaches using these architectures is still small and does not allow a meaningful comparison yet.    

\subsubsection{Meta Analysis}

The first interesting observation we can draw from the performance comparison is the performance limitation of approaches that only adapt in the feature space. 
None crosses the 50\% $\mathrm{mIoU}$ line for GTA5 as the source and with a ResNet-101 backbone. 
All other feature space approaches reach a performance between 40-50 \% $\mathrm{mIoU}$. 
The observation for SYNTHIA is similar, where only one feature-based approach exceeds the 50\% mIoU target performance. The plot showing the absolute improvement confirms this observation. Among the only feature-based approaches CaCo \cite{huang2021category} reports the best improvement of 12.6 \% absolute for GTA5 as the source and 15.0 \% absolute for SYNTHIA as the source. 
For GTA5 many approaches do not reach an improvement of more than 20\% absolute. These observations also hold for the VGG-16 backbone, where only one of the feature-based approaches reaches an improvement of more than 20\% absolute. 
The observation is, therefore, significant across both datasets and architectures. This empirical observation could indicate that aligning the synthetic and real distributions in the feature space has a limit and cannot provide full alignment standalone. \par  
The observation flips for output and hybrid space adaptation approaches. 
Both significantly outperform the feature space adaptation, particularly the hybrid approaches, which are one of the most important reasons for the performance increase and new SOTA performances in the past two years.
Remarkably, in 2022 all proposed approaches were hybrid adaptation approaches, indicating their major importance for UDA.
In strong contrast, the pure input space adaptation approaches, both quantitatively and qualitatively, play a minor role, indicating that a standalone pixel-level adaptation is insufficient to bridge the complete domain shift and provide a large performance improvement.
The best-performing input space adaptation-only approach is DS \cite{dundar2018a}, with 47.2\% mIoU with GTA5 as the source dataset even close to the feature-space performance. The observed limitations intuitively make sense since input space adaptation does not take aspects like different label distributions, different semantic content, or different geometry into account.\par
Pure output space adaptation approaches are significantly more numerous than input and feature-space. 
However, similar to the other spaces, we can observe a performance boundary, also.
For GTA5 and the ResNet-101 backbone the highest mIoU is reached by UncerDA \cite{wang2021uncertainty} with 52.6\% mIoU and the highest improvement with 16.7\% absolute by IAST \cite{mei2020instance}. 
However, in contrast to the other spaces, several other approaches \cite{wang2020classes, guan2021scale, lv2020cross, li2020content} provide improvements in a similar range like the two highlighted. 
For both SYNTHIA and the VGG-16 backbone, a similar pattern can be observed. 
Notably, for VGG-16, the improvements are slightly higher with up to $\approx20\%$ absolute, but the number of data points is limited.\\   
We can observe that the hybrid approaches made it possible to cross the line of $\approx15\%$ absolute improvement in $\mathrm{mIoU}$ for both GTA5 and SYNTHIA. 
In particular, hybrid approaches clearly carried out the latest performance raise within the last two years that exceeded the 60\% mIoU performance with GTA5. 
Before this development, between the middle of 2019 and the end of 2020, for both SYNTHIA and GTA5, we can see saturation in the performance where most of the approaches remained close to the 50 \% boundary (slightly higher for SYNTHIA).
At the beginning of 2021, several better-performing hybrid works were published, ending this saturation trend.
No standalone method worth highlighting, but the combination of known methods from other deep learning areas, like knowledge distillation, contrastive learning, self-supervised learning, and self-training, led to this increase in the SOTA performance.
ProDA \cite{zhang2021prototypical} used and combined all these techniques, which obtained a new SOTA performance of 57.5\% mIoU and an improvement of 20.9\% absolute, outperforming previous approaches by a significant margin. 
Several other works \cite{wang2021cross, chung2021exploiting, cardace2022plugging} attached additional methods to ProDA, leading to a relatively high density of approaches close to 58\% $\mathrm{mIoU}$ but without providing significant methodical progress. 
The same phenomenon can be observed for vision transformer approaches where several works build upon HRDA \cite{Hoyer2022HRDA} or DAFormer \cite{Hoyer2022DAFormer}. Those approaches \cite{koh2023consistency, vayyat2022cluda, chen2022pipa, wang2022exploring, du2022learning} provide an additional improvement of ~1-2\% absolute and reach up to 75.9\% mIoU \cite{hoyer2022mic} with the current SOTA approach masked image consistency (MIC). That marks a $\approx30\%$ mIoU increase over the respective source-only baseline. In comparison, the strongest reported improvement of a hybrid approach using a ResNet-101 backbone is 30.6\% absolute from DDB \cite{chen2022deliberated}, but a significantly weaker source-only baseline caused this large improvement. More realistically, most of the recent approaches with a similar or slightly weaker performance, compared to DDB, report smaller improvements of 22.4\% absolute \cite{xie2022sepico}, 22.5\% absolute \cite{liu2022undoing}, or 21.8\% absolute \cite{wang2021cross}. This indicates that the maximum improvement by UDA techniques for vision transformer networks is significantly larger than for ResNet-101-based architectures.\par
These observations do not directly apply to the performance with a VGG-16 backbone, where the performance is clearly saturating since 2020. 
Notably, also newer hybrid approaches do not reach new SOTA performance for this setting. 
This does not contradict the ResNet-101 trend because it is mainly caused by the fact that VGG-16 was replaced as the standard architecture by the ResNet.
Therefore newer works either entirely leave the VGG-16 out or may not perform such an extensive hyperparameter optimization as for the ResNet-101. For 2022, none of the works in the database reports their performance for the VGG-16 backbone.\par
Closely related to this is the observation that the performance variance of the feature space approaches is smaller than that of the hybrid space. Visually described, the performance band of the feature space is narrower. 
In contrast, the hybrid space shows a larger scattering of the performances where we still observe works with around 45\% mIoU and high performers with over 60\%. 
That is reasonable since hybrid approaches can strongly vary w.r.t. performance depending on the selection of components
That shows that strong performance is not directly guaranteed but requires carefully selecting the hybrid components.

\section{Discussion}
\label{sec:discussion}
In this section, both the benchmarking settings in UDA and the adaptation approaches themselves will be critically discussed. 
Based on this, this section will conclude with promising future research directions.
\label{sec:discussion}
\subsection{Benchmark Problems}\label{discuss:train_issues}
The task of unsupervised domain adaptation is defined clearly. The entire training and validation process, on the other hand, has been inconsistent so far. 
The following issues are observed when comparing different methods for unsupervised domain adaptation.
\\\textbf{Dataset splits}:
The SYNTHIA dataset does not provide an official split into a training, validation, and test set. Therefore, the whole dataset is usually used for training. 
Validation on the source domain is not considered. 
For GTA5, there is an official split of the dataset, but it is not used universally. 
Some publications train solely on the training split, but others train on the entire GTA5 dataset, i.e., on training, validation, and test set combined.
However, in both approaches, there is no validation of the method on the source domain, contributing to the next issue.
\\\textbf{Checkpoint selection}:
The training and adaptation process on the source and target domains typically takes several epochs, respectively up, to roughly 250,000 iterations. During this lengthy training and adaption process, almost all methods periodically save checkpoints of the neural network on which an evaluation is performed. The results using the checkpoint with the best performance are then reported in the publication. This leads to the following issues: First, the interval, in which checkpointing is performed, is not standardized. 
In some publications, a checkpoint is saved every 2000 iterations~\cite{vu2019advent}, and in others, every 2500 iterations~\cite{yang2020fda}.
Additionally, the total number of iterations can differ from method to method. 
Secondly, the validation for the checkpointing is often performed on the validation set of the target domain (Cityscapes), which is also used to report on in the publication. This means that the validation, which is also used for hyperparameter optimization and checkpointing, is used as a test set, which goes against the basic guidelines of machine learning. There are already some publications using a subset of the training set for validation in order to use an unseen validation set as a provisional test set~\cite{araslanov2021self,Lian2019} or criticizing this and suggesting to use the validation set for validation and the original test set with the benchmark server for testing~\cite{araslanov2021self}. Furthermore, the use of labeled target data for validation is, in our opinion, essentially misleading, as it undermines the concept of domain adaptation and misrepresents the actual performance of the methods in real world applications. 
Suppose there were labeled samples of the target domain. 
In that case, these should be used for training since significantly better results can be achieved with supervised multi-domain training than with an adaptation.
Furthermore, checkpointing on the target domain favors selecting models that perform significantly better on this domain.
This can also lead to overfitting the hyperparameters, as already described in Section~\ref{subsec:methodical_reflection} above. 
\\\textbf{Training hyperparameters}:
Another issue concerns the hyperparameters of training and evaluation, which can significantly impact performance but whose influence is often not adequately reported.
One of the factors is the resolution of the images from the source and target domains that is adopted. 
For example, images from GTA5 (source domain) are usually downsampled to $1280\,\mathrm{px}\!\times\!720\,\mathrm{px}$, and Cityscapes images (target domain) are usually downsampled to $1024\,\mathrm{px}\!\times\!512\,\mathrm{px}$ during training.
From here, there are a wide variety of strategies that some papers have followed but have not explicitly analyzed. Some papers use the full images as their input~\cite{tsai2018learning,vu2019advent} and some papers use image crops for parts of their training~\cite{araslanov2021self}. 
During the evaluation process on the target domain, most papers use the same downsampled resolution as in training, but some methods use less downsampled images, e.g.,  $1344\,\mathrm{px}\!\times\!576\,\mathrm{px}$~\cite{yang2020fda}. 
Many papers do not address the role of resolution and cropping further, although reduced resolution, in particular, can affect performance, especially for small structures.
The different choice of these hyperparameters  also contributes to the next point.
\\\textbf{Baseline performance}:
Another common issue is the lack of comparison with an own source-only baseline. Many methods only compare their adaptation performance with that of other methods. 
It is ignored that the source-only performance can already provide substantially different performances, e.g., by a different choice of hyperparameters and augmentations.
\\\textbf{Non-determinism in DL frameworks}:
Most methods that provide code cannot be re-simulated perfectly due to non-determinism in deep learning frameworks, e.g., PyTorch~\cite{pytorch_doc_determinism}. 
This is likely because either deterministic convolutions algorithms are unavailable or take significantly longer than non-deterministic ones.
Therefore, for many methods, the results may differ from the reported results. 
If no code is published, this makes it even more complicated since, for example, the choice of the random seed can also significantly influence the final results.
There are now publications that address this problem by repeating their training several times and reporting a mean value with the standard deviation as their result~\cite{Hoyer2022DAFormer}.
\subsection{Methodical Reflection}
\label{subsec:methodical_reflection}
When analyzing the current state-of-the-art methods, one can observe that the top-performing approaches tend to be complex hybrid models.
A few examples of such methods are given with ProDA \cite{zhang2021prototypical}, DPL \cite{cheng2021dual}, and MFA \cite{zhang2021multiple}. 
ProDA \cite{zhang2021prototypical} applies self supervised pre-training on image net and is comprised of four training stages in which knowledge distillation, symmetric cross-entropy, contrastive, and adversarial loss functions are applied.
DPL \cite{cheng2021dual} combines a warmup strategy with a cycle GAN for style transfer, and it applies four segmentation losses and two adversarial losses for the final training.
Finally, MFA \cite{zhang2021multiple} consists of two training stages, a warmup phase and a domain adaptation phase in which co-learning is applied.  
In the latter, six self-training losses are applied. 

Recently Sakaridis et al.\ \cite{Sakaridis2021} showed that many approaches that show good performance on the synthetic-to-real domain shift struggle on real-to-real domain changes. 
The synthetic to real domain change is exemplified through the adaptation from either the SYNTHIA \cite{ros2016synthia} or GTA5 \cite{Richter2016} dataset to the Cityscapes dataset \cite{Cordts2016}. 
The real-to-real domain change tested in Sakaridis et al.\ \cite{Sakaridis2021} is the change from Cityscapes to the ACDC dataset that contains diverse environment conditions.
There might be various reasons for the different performance of the same domain adaptation approach on a different domain adaptation benchmark. 
First, the approaches are optimized for the benchmarked synthetic-to-real domain shift, as a result of this introducing a bias in the selection and development of new approaches toward this domain change.
Second, many approaches that are comprised of many elements and hence bare a considerable amount of complexity can be finetuned toward the given benchmarks more easily by finding the optimal hyperparameters.
Such finetuning is done based on the labeled target domain validation set, hence introducing a dependency on target domain supervision. \par
Connected to the issue of approach complexity is the topic of training stability. 
Given that, in practice, often limited or no validation data is given for the target domain,  
a training process for domain adaptation should be robust against hyperparameter setting since no finetuning might be possible.
This touches on the issue of complexity but also the choice of technology.
Adversarial training, e.g., is known to be very sensitive to hyperparameter settings.
Approaches like self-training where a certain closeness of the source and target distribution is assumed, might suffer when applied to strong domain changes.
A question especially relevant for the synthetic-to-real-world domain change is whether domain adaptation also introduces domain generalization to unknown real world domains.
Since the real world target domain is comprised of a nearly infinite amount of subdomains, domain adaptation to each of them is infeasible.
Hence domain adaptation approaches must introduce domain generalization to many real world sub-domains. 
This topic is seldoml addressed in the current unsupervised domain adaptation research.\par
Connected to this question is what information or knowledge can be transferred by unsupervised domain adaptation methods from the source to the target domain.
Given, e.g., synthetic data of street scenes under rainy weather conditions, are unsupervised domain adaptation approaches capable of transferring the explicit knowledge about these scenes to the real world?
This question is seldom analyzed in current research works.
Domain adaptation approaches and papers are optimized to increase the general $\mathrm{mIoU}$ metric 
on generic domain changes such as from GTA5 to Cityscapes. 
We hence suggest two improvements. 
On the one hand, analyzing what kind of knowledge can be transferred requires datasets that offer meta tags about subdomains. 
On the other hand more sophisticated metrics than the $\mathrm{mIoU}$ are required.  

Closely related is the question of how realistic the GTA5- or SYNTHIA-to-Cityscapes domain change still is. 
Considering that current simulation engines can generate more realistic data than the synthetic datasets GTA5 and SYNTHIA, the task of unsupervised domain adaptation from synthetic to real has changed.
This probably influences the methods that could be used, and we, therefore advocate the creation of new benchmarks.\par
All UDA papers discussed in this survey deal with closed-set adaptation (see Section \ref{subsec:related_methods}) meaning both source and target domain have the same classes. 
However, for real large-scale application, it is likely that the classes between source and target are distinct, e.g., the target domain contains the class "E-Scooter" while the source does not.
The performance of the researched closed-set UDA methods for this setting is unclear. 
A stronger focus on open-set adaptation can further simplify the application for real settings.\par
Training time and the number of iterations are rarely addressed in most approaches.
Since training is relatively cheap for scenarios where the source and target domain are comprised of small datasets 
this issue is of a lesser importance. Such scenarios are, e.g., given in the current scientific benchmarks.
In contrast, real autonomous driving datasets are of a large scale and their size might increase even further in the future. 
Hence this aspect is essential and should be considered when judging a method.\par
So far, the presented benchmarks deal with adapting large-scale models only applicable in real-time, given powerful hardware.
UDA approaches should consider presenting a third knowledge distillation step showing how the learned target domain model could be used to provide a real-time capable model.\par
Finally, one of the most critical questions for unsupervised domain adaptation is its relationship with semi-supervised domain adaptation.
Semi-supervised domain adaptation might be an efficient trade-off between the cost of labeling and the performance on the target domain. 
When looking at the highly relevant synthetic-to-real domain change in autonomous driving, we already showed that no UDA method achieves equal or better performance than supervised training on the target domain.
Given that we can even achieve equal or better performance by labeling only a few images of the target domain, the question arises of where and when to apply unsupervised domain adaptation compared to semi-supervised methods.

\subsection{Further Research Directions}
In contrast to the recent performance progress, particularly the step obtained by the vision transformer architectures, the domain performance gap is still significant.
Because oracle performance is mostly not reported, it is impossible to accurately quantify that remaining gap.
However, an estimate between $\approx$ 5-10 \% $\mathrm{mIoU}$ for both CNN and vision transformer architectures is realistic. 
This purely performance-based assessment and the previously discussed issues show that the demand for research in domain-robust deep neural networks is still great.
The aspects which we will discuss in the following as future research go beyond the perspective of simply closing the performance gap for the synthetic-to-real domain gap. 
\subsubsection{Standardization of Evaluation}
As described in Section \ref{discuss:train_issues}, the current evaluation settings are based on similar architectures and the same datasets.
However, they differ in several other aspects with an unclear impact on the reported performance. 
That is not a clean scientific standard and particularly questionable for comparisons where less than one percent can be crucial for a new SOTA performance.
As a first step and necessary basis work for further research, we highly suggest standardizing the evaluation protocols.\par
Crucial points like the resolution, augmentations, architecture, dataset split, reporting standard, and statistically valid evaluation must be the same for all works and are easy to standardize.
Other points like checkpoint selection, how to treat the labeled validation set, and a standard source training setting instead might need to be the objective of a scientific discussion to find a common standard.
\subsubsection{Metrics \& Domain Shift Analysis}
As shown in Section \ref{subsec:metrics}, the variety of employed metrics or assessment tools is limited, directly impacting the acquired knowledge of the evaluations. 
More in-depth insights and a better understanding of the actual underlying mechanism within the network would further push the development of methods. 
These may better address the occurring  domain shifts the more knowledge exists. One possible direction could be the broader utilization of feature visualization tools, e.g., t-SNE~\cite{van2008visualizing}, to evaluate the alignment of the two domains. 
Another option is to develop new metrics to quantify the alignment in feature space.
With this survey, we would like to encourage researchers of future UDA works to use a more diverse set of evaluation tools without diminishing the importance of performance-based metrics like $\mathrm{mIoU}$.\par
The major focus of UDA research in the past years was reducing the performance gap.
To the best of our knowledge, no works so far focus on analyzing the network behavior under domain shift.
There are many important questions only answered at maximum implicitly by existing works: Which classes mostly affect the domain shift? Which factors (style, semantic content, class distribution) are most difficult for domain alignment? How do different network architectures respond under domain shift? Can we find a general metric for network generalization capabilities? 
Focusing more on understanding and analysis by answering these questions could be a valuable contribution to future UDA works.
\subsubsection{New Network Architectures}
As described, both \network{VGG-16} and \network{ResNet-101} were the defacto standard backbones used in UDA research for several years.
Vision transformer architectures recently gained attention in UDA research and reached new SOTA performances (see Section \ref{subsec:visiontransformers}). The full potential of these architectures now needs to be exploited after first works with promising results. 
Therefore utilizing and researching vision transformers as a new architecture type is one of the promising future research trends. 
For researchers, we recommend including vision transformer networks like DAFormer \cite{Hoyer2022DAFormer} in future publications.\par
However, it remains unclear at the current point if vision transformer networks will close the performance domain gap and how strong they perform for other settings such as domain generalization. 
For this reason another interesting research direction can be the development of a real next generation deep learning method with stronger and more human-like generalization capabilities than the vision transformer and more fundamental architectural changes as it was suggested by Marcus et al.\ \cite{marcus2018deep}. 
For this reason, another interesting research direction can be developing a real next-generation deep learning method with stronger and more human-like generalization capabilities than the vision transformer and more fundamental architectural changes, as it was suggested by Marcus et al \cite{marcus2018deep}.
The complex hybrid methods show that CNNs, by default, do not have strong generalization capabilities and therefore require much effort and additional methods (see Section \ref{subsec:methodical_reflection}) to obtain a better generalization or adaptation. 
This might be a strong motivation to push research from the methodic to the architecture level to provide much better generalization capabilities with a new generation of deep networks.
This could align with the research of other areas such as robustness and adversarial stability that also may benefit from increased generalization.
\subsubsection{Large-Scale Domain Adaptation}
There are no works that research large-scale unsupervised domain adaptation, as would be the case for industrial applications.
Instead, for example, a target dataset is utilized, which is relatively old and contains a small number of images, significantly smaller than more recent datasets like A2D2 \cite{geyer2020a2d2}, which would be an exciting  addition.
The focus of the works in this survey is clearly a scientific benchmarking setting which is unquestionably  necessary. 
However, more research is necessary to use these algorithms for real applications because the field of applicability of UDA algorithms still needs to be completed.
One step in this direction can be open set domain adaptation as described in Section \ref{subsec:related_methods}, where the target classes are distinct from the source classes and which might be a common problem for industrial applications across different countries. 
In Section \ref{subsec:methodical_reflection},  we have shown that already small settings changes can lead to severe performance decreases, so research needs to exploit how these algorithms can be transferred into real industrial settings.\par
There is currently a strong focus in research on camera-based domain adaptation, while some works tackle domain adaptation for lidar sensors \cite{triess2021survey}. 
However, from a recent point of knowledge, an entirely autonomous driving stack will come with some kind of sensor fusion that combines the camera information with the inputs from other sensors like LiDAR and RaDAR.
To our knowledge, there are currently no works covering the impact of the domain shift on the system level, which means for the entire perception system of an autonomous vehicle. Researching this question and proposing new methods to tackle the domain shift on a perception system level would be a promising and valuable future research direction. 
\subsubsection{Domain Generalization}
Our survey focuses on unsupervised domain adaptation where unlabeled data of the target domain is available to perform adaptation.
However, we must question how feasible this setting is outside a fixed research context in a large-scale industrial setting. 
The data collection might not be the largest problem since an intelligent data collection mechanism may help, depending on the application.
Data utilization for large-scale applications such as the automotive industry can lead to severe problems. 
The pure number of domains a network needs to be adapted to will cost a lot of effort in terms of computational and human resources. 
Also the formal safety verification that the networks fulfill the safety standards in all domains can be time-intensive. 
It is unlikely that the current generation of DNNs, including vision transformers, can perform well on a very large range of domains in parallel, even with good adaptation.\par
For these reasons, domain generalization can be even more promising as a future research direction because the adaptation step can be left out, making large-scale adaptation easier under the assumption that strong generalization methods are available.
This would have the major advantage that the network does not need to be adapted to every single domain, but we obtain a model that generalizes well across all or most unknown domains.
However, these methods are still missing and could be a promising research direction for their development.
Since it is unclear if current models are powerful enough for such learning strategies, this could be connected to the research direction of novel architectures.

\section{Conclusions}
\label{sec:conclusions}
In this work, we have given the most thorough review of the highly relevant and fastly evolving field of unsupervised domain adaptation for semantic segmentation. 
We have categorized and explained the ideas and methods of the vast majority of approaches that were published in this field. 
We have created a unique knowledge base that provides the reader with a comprehensive overview of the field and an extensive quantitative comparison of the approaches.
There was strong methodical progress over the past years in UDA research, successively decreasing the domain gap.
Hybrid adaptation methods, a combination of multiple standalone methods, can be highlighted as the currently most sophisticated way for UDA and the carrier of the latest SOTA performances.
Recently also, vision transformer architectures raised the SOTA performances into new dimensions.\par
However, the domain gap has yet to be closed. 
For this reason, the scope and claim of this survey were not only the categorization and the description of recent approaches but also to analyze the current status of the research critically.
Following this idea, we pointed out several issues in UDA research, like very complex approaches, bad generalization to other settings, missing strict standards for benchmarking, limited evaluation metrics, and more points.\par
Based on this critical analysis, we eventually recommend promising future directions.
Here we include aspects like new common standards, new architectures, or an application for real large-scale industrial settings. Going beyond the scope of classic UDA research, we discuss new directions like system-level adaptation, symbolic deep learning, and domain generalization. 
By doing so, we gave new impulses for domain adaptation and believe to have facilitated research further.

\bibliographystyle{IEEEtran}
\bibliography{bibliography_clean}

\begin{thebibliography}{100}
\providecommand{\url}[1]{#1}
\csname url@samestyle\endcsname
\providecommand{\newblock}{\relax}
\providecommand{\bibinfo}[2]{#2}
\providecommand{\BIBentrySTDinterwordspacing}{\spaceskip=0pt\relax}
\providecommand{\BIBentryALTinterwordstretchfactor}{4}
\providecommand{\BIBentryALTinterwordspacing}{\spaceskip=\fontdimen2\font plus
\BIBentryALTinterwordstretchfactor\fontdimen3\font minus
  \fontdimen4\font\relax}
\providecommand{\BIBforeignlanguage}[2]{{%
\expandafter\ifx\csname l@#1\endcsname\relax
\typeout{** WARNING: IEEEtran.bst: No hyphenation pattern has been}%
\typeout{** loaded for the language `#1'. Using the pattern for}%
\typeout{** the default language instead.}%
\else
\language=\csname l@#1\endcsname
\fi
#2}}
\providecommand{\BIBdecl}{\relax}
\BIBdecl

\bibitem{fingscheidt_dnndataautomateddriving}
\BIBentryALTinterwordspacing
T.~Fing\-scheidt, H.~Gottschalk, and S.~Houben, Eds., \emph{{Deep Neural
  Networks and Data for Automated Driving: Robustness, Uncertainty
  Quantification, and Insights Towards Safety}}.\hskip 1em plus 0.5em minus
  0.4em\relax Cham: {S}pringer {N}ature, 2022. [Online]. Available:
  \url{https://library.oapen.org/handle/20.500.12657/57375}
\BIBentrySTDinterwordspacing

\bibitem{Houben2022}
S.~Houben, S.~Abrecht, M.~Akila, A.~B{\"a}r, F.~Brockherde, P.~Feifel,
  T.~Fingscheidt, S.~S. Gannamaneni, S.~E. Ghobadi, A.~Hammam, A.~Haselhoff,
  F.~Hauser, C.~Heinzemann, M.~Hoffmann, N.~Kapoor, F.~Kappel, M.~Klingner,
  J.~Kronenberger, F.~K{\"u}ppers, J.~L{\"o}hdefink, M.~Mlynarski, M.~Mock,
  F.~Mualla, S.~Pavlitskaya, M.~Poretschkin, A.~Pohl, V.~Ravi-Kumar,
  J.~Rosenzweig, M.~Rottmann, S.~R{\"u}ping, T.~S{\"a}mann, J.~D. Schneider,
  E.~Schulz, G.~Schwalbe, J.~Sicking, T.~Srivastava, S.~Varghese, M.~Weber,
  S.~Wirkert, T.~Wirtz, and M.~Woehrle, \emph{{Inspect, Understand, Overcome: A
  Survey of Practical Methods for AI Safety}}.\hskip 1em plus 0.5em minus
  0.4em\relax Cham: Springer International Publishing, 2022, pp. 3--78.

\bibitem{krizhevsky2017imagenet}
A.~Krizhevsky, I.~Sutskever, and G.~E. Hinton, ``{ImageNet Classification With
  Deep Convolutional Neural Networks},'' \emph{Communications of the ACM},
  vol.~60, no.~6, pp. 84--90, 2017.

\bibitem{akhtar2021advances}
N.~Akhtar, A.~Mian, N.~Kardan, and M.~Shah, ``{Advances in Adversarial Attacks
  and Defenses in Computer Vision: A Survey},'' \emph{IEEE Access}, vol.~9, pp.
  155\,161--155\,196, 2021.

\bibitem{jaiswal2020survey}
A.~Jaiswal, A.~R. Babu, M.~Z. Zadeh, D.~Banerjee, and F.~Makedon, ``{A Survey
  on Contrastive Self-Supervised Learning},'' \emph{Technologies}, vol.~9,
  no.~1, pp. 1--22, 2020.

\bibitem{Cordts2016}
M.~Cordts, M.~Omran, S.~Ramos, T.~Rehfeld, M.~Enzweiler, R.~Benenson,
  U.~Franke, S.~Roth, and B.~Schiele, ``{The Cityscapes Dataset for Semantic
  Urban Scene Understanding},'' in \emph{Proc. of CVPR}, Las Vegas, NV, USA,
  Jun. 2016, pp. 3213--3223.

\bibitem{Meyer2018mapless}
A.~Meyer, N.~O. Salscheider, P.~F. Orzechowski, and C.~Stiller, ``{Deep
  Semantic Lane Segmentation for Mapless Driving},'' in \emph{Proc. of IROS},
  Madrid, Spain, Oct. 2018, pp. 869--875.

\bibitem{Plachetka2020}
C.~Plachetka, N.~Maier, J.~Fricke, J.-A. Termöhlen, and T.~Fing\-scheidt,
  ``{Terminology and Analysis of Map Deviations in Urban Domains: Towards
  Dependability for HD Maps in Automated Vehicles},'' in \emph{Proc. of IV},
  Las Vegas, NV, USA, Oct. 2020, pp. 63--70.

\bibitem{Plachetka2021}
C.~Plachetka, J.~Fricke, M.~Klingner, and T.~Fing\-scheidt, ``{DNN-Based
  Recognition of Pole-Like Objects in LiDAR Point Clouds},'' in \emph{Proc. of
  ITSC}, Montreal, QC, Canada, Sep. 2021, pp. 2889--2896.

\bibitem{Plachetka2022}
C.~Plachetka, B.~Sertolli, J.~Fricke, M.~Klingner, and T.~Fing\-scheidt,
  ``{3DHD CityScenes: High-Definition Maps in High-Density Point Clouds},'' in
  \emph{Proc. of ITSC}, Macau, China, Oct. 2022, pp. 627--634.

\bibitem{oza2021unsupervised}
P.~Oza, V.~A. Sindagi, V.~VS, and V.~M. Patel, ``{Unsupervised Domain
  Adaptation of Object Detectors: A Survey},'' \emph{arXiv:2105.13502}, 2021.

\bibitem{hoffman2016fcns}
J.~Hoffman, D.~Wang, F.~Yu, and T.~Darrell, ``{FCNs in the Wild: Pixel-Level
  Adversarial and Constraint-Based Adaptation},'' \emph{arXiv:1612.02649},
  2016.

\bibitem{wilson2020survey}
G.~Wilson and D.~J. Cook, ``{A Survey of Unsupervised Deep Domain
  Adaptation},'' \emph{ACM Transactions on Intelligent Systems and Technology},
  vol.~11, no.~5, pp. 1--46, 2020.

\bibitem{zhang2021survey}
Y.~Zhang, ``{A Survey of Unsupervised Domain Adaptation for Visual
  Recognition},'' \emph{arXiv:2112.06745}, 2021.

\bibitem{wang2018deep}
M.~Wang and W.~Deng, ``{Deep Visual Domain Adaptation: A Survey},''
  \emph{Neurocomputing}, vol. 312, pp. 135--153, 2018.

\bibitem{toldo2020review}
M.~Toldo, A.~Maracani, U.~Michieli, and P.~Zanuttigh, ``{Unsupervised Domain
  Adaptation in Semantic Segmentation: A Review},'' \emph{Technologies},
  vol.~8, no.~2, pp. 1--35, 2020.

\bibitem{csurka2021unsupervised}
G.~Csurka, R.~Volpi, and B.~Chidlovskii, ``{Unsupervised Domain Adaptation for
  Semantic Image Segmentation: A Comprehensive Survey},''
  \emph{arXiv:2112.03241}, 2021.

\bibitem{wang2020alleviating}
Z.~Wang, Y.~Wei, R.~Feris, J.~Xiong, W.-M. Hwu, T.~S. Huang, and H.~Shi,
  ``{Alleviating Semantic-Level Shift: A Semi-Supervised Domain Adaptation
  Method for Semantic Segmentation},'' in \emph{Proc. of CVPR - Workshops},
  Seattle, WA, USA, Jun. 2020, pp. 936--937.

\bibitem{mutze2022semi}
A.~M{\"u}tze, M.~Rottmann, and H.~Gottschalk, ``{Semi-Supervised Domain
  Adaptation With CycleGAN Guided by a Downstream Task Loss},''
  \emph{arXiv:2208.08815}, 2022.

\bibitem{chen2021semi}
Y.~Chen, X.~Ouyang, K.~Zhu, and G.~Agam, ``{Semi-Supervised Domain Adaptation
  for Semantic Segmentation},'' \emph{arXiv:2110.10639}, 2021.

\bibitem{chen2021semi2}
S.~Chen, X.~Jia, J.~He, Y.~Shi, and J.~Liu, ``{Semi-Supervised Domain
  Adaptation Based on Dual-Level Domain Mixing for Semantic Segmentation},'' in
  \emph{Proc. of CVPR}, virtual, Jun. 2021, pp. 11\,018--11\,027.

\bibitem{Hanselmann2021weakly}
N.~Hanselmann, N.~Schneider, B.~Ortelt, and A.~Geiger, ``{Learning Cascaded
  Detection Tasks With Weakly-Supervised Domain Adaptation},'' in \emph{Proc.
  of IV}, virtual, Jul. 2021, pp. 532--539.

\bibitem{paul2020domain}
S.~Paul, Y.-H. Tsai, S.~Schulter, A.~K. Roy-Chowdhury, and M.~Chandraker,
  ``{Domain Adaptive Semantic Segmentation Using Weak Labels},''
  \emph{arXiv:2007.15176}, 2020.

\bibitem{Klingner2020c}
M.~Klingner, J.-A. Term\"{o}hlen, J.~Ritterbach, and T.~Fing\-scheidt,
  ``{Unsupervised BatchNorm Adaptation (UBNA): A Domain Adaptation Method for
  Semantic Segmentation Without Using Source Domain Representations},'' in
  \emph{Proc. of WACV - Workshops}, Waikoloa, HI, USA, Jan. 2022, pp. 210--220.

\bibitem{Klingner2020d}
M.~Klingner, M.~Ayache, and T.~Fing\-scheidt, ``{Continual BatchNorm Adaptation
  (CBNA) for Semantic Segmentation},'' \emph{IEEE Transactions on Intelligent
  Transportation Systems}, vol.~23, no.~11, pp. 20\,899--20\,911, 2022.

\bibitem{Termoehlen2021}
J.-A. Term\"{o}hlen, M.~Klingner, L.~J. Brettin, N.~M. Schmidt, and
  T.~Fing\-scheidt, ``{Continual Unsupervised Domain Adaptation for Semantic
  Segmentation by Online Frequency Domain Style Transfer},'' in \emph{Proc. of
  ITSC}, virtual, Sep. 2021, pp. 2881--2888.

\bibitem{Wulfmeier2018}
M.~Wulfmeier, A.~Bewley, and I.~Posner, ``{Incremental Adversarial Domain
  Adaptation for Continually Changing Environments},'' in \emph{Proc. of ICRA},
  Brisbane, Australia, May 2018, pp. 4489--4495.

\bibitem{yue2019domain}
X.~Yue, Y.~Zhang, S.~Zhao, A.~Sangiovanni-Vincentelli, K.~Keutzer, and B.~Gong,
  ``{Domain Randomization and Pyramid Consistency: Simulation-to-Real
  Generalization Without Accessing Target Domain Data},'' in \emph{Proc. of
  ICCV}, Seoul, Korea, Oct. 2019, pp. 2100--2110.

\bibitem{lee2022wildnet}
S.~Lee, H.~Seong, S.~Lee, and E.~Kim, ``{WildNet: Learning Domain Generalized
  Semantic Segmentation From the Wild},'' in \emph{Proc. of CVPR}, New Orleans,
  LA, USA, Jun. 2022, pp. 9936--9946.

\bibitem{choi2021robustnet}
S.~Choi, S.~Jung, H.~Yun, J.~T. Kim, S.~Kim, and J.~Choo, ``{RobustNet:
  Improving Domain Generalization in Urban-Scene Segmentation via Instance
  Selective Whitening},'' in \emph{Proc. of CVPR}, virtual, Jun. 2021, pp.
  11\,580--11\,590.

\bibitem{gong2019dlow}
R.~Gong, W.~Li, Y.~Chen, and L.~V. Gool, ``{DLOW: Domain Flow for Adaptation
  and Generalization},'' in \emph{Proc. of CVPR}, Long Beach, CA, USA, Jun.
  2019, pp. 2477--2486.

\bibitem{uhlemeyer2022towards}
S.~Uhlemeyer, M.~Rottmann, and H.~Gottschalk, ``{Towards Unsupervised Open
  World Semantic Segmentation},'' \emph{arXiv:2201.01073}, 2022.

\bibitem{Kalb2021classdomain}
T.~Kalb, M.~Roschani, M.~Ruf, and J.~Beyerer, ``{Continual Learning for Class-
  And Domain-Incremental Semantic Segmentation},'' in \emph{Proc. of IV},
  virtual, Jul. 2021, pp. 1345--1351.

\bibitem{gong2021tada}
R.~Gong, M.~Danelljan, D.~Dai, W.~Wang, D.~P. Paudel, A.~Chhatkuli, F.~Yu, and
  L.~Van~Gool, ``{TADA: Taxonomy Adaptive Domain Adaptation},''
  \emph{arXiv:2109.04813}, 2021.

\bibitem{kouw2018introduction}
W.~M. Kouw and M.~Loog, ``An introduction to domain adaptation and transfer
  learning,'' \emph{arXiv preprint arXiv:1812.11806}, 2018.

\bibitem{Richter2016}
S.~Richter, V.~Vineet, S.~Roth, and V.~Koltun, ``{Playing for Data: Ground
  Truth From Computer Games},'' in \emph{Proc. of ECCV}, Amsterdam,
  Netherlands, Oct. 2016, pp. 102--118.

\bibitem{ros2016synthia}
G.~Ros, L.~Sellart, J.~Materzynska, D.~Vazquez, and A.~M. Lopez, ``{The Synthia
  Dataset: A Large Collection of Synthetic Images for Semantic Segmentation of
  Urban Scenes},'' in \emph{Proc. of CVPR}, Las Vegas, NV, USA, Jun. 2016, pp.
  3234--3243.

\bibitem{chen2017no}
Y.-H. Chen, W.-Y. Chen, Y.-T. Chen, B.-C. Tsai, Y.-C. Frank~Wang, and M.~Sun,
  ``{No More Discrimination: Cross City Adaptation of Road Scene Segmenters},''
  in \emph{Proc. of ICCV}, Venice, Italy, Oct. 2017, pp. 1992--2001.

\bibitem{Sakaridis2021}
C.~Sakaridis, D.~Dai, and L.~Van~Gool, ``{ACDC: The Adverse Conditions Dataset
  With Correspondences for Semantic Driving Scene Understanding},'' in
  \emph{Proc. of ICCV}, virtual, Oct. 2021, pp. 10\,765--10\,775.

\bibitem{dosovitskiy2017carla}
A.~Dosovitskiy, G.~Ros, F.~Codevilla, A.~Lopez, and V.~Koltun, ``{CARLA: An
  Open Urban Driving Simulator},'' in \emph{Proc. of CoRL}, Mountain View, CA,
  USA, Nov. 2017, pp. 1--16.

\bibitem{Long2015}
J.~Long, E.~Shelhamer, and T.~Darrell, ``{Fully Convolutional Networks for
  Semantic Segmentation},'' in \emph{Proc. of CVPR}, Boston, MA, USA, Jun.
  2015, pp. 3431--3440.

\bibitem{Chen2018}
L.-C. Chen, G.~Papandreou, I.~Kokkinos, K.~Murphy, and A.~L. Yuille,
  ``{DeepLab: Semantic Image Segmentation With Deep Convolutional Nets, Atrous
  Convolution, and Fully Connected CRFs},'' \emph{IEEE Transactions on Pattern
  Analysis and Machine Intelligence}, vol.~40, no.~4, pp. 834--848, Apr. 2017.

\bibitem{chao2021rethinking}
C.-H. Chao, B.-W. Cheng, and C.-Y. Lee, ``{Rethinking Ensemble-Distillation for
  Semantic Segmentation Based Unsupervised Domain Adaptation},''
  \emph{arXiv:2104.14203}, 2021.

\bibitem{toldo2021unsupervised}
M.~Toldo, U.~Michieli, and P.~Zanuttigh, ``{Unsupervised Domain Adaptation in
  Semantic Segmentation via Orthogonal and Clustered Embeddings},'' in
  \emph{Proc. of WACV}, virtual, Jan. 2021, pp. 1358--1368.

\bibitem{ye2020light}
S.~Ye, K.~Wu, M.~Zhou, Y.~Yang, S.~H. Tan, K.~Xu, J.~Song, C.~Bao, and K.~Ma,
  ``{Light-Weight Calibrator: A Separable Component for Unsupervised Domain
  Adaptation},'' in \emph{Proc. of CVPR}, virtual, Jun. 2020, pp.
  13\,736--13\,745.

\bibitem{lee2021dranet}
S.~Lee, S.~Cho, and S.~Im, ``{DRANet: Disentangling Representation and
  Adaptation Networks for Unsupervised Cross-Domain Adaptation},'' in
  \emph{Proc. of CVPR}, virtual, Jun. 2021, pp. 15\,252--15\,261.

\bibitem{Chen2019crdoco}
Y.-C. Chen, Y.-Y. Lin, M.-H. Yang, and J.-B. Huang, ``{CrDoCo: Pixel-Level
  Domain Transfer With Cross-Domain Consistency},'' in \emph{Proc. of CVPR},
  Long Beach, CA, USA, Jun. 2019, pp. 1791--1800.

\bibitem{Saito2018a}
K.~Saito, K.~Watanabe, Y.~Ushiku, and T.~Harada, ``{Maximum Classifier
  Discrepancy for Unsupervised Domain Adaptation},'' in \emph{Proc. of CVPR},
  Salt Lake City, UT, USA, Jun. 2018, pp. 3723--3732.

\bibitem{shan2020semantic}
Y.~Shan, C.~M. Chew, and W.~F. Lu, ``{Semantic-Aware Short Path Adversarial
  Training for Cross-Domain Semantic Segmentation},'' \emph{Neurocomputing},
  vol. 380, pp. 125--132, 2020.

\bibitem{Iqbal2020}
J.~Iqbal and M.~Ali, ``{MLSL: Multi-Level Self-Supervised Learning for Domain
  Adaptation With Spatially Independent and Semantically Consistent
  Labeling},'' in \emph{Proc. of WACV}, Aspen, CO, USA, Mar. 2020, pp.
  1864--1873.

\bibitem{Lian2019}
Q.~Lian, F.~Lv, L.~Duan, and B.~Gong, ``{Constructing Self-Motivated Pyramid
  Curriculums for Cross-Domain Semantic Segmentation: A Non-Adversarial
  Approach},'' in \emph{Proc. of ICCV}, Seoul, Korea, Oct. 2019, pp.
  6758--6767.

\bibitem{Zou2018}
Y.~Zou, Z.~Yu, B.~V.~K. Vijaya~Kumar, and J.~Wang, ``{Unsupervised Domain
  Adaptation for Semantic Segmentation via Class-Balanced Self-Training},'' in
  \emph{Proc. of ECCV}, Munich, Germany, Sep. 2018, pp. 289--305.

\bibitem{Dong2019a}
J.~Dong, Y.~Cong, G.~Sun, and D.~Hou, ``{Semantic-Transferable
  Weakly-Supervised Endoscopic Lesions Segmentation},'' in \emph{Proc. of
  ICCV}, Seoul, Korea, Oct. 2019, pp. 10\,712--10\,721.

\bibitem{li2020variational}
Z.~Li, R.~Togo, T.~Ogawa, and M.~Haseyama, ``{Variational Autoencoder Based
  Unsupervised Domain Adaptation for Semantic Segmentation},'' in \emph{Proc
  of. ICIP}, Abu Dhabi, United Arab Emirates, Oct. 2020, pp. 2426--2430.

\bibitem{Hoyer2022DAFormer}
L.~Hoyer, D.~Dai, and L.~Van~Gool, ``{DAFormer: Improving Network Architectures
  and Training Strategies For Domain-Adaptive Semantic Segmentation},'' in
  \emph{Proc. of CVPR}, New Orleans, LA, USA, Jun. 2022, pp. 9924--9935.

\bibitem{xie2021segformer}
E.~Xie, W.~Wang, Z.~Yu, A.~Anandkumar, J.~M. Alvarez, and P.~Luo, ``{SegFormer:
  Simple and Efficient Design for Semantic Segmentation With Transformers},''
  in \emph{Proc. of NeurIPS}, virtual, Dec. 2021, pp. 12\,077--12\,090.

\bibitem{chen2022smoothing}
R.~Chen, Y.~Rong, S.~Guo, J.~Han, F.~Sun, T.~Xu, and W.~Huang, ``{Smoothing
  Matters: Momentum Transformer for Domain Adaptive Semantic Segmentation},''
  \emph{arXiv:2203.07988}, 2022.

\bibitem{liu2021swin}
Z.~Liu, Y.~Lin, Y.~Cao, H.~Hu, Y.~Wei, Z.~Zhang, S.~Lin, and B.~Guo, ``{Swin
  Transformer: Hierarchical Vision Transformer Using Shifted Windows},'' in
  \emph{Proc. of ICCV}, virtual, Oct. 2021, pp. 10\,012--10\,022.

\bibitem{toldo2020unsupervised}
M.~Toldo, U.~Michieli, G.~Agresti, and P.~Zanuttigh, ``{Unsupervised Domain
  Adaptation for Mobile Semantic Segmentation Based on Cycle Consistency and
  Feature Alignment},'' \emph{Image and Vision Computing}, vol.~95, pp.
  103\,889--103\,899, 2020.

\bibitem{dundar2018a}
A.~Dundar, M.-Y. Liu, T.-C. Wang, J.~Zedlewski, and J.~Kautz, ``{Domain
  Stylization: A Strong, Simple Baseline for Synthetic to Real Image Domain
  Adaptation},'' \emph{arXiv:1807.09384}, Jul. 2018.

\bibitem{gkitsas2019restyling}
V.~Gkitsas, A.~Karakottas, N.~Zioulis, D.~Zarpalas, and P.~Daras, ``{Restyling
  Data: Application to Unsupervised Domain Adaptation},''
  \emph{arXiv:1909.10900}, 2019.

\bibitem{li2020generating}
R.~Li, W.~Cao, S.~Wu, and H.-S. Wong, ``{Generating Target Image-Label Pairs
  for Unsupervised Domain Adaptation},'' \emph{IEEE Transactions on Image
  Processing}, vol.~29, pp. 7997--8011, 2020.

\bibitem{lin2019adapting}
Y.-X. Lin, D.~S. Tan, W.-H. Cheng, and K.-L. Hua, ``{Adapting Semantic
  Segmentation of Urban Scenes via Mask-Aware Gated Discriminator},'' in
  \emph{Proc. of ICME}, Shanghai, China, Jul. 2019, pp. 218--223.

\bibitem{Zhu2017}
J.-Y. Zhu, T.~Park, P.~Isola, and A.~A. Efros, ``{Unpaired Image-to-Image
  Translation Using Cycle-Consistent Adversarial Networks},'' in \emph{Proc. of
  ICCV}, Venice, Italy, Oct. 2017, pp. 2223--2232.

\bibitem{Huang2017b}
X.~Huang and S.~Belongie, ``{Arbitrary Style Transfer in Real-Time With
  Adaptive Instance Normalization},'' in \emph{Proc. of ICCV}, Venice, Italy,
  Oct. 2017, pp. 1501--1510.

\bibitem{huang2021cross}
J.~Huang, D.~Guan, A.~Xiao, and S.~Lu, ``{Cross-View Regularization for Domain
  Adaptive Panoptic Segmentation},'' \emph{arXiv:2103.02584}, 2021.

\bibitem{ma2021coarse}
H.~Ma, X.~Lin, Z.~Wu, and Y.~Yu, ``{Coarse-to-Fine Domain Adaptive Semantic
  Segmentation With Photometric Alignment and Category-Center
  Regularization},'' in \emph{Proc. of CVPR}, 2021, pp. 4051--4060.

\bibitem{lu2022bidirectional}
Y.~Lu, Y.~Luo, L.~Zhang, Z.~Li, Y.~Yang, and J.~Xiao, ``{Bidirectional
  Self-Training With Multiple Anisotropic Prototypes for Domain Adaptive
  Semantic Segmentation},'' \emph{arXiv:2204.07730}, pp. 1--11, Apr. 2022.

\bibitem{yang2020fda}
Y.~Yang and S.~Soatto, ``{FDA: Fourier Domain Adaptation for Semantic
  Segmentation},'' in \emph{Proc. of CVPR}, Seattle, WA, USA, Jun. 2020, pp.
  4085--4095.

\bibitem{zhang2021spectral}
J.~Zhang, J.~Huang, and S.~Lu, ``{Spectral Unsupervised Domain Adaptation for
  Visual Recognition},'' \emph{arXiv:2106.06112}, 2021.

\bibitem{tranheden2021dacs}
W.~Tranheden, V.~Olsson, J.~Pinto, and L.~Svensson, ``{DACS: Domain Adaptation
  via Cross-Domain Mixed Sampling},'' in \emph{Proc. of WACV}, Waikoloa, HI,
  USA, Jan. 2021, pp. 1379--1389.

\bibitem{Gao2021DSP}
L.~Gao, J.~Zhang, L.~Zhang, and D.~Tao, ``{DSP: Dual Soft-Paste for
  Unsupervised Domain Adaptive Semantic Segmentation},'' in \emph{Proc. of ACM
  International Conference on Multimedia}, Chengdu, China, Oct. 2021, pp.
  2825--2833.

\bibitem{melas2021pixmatch}
L.~Melas-Kyriazi and A.~K. Manrai, ``{PixMatch: Unsupervised Domain Adaptation
  via Pixelwise Consistency Training},'' in \emph{Proc. of CVPR}, virtual, Jun.
  2021, pp. 12\,435--12\,445.

\bibitem{liu2021bapa}
Y.~Liu, J.~Deng, X.~Gao, W.~Li, and L.~Duan, ``{BAPA-Net: Boundary Adaptation
  and Prototype Alignment for Cross-Domain Semantic Segmentation},'' in
  \emph{Proc. of ICCV}, virtual, Oct. 2021, pp. 8801--8811.

\bibitem{zhou2021domain}
Q.~Zhou, C.~Zhuang, R.~Yi, X.~Lu, and L.~Ma, ``{Domain Adaptive Semantic
  Segmentation via Regional Contrastive Consistency Regularization},''
  \emph{arXiv}, no. 2110.05170, pp. 1--7, Oct. 2021.

\bibitem{zhou2021context}
Q.~Zhou, Z.~Feng, Q.~Gu, J.~Pang, G.~Cheng, X.~Lu, J.~Shi, and L.~Ma,
  ``{Context-Aware Mixup for Domain Adaptive Semantic Segmentation},''
  \emph{arXiv:2108.03557}, 2021.

\bibitem{wang2021domain}
Q.~Wang, D.~Dai, L.~Hoyer, L.~Van~Gool, and O.~Fink, ``{Domain Adaptive
  Semantic Segmentation With Self-Supervised Depth Estimation},'' in
  \emph{Proc. of ICCV}, virtual, Oct. 2021, pp. 8515--8525.

\bibitem{huo2022domain}
X.~Huo, L.~Xie, H.~Hu, W.~Zhou, H.~Li, and Q.~Tian, ``{Domain-Agnostic Prior
  for Transfer Semantic Segmentation},'' \emph{arXiv:2204.02684}, 2022.

\bibitem{kim2020learning}
M.~Kim and H.~Byun, ``{Learning Texture Invariant Representation for Domain
  Adaptation of Semantic Segmentation},'' in \emph{Proc. of CVPR}, Seattle, WA,
  USA, Jun. 2020, pp. 12\,975--12\,984.

\bibitem{huang2021rda}
J.~Huang, D.~Guan, A.~Xiao, and S.~Lu, ``{RDA: Robust Domain Adaptation via
  Fourier Adversarial Attacking},'' in \emph{Proc. of ICCV}, virtual, Oct.
  2021, pp. 8988--8999.

\bibitem{zhang2022unsupervised}
F.~Zhang, V.~Koltun, P.~Torr, R.~Ranftl, and S.~R. Richter, ``{Unsupervised
  Contrastive Domain Adaptation for Semantic Segmentation},''
  \emph{arXiv:2204.08399}, 2022.

\bibitem{zhou2020uncertainty}
Q.~Zhou, Z.~Feng, G.~Cheng, X.~Tan, J.~Shi, and L.~Ma, ``{Uncertainty-Aware
  Consistency Regularization for Cross-Domain Semantic Segmentation},''
  \emph{arXiv:2004.08878}, 2020.

\bibitem{musto2020semantically}
L.~Musto and A.~Zinelli, ``{Semantically Adaptive Image-to-Image Translation
  for Domain Adaptation of Semantic Segmentation},'' \emph{arXiv:2009.01166},
  2020.

\bibitem{luo2020adversarial}
Y.~Luo, P.~Liu, T.~Guan, J.~Yu, and Y.~Yang, ``{Adversarial Style Mining for
  One-Shot Unsupervised Domain Adaptation},'' in \emph{Proc. of NeurIPS},
  virtual, Dec. 2020, pp. 20\,612--20\,623.

\bibitem{choi2019self}
J.~Choi, T.~Kim, and C.~Kim, ``{Self-Ensembling With GAN-Based Data
  Augmentation for Domain Adaptation in Semantic Segmentation},'' in
  \emph{Proc. of ICCV}, Seoul, Korea, Oct. 2019, pp. 6830--6840.

\bibitem{wang2020consistency}
K.~Wang, C.~Yang, and M.~Betke, ``{Consistency Regularization With
  High-Dimensional Non-Adversarial Source-Guided Perturbation for Unsupervised
  Domain Adaptation in Segmentation},'' \emph{arXiv:2009.08610}, 2020.

\bibitem{tang2021unsupervised}
S.~Tang, P.~Tang, Y.~Gong, Z.~Ma, and M.~Xie, ``{Unsupervised Domain Adaptation
  via Coarse-to-Fine Feature Alignment Method Using Contrastive Learning},''
  \emph{arXiv:2103.12371}, 2021.

\bibitem{Ulyanov2016}
D.~Ulyanov, A.~Vedaldi, and V.~Lempitsky, ``{Instance Normalization: The
  Missing Ingredient for Fast Stylization},'' \emph{arXiv:1607.08022}, Jul.
  2016.

\bibitem{wu2018dcan}
Z.~Wu, X.~Han, Y.-L. Lin, M.~G. Uzunbas, T.~Goldstein, S.~N. Lim, and L.~S.
  Davis, ``{DCAN: Dual Channel-Wise Alignment Networks for Unsupervised Scene
  Adaptation},'' in \emph{Proc. of ECCV}, Munich, Germany, Sep. 2018, pp.
  535--552.

\bibitem{chen2019learning}
Y.~Chen, W.~Li, X.~Chen, and L.~V. Gool, ``{Learning Semantic Segmentation From
  Synthetic Data: A Geometrically Guided Input-Output Adaptation Approach},''
  in \emph{Proc. of CVPR}, Long Beach, CA, USA, Jun. 2019, pp. 1841--1850.

\bibitem{sankaranarayanan2018learning}
S.~Sankaranarayanan, Y.~Balaji, A.~Jain, S.~N. Lim, and R.~Chellappa,
  ``{Learning From Synthetic Data: Addressing Domain Shift for Semantic
  Segmentation},'' in \emph{Proc. of CVPR}, Salt Lake City, UT, USA, Jun. 2018,
  pp. 3752--3761.

\bibitem{chang2019all}
W.-L. Chang, H.-P. Wang, W.-H. Peng, and W.-C. Chiu, ``{All About Structure:
  Adapting Structural Information Across Domains for Boosting Semantic
  Segmentation},'' in \emph{Proc. of CVPR}, Long Beach, CA, USA, Jun. 2019, pp.
  1900--1909.

\bibitem{yang2020phase}
Y.~Yang, D.~Lao, G.~Sundaramoorthi, and S.~Soatto, ``{Phase Consistent
  Ecological Domain Adaptation},'' in \emph{Proc. of CVPR}, Seattle, WA, USA,
  Jun. 2020, pp. 9011--9020.

\bibitem{song2020learning}
L.~Song, Y.~Xu, L.~Zhang, B.~Du, Q.~Zhang, and X.~Wang, ``{Learning From
  Synthetic Images via Active Pseudo-Labeling},'' \emph{IEEE Transactions on
  Image Processing}, vol.~29, pp. 6452--6465, 2020.

\bibitem{hoffman2018cycada}
J.~Hoffman, E.~Tzeng, T.~Park, J.-Y. Zhu, P.~Isola, K.~Saenko, A.~Efros, and
  T.~Darrell, ``{CyCADA: Cycle-Consistent Adversarial Domain Adaptation},'' in
  \emph{Proc. of ICML}, Stockholm, Sweden, Jul. 2018, pp. 1989--1998.

\bibitem{yang2021context}
J.~Yang, W.~An, C.~Yan, P.~Zhao, and J.~Huang, ``{Context-Aware Domain
  Adaptation in Semantic Segmentation},'' in \emph{Proc. of WACV}, virtual,
  Jan. 2021, pp. 514--524.

\bibitem{ramirez2018exploiting}
P.~Z. Ramirez, A.~Tonioni, and L.~Di~Stefano, ``{Exploiting Semantics in
  Adversarial Training for Image-Level Domain Adaptation},'' in \emph{Proc. of
  IPAS)}, Sophia Antipolis, France, Dec. 2018, pp. 49--54.

\bibitem{li2019bidirectional}
Y.~Li, L.~Yuan, and N.~Vasconcelos, ``{Bidirectional Learning for Domain
  Adaptation of Semantic Segmentation},'' in \emph{Proc. of CVPR}, Long Beach,
  CA, USA, Jun. 2019, pp. 6936--6945.

\bibitem{kim2020cross}
M.~Kim, S.~Joung, S.~Kim, J.~Park, I.-J. Kim, and K.~Sohn, ``{Cross-Domain
  Grouping and Alignment for Domain Adaptive Semantic Segmentation},''
  \emph{arXiv:2012.08226}, 2020.

\bibitem{zhou2020affinity}
W.~Zhou, Y.~Wang, J.~Chu, J.~Yang, X.~Bai, and Y.~Xu, ``{Affinity Space
  Adaptation for Semantic Segmentation Across Domains},'' \emph{IEEE
  Transactions on Image Processing}, vol.~30, pp. 2549--2561, 2021.

\bibitem{wang2020differential}
Z.~Wang, M.~Yu, Y.~Wei, R.~Feris, J.~Xiong, W.~Hwu, T.~S. Huang, and H.~Shi,
  ``{Differential Treatment for Stuff and Things: A Simple Unsupervised Domain
  Adaptation Method for Semantic Segmentation},'' in \emph{Proc. of CVPR},
  Seattle, WA, USA, Jun. 2020, pp. 12\,635--12\,644.

\bibitem{yang2020label}
J.~Yang, W.~An, S.~Wang, X.~Zhu, C.~Yan, and J.~Huang, ``{Label-Driven
  Reconstruction for Domain Adaptation in Semantic Segmentation},'' in
  \emph{Proc. of ECCV}, Glasgow, UK, Aug. 2020, pp. 480--498.

\bibitem{lee2018spigan}
K.-H. Lee, G.~Ros, J.~Li, and A.~Gaidon, ``{SPIGAN: Privileged Adversarial
  Learning From Simulation},'' in \emph{Proc. of ICLR}, New Orleans, LA, USA,
  Apr. 2019, pp. 1--14.

\bibitem{lee2020unsupervised}
S.~Lee, J.~Hyun, H.~Seong, and E.~Kim, ``{Unsupervised Domain Adaptation for
  Semantic Segmentation by Content Transfer},'' \emph{arXiv:2012.12545}, 2020.

\bibitem{chung2021maximizing}
I.~Chung, D.~Kim, and N.~Kwak, ``{Maximizing Cosine Similarity Between Spatial
  Features for Unsupervised Domain Adaptation in Semantic Segmentation},''
  \emph{arXiv:2102.13002}, 2021.

\bibitem{cheng2021dual}
Y.~Cheng, F.~Wei, J.~Bao, D.~Chen, F.~Wen, and W.~Zhang, ``{Dual Path Learning
  for Domain Adaptation of Semantic Segmentation},'' in \emph{Proc. of ICCV},
  virtual, Oct. 2021, pp. 9082--9091.

\bibitem{xu2021self}
Y.~Xu, F.~He, B.~Du, L.~Zhang, and D.~Tao, ``{Self-Ensembling GAN for
  Cross-Domain Semantic Segmentation},'' \emph{arXiv:2112.07999}, 2021.

\bibitem{gao2021addressing}
L.~Gao, L.~Zhang, and Q.~Zhang, ``{Addressing Domain Gap via Content Invariant
  Representation for Semantic Segmentation},'' in \emph{Proc. of AAAI},
  vol.~35, no.~9, 2021, pp. 7528--7536.

\bibitem{dong2021and}
J.~Dong, Y.~Cong, G.~Sun, Z.~Fang, and Z.~Ding, ``{Where and How to Transfer:
  Knowledge Aggregation-Induced Transferability Perception for Unsupervised
  Domain Adaptation},'' \emph{IEEE Transactions on Pattern Analysis and Machine
  Intelligence}, 2021.

\bibitem{saporta2020esl}
A.~Saporta, T.-H. Vu, M.~Cord, and P.~P{\'e}rez, ``{ESL: Entropy-Guided
  Self-Supervised Learning for Domain Adaptation in Semantic Segmentation},''
  \emph{arXiv:2006.08658}, 2020.

\bibitem{li2020unsupervised}
Z.~Li, R.~Togo, T.~Ogawa, and M.~Haseyama, ``{Unsupervised Domain Adaptation
  for Semantic Segmentation With Symmetric Adaptation Consistency},'' in
  \emph{Proc. of ICASSP}, Barcelona, Spain, May 2020, pp. 2263--2267.

\bibitem{cai2020exploiting}
B.~Cai, H.~Fu, R.~Jia, B.~Zhao, H.~Li, and Y.~Xu, ``{Exploiting Diverse
  Characteristics and Adversarial Ambivalence for Domain Adaptive
  Segmentation},'' \emph{arXiv:2012.05608}, 2020.

\bibitem{li2020simplified}
R.~Li, W.~Cao, Q.~Jiao, S.~Wu, and H.-S. Wong, ``{Simplified Unsupervised Image
  Translation for Semantic Segmentation Adaptation},'' \emph{Pattern
  Recognition}, vol. 105, p. 107343, 2020.

\bibitem{chiou2022beyond}
E.~Chiou, E.~Panagiotaki, and I.~Kokkinos, ``{Beyond Deterministic Translation
  for Unsupervised Domain Adaptation},'' \emph{arXiv:2202.07778}, 2022.

\bibitem{araslanov2021self}
N.~Araslanov and S.~Roth, ``{Self-Supervised Augmentation Consistency for
  Adapting Semantic Segmentation},'' in \emph{Proc. of CVPR}, virtual, Jun.
  2021, pp. 15\,384--15\,394.

\bibitem{huang2021mlan}
J.~Huang, D.~Guan, S.~Lu, and A.~Xiao, ``{MLAN: Multi-Level Adversarial Network
  for Domain Adaptive Semantic Segmentation},'' 2021.

\bibitem{Li_2018_ECCV}
Y.~Li, M.-Y. Liu, X.~Li, M.-H. Yang, and J.~Kautz, ``{A Closed-Form Solution to
  Photorealistic Image Stylization},'' in \emph{Proc. of ECCV}, Munich,
  Germany, September 2018.

\bibitem{Li2017c}
Y.~Li, C.~Fang, J.~Yang, Z.~Wang, X.~Lu, and M.-H. Yang, ``{Universal Style
  Transfer via Feature Transforms},'' in \emph{Proc. of NeurIPS}, Long Beach,
  CA, USA, Dec. 2017, pp. 386--396.

\bibitem{Simonyan2015}
K.~Simonyan and A.~Zisserman, ``{Very Deep Convolutional Networks for
  Large-Scale Image Recognition},'' in \emph{Proc. of ICLR}, San Diego, CA,
  USA, May 2015, pp. 1--27.

\bibitem{Huang2018multimodal}
X.~Huang, M.-Y. Liu, S.~Belongie, and J.~Kautz, ``{Multimodal Unsupervised
  Image-to-Image Translation},'' in \emph{Proc. of ECCV}, Munich, Germany, Sep.
  2018, pp. 1--18.

\bibitem{Pizer1987histogram}
S.~M. Pizer, E.~P. Amburn, J.~D. Austin, R.~Cromartie, A.~Geselowitz, T.~Greer,
  B.~T.~H. Romeny, and J.~B. Zimmerman, ``{Adaptive Histogram Equalization and
  Its Variations},'' \emph{Comput. Vision Graph. Image Process.}, vol.~39,
  no.~3, pp. 355--368, sep 1987.

\bibitem{Mao2016}
X.~Mao, Q.~Li, H.~Xie, R.~Y.~K. Lau, Z.~Wang, and S.~P. Smolley, ``{Least
  Squares Generative Adversarial Networks},'' in \emph{Proc. of ICCV}, Venice,
  Italy, Oct. 2017, pp. 2794--2802.

\bibitem{Johnson2016Perceptual}
{Justin Johnson and Alexandre Alahi and Li Fei-Fei}, ``{Perceptual Losses for
  Real-Time Style Transfer and Super-Resolution},'' in \emph{Proc. of ECCV},
  Amsterdam, Netherlands, Oct. 2016, pp. 694--711.

\bibitem{Mirza2014}
M.~Mirza and S.~Osindero, ``{Conditional Generative Adversarial Nets},''
  \emph{arXiv}, Nov. 2014, (arXiv:1411.1784).

\bibitem{Choi2018stargan}
Y.~Choi, M.~Choi, M.~Kim, J.-W. Ha, S.~Kim, and J.~Choo, ``{StarGAN: Unified
  Generative Adversarial Networks for Multi-Domain Image-to-Image
  Translation},'' in \emph{Proc. of CVPR}, Salt Lake City, UT, USA, Jun. 2018,
  pp. 8789--8797.

\bibitem{Zhu2017bicyclegan}
J.-Y. Zhu, R.~Zhang, D.~Pathak, T.~Darrell, A.~A. Efros, O.~Wang, and
  E.~Shechtman, ``{Toward Multimodal Image-to-Image Translation},'' in
  \emph{Proc. of NeurIPS}, Long Beach, CA, USA, Dec. 2017, pp. 465--476.

\bibitem{Benaim2019}
S.~Benaim, M.~Khaitov, T.~Galanti, and L.~Wolf, ``{Domain Intersection and
  Domain Difference},'' in \emph{Proc. of ICCV}, Seoul, Korea, Oct. 2019, pp.
  3445--3453.

\bibitem{Olsson2021classmix}
V.~Olsson, W.~Tranheden, J.~Pinto, and L.~Svensson, ``{ClassMix:
  Segmentation-Based Data Augmentation for Semi-Supervised Learning},'' in
  \emph{Proc. of WACV}, Waikoloa, HI, USA, Jan. 2021, pp. 1369--1378.

\bibitem{yun2019cutmix}
S.~Yun, D.~Han, S.~J. Oh, S.~Chun, J.~Choe, and Y.~Yoo, ``{Cutmix:
  Regularization Strategy to Train Strong Classifiers With Localizable
  Features},'' in \emph{Proc. of ICCV}, Seoul, Korea, Oct. 2019, pp.
  6023--6032.

\bibitem{cardace2022plugging}
A.~Cardace, L.~De~Luigi, P.~Z. Ramirez, S.~Salti, and L.~Di~Stefano,
  ``{Plugging Self-Supervised Monocular Depth Into Unsupervised Domain
  Adaptation for Semantic Segmentation},'' in \emph{Proc. of WACV}, Waikoloa,
  HI, USA, Jan. 2022, pp. 1129--1139.

\bibitem{zhang2021prototypical}
P.~Zhang, B.~Zhang, T.~Zhang, D.~Chen, Y.~Wang, and F.~Wen, ``{Prototypical
  Pseudo Label Denoising and Target Structure Learning for Domain Adaptive
  Semantic Segmentation},'' \emph{arXiv:2101.10979}, 2021.

\bibitem{Hoffman2016}
J.~Hoffman, D.~Wang, F.~Yu, and T.~Darrell, ``{FCNs in the Wild: Pixel-Level
  Adversarial and Constraint-Based Adaptation},'' \emph{arXiv}, no. 1612.02649,
  Dec. 2016.

\bibitem{du2019ssf}
L.~Du, J.~Tan, H.~Yang, J.~Feng, X.~Xue, Q.~Zheng, X.~Ye, and X.~Zhang,
  ``{SSF-DAN: Separated Semantic Feature Based Domain Adaptation Network for
  Semantic Segmentation},'' in \emph{Proc. of ICCV}, Seoul, Korea, Oct. 2019,
  pp. 982--991.

\bibitem{li2020spatial}
C.~Li, D.~Du, L.~Zhang, L.~Wen, T.~Luo, Y.~Wu, and P.~Zhu, ``{Spatial Attention
  Pyramid Network for Unsupervised Domain Adaptation},'' in \emph{Proc. of
  ECCV}, virtual, Aug. 2020, pp. 481--497.

\bibitem{luo2019significance}
Y.~Luo, P.~Liu, T.~Guan, J.~Yu, and Y.~Yang, ``{Significance-Aware Information
  Bottleneck for Domain Adaptive Semantic Segmentation},'' in \emph{Proc. of
  ICCV}, Seoul, Korea, Oct. 2019, pp. 6778--6787.

\bibitem{luo2019taking}
Y.~Luo, L.~Zheng, T.~Guan, J.~Yu, and Y.~Yang, ``{Taking a Closer Look at
  Domain Shift: Category-Level Adversaries for Semantics Consistent Domain
  Adaptation},'' in \emph{Proc. of CVPR}, Long Beach, CA, USA, Jun. 2019, pp.
  2507--2516.

\bibitem{Hu2020SemanticDA}
D.~Hu, J.~Liang, Q.-B. Hou, H.~Yan, Y.~Chen, S.~Yan, and J.~Feng, ``{Semantic
  Domain Adversarial Networks for Unsupervised Domain Adaptation},'' 2020.

\bibitem{tsai2019domain}
Y.-H. Tsai, K.~Sohn, S.~Schulter, and M.~Chandraker, ``{Domain Adaptation for
  Structured Output via Discriminative Patch Representations},'' in \emph{Proc.
  of ICCV}, Seoul, Korea, Oct. 2019, pp. 1456--1465.

\bibitem{zheng2019deep}
Q.~Zheng, J.~Chen, Z.~Wang, J.~Jiang, and C.~Liang, ``{Deep Segmentation Domain
  Adaptation Network With Weighted Boundary Constraint},'' \emph{IEEE Access},
  vol.~7, pp. 93\,909--93\,918, 2019.

\bibitem{yu2021dast}
F.~Yu, M.~Zhang, H.~Dong, S.~Hu, B.~Dong, and L.~Zhang, ``{DAST: Unsupervised
  Domain Adaptation in Semantic Segmentation Based on Discriminator Attention
  and Self-Training},'' 2021.

\bibitem{wang2020classes}
H.~Wang, T.~Shen, W.~Zhang, L.-Y. Duan, and T.~Mei, ``{Classes Matter: A
  Fine-Grained Adversarial Approach to Cross-Domain Semantic Segmentation},''
  in \emph{Proc. of ECCV}, virtual, Aug. 2020, pp. 642--659.

\bibitem{luo2021away}
Y.~Luo, Z.~Wang, D.~Huang, N.~Ge, and J.~Lu, ``{Get Away From Style:
  Category-Guided Domain Adaptation for Semantic Segmentation},'' 2021.

\bibitem{dong2020cscl}
J.~Dong, Y.~Cong, G.~Sun, Y.~Liu, and X.~Xu, ``{CSCL: Critical
  Semantic-Consistent Learning for Unsupervised Domain Adaptation},'' in
  \emph{Proc. of ECCV}, virtual, Aug. 2020, pp. 745--762.

\bibitem{huang2018a}
H.~Huang, Q.~Huang, and P.~Krahenbuhl, ``{Domain Tansfer Through Deep
  Activation Matching},'' in \emph{Proc. of ECCV}, Munich, Germany, Sep. 2018,
  pp. 590--605.

\bibitem{wang2019class}
Y.~Wang, Y.~Li, J.~H. Elder, R.~Wu, and H.~Lu, ``{Class-Conditional Domain
  Adaptation on Semantic Segmentation},'' \emph{arXiv:1911.11981}, 2019.

\bibitem{chen2018road}
Y.~Chen, W.~Li, and L.~Van~Gool, ``{Road: Reality Oriented Adaptation for
  Semantic Segmentation of Urban Scenes},'' in \emph{Proc. of CVPR}, Salt Lake
  City, UT, USA, Jun. 2018, pp. 7892--7901.

\bibitem{chen2021unsupervised}
H.~Chen, C.~Wu, Y.~Xu, and B.~Du, ``{Unsupervised Domain Adaptation for
  Semantic Segmentation via Low-Level Edge Information Transfer},''
  \emph{arXiv:2109.08912}, 2021.

\bibitem{chen2020classification}
T.~Chen, J.~Zhang, G.-S. Xie, Y.~Yao, X.~Huang, and Z.~Tang, ``{Classification
  Constrained Discriminator for Domain Adaptive Semantic Segmentation},'' in
  \emph{Proc. of ICME}, virtual, Jul. 2020, pp. 1--6.

\bibitem{zhang2020transferring}
Y.~Zhang, Z.~Qiu, T.~Yao, C.-W. Ngo, D.~Liu, and T.~Mei, ``{Transferring and
  Regularizing Prediction for Semantic Segmentation},'' in \emph{Proc. of
  CVPR}, virtual, Jun. 2020, pp. 9621--9630.

\bibitem{saha2021learning}
S.~Saha, A.~Obukhov, D.~P. Paudel, M.~Kanakis, Y.~Chen, S.~Georgoulis, and
  L.~Van~Gool, ``{Learning to Relate Depth and Semantics for Unsupervised
  Domain Adaptation},'' in \emph{Proc. of CVPR}, virtual, Jul. 2021, pp.
  8197--8207.

\bibitem{Bolte2019a}
J.-A. Bolte, M.~Kamp, A.~Breuer, S.~Homoceanu, P.~Schlicht, F.~{H{\"u}ger},
  D.~Lipinski, and T.~Fing\-scheidt, ``{Unsupervised Domain Adaptation to
  Improve Image Segmentation Quality Both in the Source and Target Domain},''
  in \emph{Proc. of CVPR - Workshops}, Long Beach, CA, USA, Jun. 2019, pp.
  1404--1413.

\bibitem{li2020bi}
S.~Li, F.~Lv, B.~Xie, C.~H. Liu, J.~Liang, and C.~Qin, ``{Bi-Classifier
  Determinacy Maximization for Unsupervised Domain Adaptation},''
  \emph{arXiv:2012.06995}, 2020.

\bibitem{lee2019sliced}
C.-Y. Lee, T.~Batra, M.~H. Baig, and D.~Ulbricht, ``{Sliced Wasserstein
  Discrepancy for Unsupervised Domain Adaptation},'' in \emph{Proc. of CVPR},
  Long Beach, CA, USA, Jun. 2019, pp. 10\,285--10\,295.

\bibitem{gao2021dual}
L.~Gao, J.~Zhang, L.~Zhang, and D.~Tao, ``{DSP: Dual Soft-Paste for
  Unsupervised Domain Adaptive Semantic Segmentation},'' in \emph{Proc. of ACM
  International Conference on Multimedia}, virtual, Oct. 2021, pp. 2825--2833.

\bibitem{Hong2018}
W.~Hong, Z.~Wang, M.~Yang, and J.~Yuan, ``{Conditional Generative Adversarial
  Network for Structured Domain Adaptation},'' in \emph{Proc. of CVPR}, Salt
  Lake City, UT, USA, Jun. 2018, pp. 1335--1344.

\bibitem{zhu2018penalizing}
Y.~Zhu, H.~Zhou, C.~Yang, J.~Shi, and D.~Lin, ``{Penalizing Top Performers:
  Conservative Loss for Semantic Segmentation Adaptation},'' in \emph{Proc. of
  ECCV}, Munich, Germany, Sep. 2018, pp. 568--583.

\bibitem{ruan2019category}
\BIBentryALTinterwordspacing
C.~Ruan, W.~Wang, H.~Hu, and D.~Chen, ``{Category-Level Adversaries for
  Semantic Domain Adaptation},'' \emph{IEEE Access}, vol.~7, pp.
  83\,198--83\,208, 2019. [Online]. Available:
  \url{https://ieeexplore.ieee.org/stamp/stamp.jsp?arnumber=8731847}
\BIBentrySTDinterwordspacing

\bibitem{Romijnders2019}
R.~Romijnders, P.~Meletis, and G.~Dubbelman, ``{A Domain Agnostic Normalization
  Layer for Unsupervised Adversarial Domain Adaptation},'' in \emph{Proc. of
  WACV}, Waikoloa, Hawaii, Jan. 2019, pp. 1866--1875.

\bibitem{Ioffe2017}
S.~Ioffe, ``{Batch Renormalization: Towards Reducing Minibatch Dependence in
  Batch-Normalized Models},'' in \emph{Proc. of NIPS}, Long Beach, CA, USA,
  Dec. 2017, pp. 1945--1953.

\bibitem{Li2018d}
Y.~Li, N.~Wang, J.~Shi, X.~Hou, and J.~Liu, ``{Adaptive Batch Normalization for
  Practical Domain Adaptation},'' \emph{Pattern Recognition}, vol.~80, pp.
  109--117, Aug. 2018.

\bibitem{marsden2021contrastive}
R.~A. Marsden, A.~Bartler, M.~Döbler, and B.~Yang, ``{Contrastive Learning and
  Self-Training for Unsupervised Domain Adaptation in Semantic Segmentation},''
  2021.

\bibitem{xie2021spcl}
B.~Xie, K.~Yin, S.~Li, and X.~Chen, ``{SPCL: A New Framework for Domain
  Adaptive Semantic Segmentation via Semantic Prototype-Based Contrastive
  Learning},'' \emph{arXiv:2111.12358}, 2021.

\bibitem{shim2021learning}
D.~Shim and H.~J. Kim, ``{Learning a Domain-Agnostic Visual Representation for
  Autonomous Driving via Contrastive Loss},'' 2021.

\bibitem{liu2021domain}
W.~Liu, D.~Ferstl, S.~Schulter, L.~Zebedin, P.~Fua, and C.~Leistner, ``{Domain
  Adaptation for Semantic Segmentation via Patch-Wise Contrastive Learning},''
  \emph{arXiv:2104.11056}, 2021.

\bibitem{kang2020pixel}
G.~Kang, Y.~Wei, Y.~Yang, Y.~Zhuang, and A.~G. Hauptmann, ``{Pixel-Level Cycle
  Association: A New Perspective for Domain Adaptive Semantic Segmentation},''
  \emph{arXiv:2011.00147}, 2020.

\bibitem{huang2021category}
J.~Huang, D.~Guan, A.~Xiao, S.~Lu, and L.~Shao, ``{Category Contrast for
  Unsupervised Domain Adaptation in Visual Tasks},'' \emph{arXiv:2106.02885},
  2021.

\bibitem{niemeijer2021combining}
J.~Niemeijer and P.~Schäfer, ``{Combining Semantic Self-Supervision and
  Self-Training for Domain Adaptation in Semantic Segmentation},'' in
  \emph{Proc. of IV - Workshops}, virtual, Jul. 2021, pp. 364--371.

\bibitem{wang2021separable}
S.~Wang, D.~Zhao, Y.~Li, C.~Zhang, Y.~Guo, Q.~Zang, B.~Hou, and L.~Jiao,
  ``{More Separable and Easier to Segment: A Cluster Alignment Method for
  Cross-Domain Semantic Segmentation},'' 2021.

\bibitem{li2021semantic}
S.~Li, B.~Xie, B.~Zang, C.~H. Liu, X.~Cheng, R.~Yang, and G.~Wang, ``{Semantic
  Distribution-Aware Contrastive Adaptation for Semantic Segmentation},'' 2021.

\bibitem{guizilini2021geometric}
V.~Guizilini, J.~Li, R.~Ambrus, and A.~Gaidon, ``{Geometric Unsupervised Domain
  Adaptation for Semantic Segmentation},'' \emph{arXiv:2103.16694}, 2021.

\bibitem{sun2019unsupervised}
Y.~Sun, E.~Tzeng, T.~Darrell, and A.~A. Efros, ``{Unsupervised Domain
  Adaptation Through Self-Supervision},'' \emph{arXiv:1909.11825}, 2019.

\bibitem{xu2019self2}
J.~Xu, L.~Xiao, and A.~M. L{\'o}pez, ``{Self-Supervised Domain Adaptation for
  Computer Vision Tasks},'' \emph{IEEE Access}, vol.~7, pp. 156\,694--156\,706,
  2019.

\bibitem{lv2020cross}
F.~Lv, T.~Liang, X.~Chen, and G.~Lin, ``{Cross-Domain Semantic Segmentation via
  Domain-Invariant Interactive Relation Transfer},'' in \emph{Proc. of CVPR},
  Seattle, WA, USA, Jun. 2020, pp. 4334--4343.

\bibitem{Ganin2015}
Y.~Ganin and V.~Lempitsky, ``{Unsupervised Domain Adaptation by
  Backpropagation},'' in \emph{Proc. of ICML}, Lille, France, Jul. 2015, pp.
  1180--1189.

\bibitem{Goodfellow2014}
I.~Goodfellow, J.~Pouget-Abadie, M.~Mirza, B.~Xu, D.~Warde-Farley, S.~Ozair,
  A.~Courville, and Y.~Bengio, ``{Generative Adversarial Nets},'' in
  \emph{Proc. of NIPS}, Montr{\'{e}}al, Canada, Dec. 2014, pp. 2672--2680.

\bibitem{huang2018domain}
H.~Huang, Q.~Huang, and P.~Krahenbuhl, ``{Domain Transfer Through Deep
  Activation Matching},'' in \emph{Proc. of ECCV}, Munich, Germany, Sep. 2018,
  pp. 590--605.

\bibitem{zhang2020towards}
B.~Zhang, S.~Zhao, and R.~Zhang, ``{Towards Adaptive Semantic Segmentation by
  Progressive Feature Refinement},'' in \emph{Proc. of ICIP}, Abu Dhabi, United
  Arab Emirates, Oct. 2020, pp. 2221--2225.

\bibitem{gatys2016image}
L.~A. Gatys, A.~S. Ecker, and M.~Bethge, ``{Image Style Transfer Using
  Convolutional Neural Networks},'' in \emph{Proc. of CVPR}, Las Vegas, NV,
  USA, Jun. 2016, pp. 2414--2423.

\bibitem{SelfSuperSemSeg}
\BIBentryALTinterwordspacing
Y.~Sun, E.~Tzeng, T.~Darrell, and A.~A. Efros, ``{Unsupervised Domain
  Adaptation Through Self-Supervision},'' \emph{CoRR}, vol. abs/1909.11825,
  2019. [Online]. Available: \url{http://arxiv.org/abs/1909.11825}
\BIBentrySTDinterwordspacing

\bibitem{chen2020simple}
T.~Chen, S.~Kornblith, M.~Norouzi, and G.~Hinton, ``{A Simple Framework for
  Contrastive Learning of Visual Representations},'' in \emph{Proc. of ICML},
  virtual, Jul. 2020, pp. 1597--1607.

\bibitem{saito2020universal}
K.~Saito, D.~Kim, S.~Sclaroff, and K.~Saenko, ``{Universal Domain Adaptation
  Through Self Supervision},'' 2020.

\bibitem{barbato2021latent}
F.~Barbato, M.~Toldo, U.~Michieli, and P.~Zanuttigh, ``{Latent Space
  Regularization for Unsupervised Domain Adaptation in Semantic
  Segmentation},'' 2021.

\bibitem{Xu2019ICCV}
R.~Xu, G.~Li, J.~Yang, and L.~Lin, ``Larger norm more transferable: An adaptive
  feature norm approach for unsupervised domain adaptation,'' in
  \emph{Proceedings of the IEEE/CVF International Conference on Computer Vision
  (ICCV)}, October 2019.

\bibitem{stan2020unsupervised}
S.~Stan and M.~Rostami, ``{Unsupervised Model Adaptation for Continual Semantic
  Segmentation},'' \emph{arXiv:2009.12518}, 2020.

\bibitem{chen2019domain}
M.~Chen, H.~Xue, and D.~Cai, ``{Domain Adaptation for Semantic Segmentation
  With Maximum Squares Loss},'' in \emph{Proc. of ICCV}, Seoul, Korea, Oct.
  2019, pp. 2090--2099.

\bibitem{xie2022sepico}
B.~Xie, S.~Li, M.~Li, C.~H. Liu, G.~Huang, and G.~Wang, ``{SePiCo:
  Semantic-Guided Pixel Contrast for Domain Adaptive Semantic Segmentation},''
  \emph{arXiv:2204.08808}, 2022.

\bibitem{zou2019confidence}
Y.~Zou, Z.~Yu, X.~Liu, B.~Kumar, and J.~Wang, ``{Confidence Regularized
  Self-Training},'' in \emph{Proc. of ICCV}, Seoul, Korea, Oct. 2019, pp.
  5982--5991.

\bibitem{mei2020instance}
K.~Mei, C.~Zhu, J.~Zou, and S.~Zhang, ``{Instance Adaptive Self-Training for
  Unsupervised Domain Adaptation},'' pp. 415--430, Aug. 2020.

\bibitem{li2020content}
G.~Li, G.~Kang, W.~Liu, Y.~Wei, and Y.~Yang, ``{Content-Consistent Matching for
  Domain Adaptive Semantic Segmentation},'' in \emph{Proc. of ECCV}, virtual,
  Aug. 2020, pp. 440--456.

\bibitem{zheng2019unsupervised}
Z.~Zheng and Y.~Yang, ``{Unsupervised Scene Adaptation With Memory
  Regularization in Vivo},'' \emph{arXiv:1912.11164}, 2019.

\bibitem{chung2021exploiting}
I.~Chung, J.~Yoo, and N.~Kwak, ``{Exploiting Inter-Pixel Correlations in
  Unsupervised Domain Adaptation for Semantic Segmentation},''
  \emph{arXiv:2110.10916}, 2021.

\bibitem{huang2020contextual}
J.~Huang, S.~Lu, D.~Guan, and X.~Zhang, ``{Contextual-Relation Consistent
  Domain Adaptation for Semantic Segmentation},'' in \emph{Proc of ECCV},
  Glasow, UK, Aug. 2020, pp. 705--722.

\bibitem{zhang2017a}
K.~Zhang, W.~Zuo, S.~Gu, and L.~Zhang, ``{Learning Deep CNN Denoiser Prior for
  Image Restoration},'' in \emph{Proc. of CVPR}, Honulu, HI, USA, Jul. 2017,
  pp. 3929--3938.

\bibitem{liu2021adversarial}
X.~Liu, Z.~Guo, S.~Li, F.~Xing, J.~You, C.-C.~J. Kuo, G.~El~Fakhri, and J.~Woo,
  ``{Adversarial Unsupervised Domain Adaptation With Conditional and Label
  Shift: Infer, Align and Iterate},'' in \emph{Proc. of ICCV}, virtual, Oct.
  2021, pp. 10\,367--10\,376.

\bibitem{guan2021scale}
D.~Guan, J.~Huang, S.~Lu, and A.~Xiao, ``{Scale Variance Minimization for
  Unsupervised Domain Adaptation in Image Segmentation},'' \emph{Pattern
  Recognition}, vol. 112, p. 107764, 2021.

\bibitem{Subhani2020}
M.~N. Subhani and M.~Ali, ``{Learning From Scale-Invariant Examples for Domain
  Adaptation in Semantic Segmentation},'' in \emph{Proc. of ECCV}, Glasgow, UK,
  Aug. 2020, pp. 290--306.

\bibitem{wang2021cross}
Z.~Wang, X.~Liu, M.~Suganuma, and T.~Okatani, ``{Cross-Region Domain Adaptation
  for Class-Level Alignment},'' \emph{arXiv:2109.06422}, 2021.

\bibitem{wang2021uncertainty}
Y.~Wang, J.~Peng, and Z.~Zhang, ``{Uncertainty-Aware Pseudo Label Refinery for
  Domain Adaptive Semantic Segmentation},'' in \emph{Proc. of ICCV}, virtual,
  Oct. 2021, pp. 9092--9101.

\bibitem{truong2021bimal}
T.-D. Truong, C.~N. Duong, N.~Le, S.~L. Phung, C.~Rainwater, and K.~Luu,
  ``{BiMal: Bijective Maximum Likelihood Approach to Domain Adaptation in
  Semantic Scene Segmentation},'' in \emph{Proc. of ICCV}, virtual, Oct. 2021,
  pp. 8548--8557.

\bibitem{pan2020unsupervised}
F.~Pan, I.~Shin, F.~Rameau, S.~Lee, and I.~S. Kweon, ``{Unsupervised
  Intra-Domain Adaptation for Semantic Segmentation Through
  Self-Supervision},'' in \emph{Proc. of CVPR}, Seattle, WA, USA, Jun. 2020,
  pp. 3764--3773.

\bibitem{yang2020adversarial}
J.~Yang, R.~Xu, R.~Li, X.~Qi, X.~Shen, G.~Li, and L.~Lin, ``{An Adversarial
  Perturbation Oriented Domain Adaptation Approach for Semantic
  Segmentation},'' in \emph{Proc. of AAAI}, vol.~34, no.~07, 2020, pp.
  12\,613--12\,620.

\bibitem{zheng2021rectifying}
Z.~Zheng and Y.~Yang, ``{Rectifying Pseudo Label Learning via Uncertainty
  Estimation for Domain Adaptive Semantic Segmentation},'' \emph{International
  Journal of Computer Vision}, pp. 1--15, 2021.

\bibitem{xu2022unsupervised}
W.~Xu, Z.~Wang, and W.~Bian, ``{Unsupervised Domain Adaptation With Implicit
  Pseudo Supervision for Semantic Segmentation},'' \emph{arXiv:2204.06747},
  2022.

\bibitem{michieli2020adversarial}
U.~Michieli, M.~Biasetton, G.~Agresti, and P.~Zanuttigh, ``{Adversarial
  Learning and Self-Teaching Techniques for Domain Adaptation in Semantic
  Segmentation},'' \emph{IEEE Transactions on Intelligent Vehicles}, vol.~5,
  no.~3, pp. 508--518, 2020.

\bibitem{spadotto2020unsupervised}
T.~Spadotto, M.~Toldo, U.~Michieli, and P.~Zanuttigh, ``{Unsupervised Domain
  Adaptation With Multiple Domain Discriminators and Adaptive Self-Training},''
  \emph{arXiv:2004.12724}, 2020.

\bibitem{shen2019regularizing}
T.~Shen, D.~Gong, W.~Zhang, C.~Shen, and T.~Mei, ``{Regularizing Proxies With
  Multi-Adversarial Training for Unsupervised Domain-Adaptive Semantic
  Segmentation},'' \emph{arXiv:1907.12282}, 2019.

\bibitem{tsai2018learning}
Y.-H. Tsai, W.-C. Hung, S.~Schulter, K.~Sohn, M.-H. Yang, and M.~Chandraker,
  ``{Learning to Adapt Structured Output Space for Semantic Segmentation},'' in
  \emph{Proc. of CVPR}, Salt Lake City, UT, USA, Jun. 2018, pp. 7472--7481.

\bibitem{liu2021source}
Y.~Liu, W.~Zhang, and J.~Wang, ``{Source-Free Domain Adaptation for Semantic
  Segmentation},'' \emph{arXiv:2103.16372}, 2021.

\bibitem{zhang2020joint}
\BIBentryALTinterwordspacing
Y.~Zhang and Z.~Wang, ``{Joint Adversarial Learning for Domain Adaptation in
  Semantic Segmentation},'' in \emph{Proceedings of the AAAI Conference on
  Artificial Intelligence}, vol.~34, no.~04, 2020, pp. 6877--6884. [Online].
  Available: \url{https://ojs.aaai.org/index.php/AAAI/article/view/6169/6025}
\BIBentrySTDinterwordspacing

\bibitem{vu2019a}
\BIBentryALTinterwordspacing
T.-H. Vu, H.~Jain, M.~Bucher, M.~Cord, and P.~P{\'e}rez, ``{DADA: Depth-Aware
  Domain Adaptation in Semantic Segmentation},'' in \emph{Proceedings of the
  IEEE/CVF International Conference on Computer Vision}, Seoul, Korea, Oct.
  2019, pp. 7364--7373. [Online]. Available:
  \url{https://openaccess.thecvf.com/content_ICCV_2019/papers/Vu_DADA_Depth-Aware_Domain_Adaptation_in_Semantic_Segmentation_ICCV_2019_paper.pdf}
\BIBentrySTDinterwordspacing

\bibitem{yan2021pixel}
Z.~Yan, X.~Yu, Y.~Qin, Y.~Wu, X.~Han, and S.~Cui, ``{Pixel-Level Intra-Domain
  Adaptation for Semantic Segmentation},'' in \emph{Proc. of ACM International
  Conference on Multimedia}, 2021, pp. 404--413.

\bibitem{cicek2020spatial}
S.~Cicek, N.~Xu, Z.~Wang, H.~Jin, and S.~Soatto, ``{Spatial Class Distribution
  Shift in Unsupervised Domain Adaptation: Local Alignment Comes to Rescue},''
  in \emph{Proc. of the ACCV}, Kyoto, Japan, Dec. 2020.

\bibitem{tang2020towards}
H.~Tang, X.~Zhu, K.~Chen, K.~Jia, and C.~Chen, ``{Towards Uncovering the
  Intrinsic Data Structures for Unsupervised Domain Adaptation Using
  Structurally Regularized Deep Clustering},'' \emph{arXiv:2012.04280}, 2020.

\bibitem{vu2019advent}
T.-H. Vu, H.~Jain, M.~Bucher, M.~Cord, and P.~P{\'e}rez, ``{ADVENT: Adversarial
  Entropy Minimization for Domain Adaptation in Semantic Segmentation},'' in
  \emph{Proc. of CVPR}, Long Beach, CA, USA, Jun. 2019, pp. 2517--2526.

\bibitem{zhang2021multiple}
K.~Zhang, Y.~Sun, R.~Wang, H.~Li, and X.~Hu, ``{Multiple Fusion Adaptation: A
  Strong Framework for Unsupervised Semantic Segmentation Adaptation},''
  \emph{arXiv:2112.00295}, 2021.

\bibitem{xu2019self}
Y.~Xu, B.~Du, L.~Zhang, Q.~Zhang, G.~Wang, and L.~Zhang, ``{Self-Ensembling
  Attention Networks: Addressing Domain Shift for Semantic Segmentation},'' in
  \emph{Proc. of AAAI}, Honolulu, HI, USA, Jan. 2019, pp. 5581--5588.

\bibitem{Vu2019dada}
T.-H. Vu, H.~Jain, M.~Bucher, M.~Cord, and P.~P\'{e}rez, ``{DADA: Depth-Aware
  Domain Adaptation in Semantic Segmentation},'' in \emph{Proc. of ICCV},
  Seoul, Korea, Oct. 2019, pp. 7364--7373.

\bibitem{zhang2019category}
Q.~Zhang, J.~Zhang, W.~Liu, and D.~Tao, ``{Category Anchor-Guided Unsupervised
  Domain Adaptation for Semantic Segmentation},'' \emph{arXiv:1910.13049},
  2019.

\bibitem{li2018pyramid}
H.~Li, P.~Xiong, J.~An, and L.~Wang, ``{Pyramid Attention Network for Semantic
  Segmentation},'' \emph{arXiv}, 2018, (arXiv:1805.10180).

\bibitem{bruhn2005towards}
A.~Bruhn and J.~Weickert, ``{Towards Ultimate Motion Estimation: Combining
  Highest Accuracy With Real-Time Performance},'' in \emph{Proc. of ICCV},
  vol.~1, 2005, pp. 749--755.

\bibitem{abdar2021review}
M.~Abdar, F.~Pourpanah, S.~Hussain, D.~Rezazadegan, L.~Liu, M.~Ghavamzadeh,
  P.~Fieguth, X.~Cao, A.~Khosravi, U.~R. Acharya \emph{et~al.}, ``{A Review of
  Uncertainty Quantification in Deep Learning: Techniques, Applications and
  Challenges},'' \emph{Information Fusion}, vol.~76, pp. 243--297, 2021.

\bibitem{lu2020stochastic}
Z.~Lu, Y.~Yang, X.~Zhu, C.~Liu, Y.-Z. Song, and T.~Xiang, ``Stochastic
  classifiers for unsupervised domain adaptation,'' in \emph{Proceedings of the
  IEEE/CVF Conference on Computer Vision and Pattern Recognition}, 2020, pp.
  9111--9120.

\bibitem{guo2021metacorrection}
X.~Guo, C.~Yang, B.~Li, and Y.~Yuan, ``{MetaCorrection: Domain-Aware Meta Loss
  Correction for Unsupervised Domain Adaptation in Semantic Segmentation},''
  2021.

\bibitem{ester1996density}
M.~Ester, H.-P. Kriegel, J.~Sander, X.~Xu \emph{et~al.}, ``{A Density-Based
  Algorithm for Discovering Clusters in Large Spatial Databases With Noise.}''
  in \emph{Proc. of KDD}, Portland, OR, USA, Aug. 1996, pp. 226--231.

\bibitem{choi2019a}
\BIBentryALTinterwordspacing
J.~Choi, T.~Kim, and C.~Kim, ``{Self-Ensembling With Gan-Based Data
  Augmentation for Domain Adaptation in Semantic Segmentation},'' in
  \emph{Proceedings of the IEEE/CVF International Conference on Computer
  Vision}, Seoul, Korea, Oct. 2019, pp. 6830--6840. [Online]. Available:
  \url{https://openaccess.thecvf.com/content_ICCV_2019/papers/Choi_Self-Ensembling_With_GAN-Based_Data_Augmentation_for_Domain_Adaptation_in_Semantic_ICCV_2019_paper.pdf}
\BIBentrySTDinterwordspacing

\bibitem{Niemeijer2022LowCoplex}
\BIBentryALTinterwordspacing
J.~Niemeijer and J.~P. Sch\"afer, ``Domain adaptation and generalization: A
  low-complexity approach,'' in \emph{Proceedings of The 6th Conference on
  Robot Learning}, ser. Proceedings of Machine Learning Research, K.~Liu,
  D.~Kulic, and J.~Ichnowski, Eds., vol. 205.\hskip 1em plus 0.5em minus
  0.4em\relax PMLR, 14--18 Dec 2022, pp. 1081--1091. [Online]. Available:
  \url{https://proceedings.mlr.press/v205/niemeijer23a.html}
\BIBentrySTDinterwordspacing

\bibitem{mikolov2013distributed}
T.~Mikolov, I.~Sutskever, K.~Chen, G.~S. Corrado, and J.~Dean, ``{Distributed
  Representations of Words and Phrases and Their Compositionality},''
  \emph{Advances in neural information processing systems}, vol.~26, 2013.

\bibitem{vaswani2017attention}
A.~Vaswani, N.~Shazeer, N.~Parmar, J.~Uszkoreit, L.~Jones, A.~N. Gomez,
  {\L}.~Kaiser, and I.~Polosukhin, ``{Attention Is All You Need},'' in
  \emph{Proc. of NIPS}, Long Beach, CA, USA, Dec. 2017, pp. 5998--6008.

\bibitem{dosovitskiy2020image}
A.~Dosovitskiy, L.~Beyer, A.~Kolesnikov, D.~Weissenborn, X.~Zhai,
  T.~Unterthiner, M.~Dehghani, M.~Minderer, G.~Heigold, S.~Gelly \emph{et~al.},
  ``{An Image Is Worth 16x16 Words: Transformers for Image Recognition at
  Scale},'' \emph{arXiv:2010.11929}, 2020.

\bibitem{khan2022transformers}
S.~Khan, M.~Naseer, M.~Hayat, S.~W. Zamir, F.~S. Khan, and M.~Shah,
  ``{Transformers in Vision: A Survey},'' \emph{ACM computing surveys (CSUR)},
  vol.~54, no. 10s, pp. 1--41, 2022.

\bibitem{wang2021pyramid}
W.~Wang, E.~Xie, X.~Li, D.-P. Fan, K.~Song, D.~Liang, T.~Lu, P.~Luo, and
  L.~Shao, ``{Pyramid Vision Transformer: A Versatile Backbone for Dense
  Prediction Without Convolutions},'' in \emph{Proc. of ICCV}, virtual, Oct.
  2021, pp. 568--578.

\bibitem{gu2022multi}
J.~Gu, H.~Kwon, D.~Wang, W.~Ye, M.~Li, Y.-H. Chen, L.~Lai, V.~Chandra, and
  D.~Z. Pan, ``{Multi-Scale High-Resolution Vision Transformer for Semantic
  Segmentation},'' in \emph{Proc. of CVPR}, 2022, pp. 12\,094--12\,103.

\bibitem{shao2021adversarial}
R.~Shao, Z.~Shi, J.~Yi, P.-Y. Chen, and C.-J. Hsieh, ``{On the Adversarial
  Robustness of Vision Transformers},'' \emph{arXiv:2103.15670}, 2021.

\bibitem{naseer2021intriguing}
M.~M. Naseer, K.~Ranasinghe, S.~H. Khan, M.~Hayat, F.~Shahbaz~Khan, and M.-H.
  Yang, ``{Intriguing Properties of Vision Transformers},'' in \emph{Proc. of
  NeurIPS}, virtual, Dec. 2021, pp. 23\,296--23\,308.

\bibitem{mao2022towards}
X.~Mao, G.~Qi, Y.~Chen, X.~Li, R.~Duan, S.~Ye, Y.~He, and H.~Xue, ``{Towards
  Robust Vision Transformer},'' in \emph{Proc. of CVPR}, New Orleans, LA, USA,
  Jun. 2022, pp. 12\,042--12\,051.

\bibitem{bai2021transformers}
Y.~Bai, J.~Mei, A.~L. Yuille, and C.~Xie, ``{Are Transformers More Robust Than
  CNNs?}'' \emph{Advances in Neural Information Processing Systems}, vol.~34,
  pp. 26\,831--26\,843, 2021.

\bibitem{wang2022can}
Z.~Wang, Y.~Bai, Y.~Zhou, and C.~Xie, ``{Can CNNs Be More Robust Than
  Transformers?}'' \emph{arXiv:2206.03452}, 2022.

\bibitem{liu2022convnet}
Z.~Liu, H.~Mao, C.-Y. Wu, C.~Feichtenhofer, T.~Darrell, and S.~Xie, ``{A
  ConvNet for the 2020s},'' in \emph{Proc. of CVPR}, New Orleans, LA, USA, Jun.
  2022, pp. 11\,976--11\,986.

\bibitem{raghu2021vision}
M.~Raghu, T.~Unterthiner, S.~Kornblith, C.~Zhang, and A.~Dosovitskiy, ``{Do
  Vision Transformers See Like Convolutional Neural Networks?}'' in \emph{Proc.
  of NeurIPS}, virtual, Dec. 2021, pp. 12\,116--12\,128.

\bibitem{geirhos2018imagenet}
R.~Geirhos, P.~Rubisch, C.~Michaelis, M.~Bethge, F.~A. Wichmann, and
  W.~Brendel, ``{ImageNet-Trained CNNs Are Biased Towards Texture; Increasing
  Shape Bias Improves Accuracy and Robustness},'' \emph{arXiv:1811.12231},
  2018.

\bibitem{Chen2018a}
L.-C. Chen, Y.~Zhu, G.~Papandreou, F.~Schroff, and H.~Adam, ``{Encoder-Decoder
  With Atrous Separable Convolution for Semantic Image Segmentation},'' in
  \emph{Proc. of ECCV}, Munich, Germany, Sep. 2018, pp. 801--818.

\bibitem{Hoyer2022HRDA}
L.~Hoyer, D.~Dai, and L.~Van~Gool, ``{HRDA: Context-Aware High-Resolution
  Domain-Adaptive Semantic Segmentation},'' 2022.

\bibitem{ettedgui2022procst}
S.~Ettedgui, S.~Abu-Hussein, and R.~Giryes, ``{ProCST: Boosting Semantic
  Segmentation Using Progressive Cyclic Style-Transfer},''
  \emph{arXiv:2204.11891}, 2022.

\bibitem{vayyat2022cluda}
M.~Vayyat, J.~Kasi, A.~Bhattacharya, S.~Ahmed, and R.~Tallamraju, ``{CLUDA:
  Contrastive Learning in Unsupervised Domain Adaptation for Semantic
  Segmentation},'' \emph{arXiv:2208.14227}, 2022.

\bibitem{du2022learning}
Y.~Du, Y.~Shen, H.~Wang, J.~Fei, W.~Li, L.~Wu, R.~Zhao, Z.~Fu, and Q.~Liu,
  ``{Learning From Future: A Novel Self-Training Framework for Semantic
  Segmentation},'' \emph{arXiv:2209.06993}, 2022.

\bibitem{van2008visualizing}
L.~Van~der Maaten and G.~Hinton, ``{ Visualizing High-Dimensional Data Using
  t-SNE},'' \emph{Journal of Machine Learning Research}, vol.~9, no.~11, 2008.

\bibitem{geyer2020a2d2}
\BIBentryALTinterwordspacing
J.~Geyer, Y.~Kassahun, M.~Mahmudi, X.~Ricou, R.~Durgesh, A.~S. Chung,
  L.~Hauswald, V.~H. Pham, M.~M{\"u}hlegg, S.~Dorn, T.~Fernandez,
  M.~J{\"a}nicke, S.~Mirashi, C.~Savani, M.~Sturm, O.~Vorobiov, M.~Oelker,
  S.~Garreis, and P.~Schuberth, ``{A2D2: Audi Autonomous Driving Dataset},''
  2020. [Online]. Available: \url{https://www.a2d2.audi}
\BIBentrySTDinterwordspacing

\bibitem{yu2020bdd100k}
F.~Yu, H.~Chen, X.~Wang, W.~Xian, Y.~Chen, F.~Liu, V.~Madhavan, and T.~Darrell,
  ``{BDD100K: A Diverse Driving Dataset for Heterogeneous Multitask
  Learning},'' in \emph{Proc. of CVPR}, virtual, Jun. 2020, pp. 1--14.

\bibitem{koh2023consistency}
K.~B. Koh and B.~Fernando, ``Consistency regularization for domain
  adaptation,'' in \emph{Computer Vision--ECCV 2022 Workshops: Tel Aviv,
  Israel, October 23--27, 2022, Proceedings, Part VIII}.\hskip 1em plus 0.5em
  minus 0.4em\relax Springer, 2023, pp. 347--359.

\bibitem{chen2022pipa}
M.~Chen, Z.~Zheng, Y.~Yang, and T.-S. Chua, ``Pipa: Pixel-and patch-wise
  self-supervised learning for domain adaptative semantic segmentation,''
  \emph{arXiv preprint arXiv:2211.07609}, 2022.

\bibitem{wang2022exploring}
K.~Wang, D.~Kim, R.~Feris, K.~Saenko, and M.~Betke, ``Exploring consistency in
  cross-domain transformer for domain adaptive semantic segmentation,''
  \emph{arXiv preprint arXiv:2211.14703}, 2022.

\bibitem{hoyer2022mic}
L.~Hoyer, D.~Dai, H.~Wang, and L.~Van~Gool, ``Mic: Masked image consistency for
  context-enhanced domain adaptation,'' \emph{arXiv preprint arXiv:2212.01322},
  2022.

\bibitem{chen2022deliberated}
L.~Chen, Z.~Wei, X.~Jin, H.~Chen, M.~Zheng, K.~Chen, and Y.~Jin, ``Deliberated
  domain bridging for domain adaptive semantic segmentation,'' \emph{arXiv
  preprint arXiv:2209.07695}, 2022.

\bibitem{liu2022undoing}
Y.~Liu, J.~Deng, J.~Tao, T.~Chu, L.~Duan, and W.~Li, ``Undoing the damage of
  label shift for cross-domain semantic segmentation,'' in \emph{Proceedings of
  the IEEE/CVF Conference on Computer Vision and Pattern Recognition}, 2022,
  pp. 7042--7052.

\bibitem{pytorch_doc_determinism}
{PyTorch Contributors}, ``{PyTorch Documentation: Reproducibility},''
  \url{https://pytorch.org/docs/stable/notes/randomness.html}, 2022, accessed:
  2022-09-19.

\bibitem{marcus2018deep}
G.~Marcus, ``{Deep Learning: A Critical Appraisal},'' \emph{arXiv:1801.00631},
  2018.

\bibitem{triess2021survey}
L.~T. Triess, M.~Dreissig, C.~B. Rist, and J.~M. Z{\"o}llner, ``{A Survey on
  Deep Domain Adaptation for LiDAR Perception},'' in \emph{Proc. of IV -
  Workshops}, virtual, Jul. 2021, pp. 350--357.

\end{thebibliography}

\EOD

\end{document}